\documentclass[10pt,twocolumn,letterpaper]{article}
\usepackage[symbol]{footmisc}
\usepackage{float}
\usepackage{iccv2023}
\usepackage{times}
\usepackage{epsfig}
\usepackage{graphicx}
\usepackage{amsmath}
\usepackage{amssymb}
\usepackage{subcaption}
\usepackage{placeins}
\usepackage{makecell}
\usepackage{xcolor}
\usepackage{booktabs}

\definecolor{human_color}{RGB}{255,0,0}
\definecolor{background_color}{RGB}{29,162,216}
\definecolor{fish_color}{RGB}{255,255,0}
\definecolor{sand_color}{RGB}{194,178,128}
\definecolor{rubble_color}{RGB}{161,153,128}
\definecolor{unknown_hard_substrate_color}{RGB}{125,125,125}
\definecolor{algae_covered_substrate_color}{RGB}{125,163,125}
\definecolor{dark_color}{RGB}{31,31,31}
\definecolor{branching_bleached_color}{RGB}{252,231,240}
\definecolor{branching_dead_color}{RGB}{123,50,86}
\definecolor{branching_alive_color}{RGB}{226,91,157}
\definecolor{stylophora_alive_color}{RGB}{255,111,194}
\definecolor{pocillopora_alive_color}{RGB}{255,146,150}
\definecolor{acropora_alive_color}{RGB}{236,128,255}
\definecolor{table_acropora_alive_color}{RGB}{189,119,255}
\definecolor{table_acropora_dead_color}{RGB}{85,53,116}
\definecolor{millepora_color}{RGB}{244,150,115}
\definecolor{turbinaria_color}{RGB}{228,255,119}
\definecolor{other_coral_bleached_color}{RGB}{250,224,225}
\definecolor{other_coral_dead_color}{RGB}{114,60,61}
\definecolor{other_coral_alive_color}{RGB}{224,118,119}
\definecolor{massive_meandering_alive_color}{RGB}{236,150,21}
\definecolor{massive_meandering_dead_color}{RGB}{134,86,18}
\definecolor{massive_meandering_bleached_color}{RGB}{255,248,228}
\definecolor{meandering_alive_color}{RGB}{230,193,0}
\definecolor{meandering_dead_color}{RGB}{119,100,14}
\definecolor{meandering_bleached_color}{RGB}{251,243,216}
\definecolor{transect_line_color}{RGB}{0,255,0}
\definecolor{transect_tools_color}{RGB}{8,205,12}
\definecolor{sea_urchin_color}{RGB}{0,142,255}
\definecolor{sea_cucumber_color}{RGB}{0,231,255}
\definecolor{anemone_color}{RGB}{0,255,189}
\definecolor{sponge_color}{RGB}{240,80,80}
\definecolor{clam_color}{RGB}{189,255,234}
\definecolor{other_animal_color}{RGB}{0,255,255}
\definecolor{trash_color}{RGB}{255,0,134}
\definecolor{seagrass_color}{RGB}{125,222,125}
\definecolor{crown_of_thorn_color}{RGB}{179,245,234}
\definecolor{dead_clam_color}{RGB}{89,155,134}

\usepackage[pagebackref=true,breaklinks=true,letterpaper=true,colorlinks,bookmarks=false]{hyperref}
\iccvfinalcopy % *** Uncomment this line for the final submission

\def\iccvPaperID{*****} % *** Enter the ICCV Paper ID here

\ificcvfinal\fi

\newcommand{\red}[1]{\textcolor[RGB]{224, 118, 119}{#1}}
\newcommand{\green}[1]{\textcolor[RGB]{75,166,211}{#1}}

\begin{document}

%%%%%%%%% TITLE
\title{The Coralscapes Dataset: Semantic Scene Understanding in Coral Reefs}

\author{%
  \vspace{-10pt}
  Jonathan Sauder$^{*,1,2}$
  \and Viktor Domazetoski$^{*,3,4}$
  \and Guilhem Banc-Prandi$^{2}$ 
  \and Gabriela Perna$^{2}$
  \and Anders Meibom$^{2,5}$
  \and Devis Tuia$^{1}$ 
  \and \vspace{-5pt}\scriptsize{$^{1}$Environmental Computational Science and Earth Observation Laboratory, École Polytechnique Fédérale de Lausanne, Switzerland}\\\vspace{-5pt}
    \scriptsize{$^{2}$Laboratory for Biological Geochemistry, École Polytechnique Fédérale de Lausanne, Switzerland}\\\vspace{-5pt}
    \scriptsize{$^{3}$Centre for Ecology and Conservation, University of Exeter, United Kingdom}\\\vspace{-5pt}
    \scriptsize{$^{4}$School of the Environment, The University of Queensland, Australia}\\\vspace{-5pt}
    \scriptsize{$^{5}$Center for Advanced Surface Analysis, University of Lausanne, Switzerland}\\\vspace{-5pt}
  \scriptsize{$\mbox{}^*$ Equal contribution}
  \vspace{-20pt}
}

%\and

\twocolumn[{\maketitle
\begin{center}
    \centering
%    \begin{figure*}[t!]
\includegraphics[trim={0px 10px 0px 10px},clip,width=0.32\textwidth]{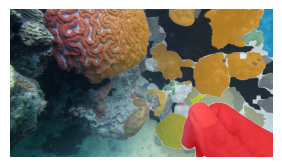}
\includegraphics[trim={0px 10px 0px 10px},clip,width=0.32\textwidth]{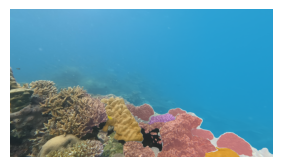}
\includegraphics[trim={0px 10px 0px 10px},clip,width=0.32\textwidth]{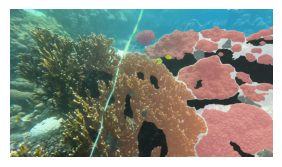}
%\end{figure*}
\end{center}
}]

\ificcvfinal\thispagestyle{empty}\fi

%%%%%%%%% ABSTRACT
\begin{abstract}
Coral reefs are declining worldwide due to climate change and local stressors. To inform effective conservation or restoration, monitoring at the highest possible spatial and temporal resolution is necessary. Conventional coral reef surveying methods are limited in scalability due to their reliance on expert labor time, motivating the use of computer vision tools to automate the identification and abundance estimation of live corals from images. However, the design and evaluation of such tools has been impeded by the lack of large high quality datasets. We release the Coralscapes dataset, the first general-purpose dense semantic segmentation dataset for coral reefs, covering 2075 images, 39 benthic classes, and 174k segmentation masks annotated by experts. Coralscapes has a similar scope and the same structure as the widely used Cityscapes dataset for urban scene segmentation, allowing benchmarking of semantic segmentation models in a new challenging domain which requires expert knowledge to annotate. We benchmark a wide range of semantic segmentation models, and find that transfer learning from Coralscapes to existing smaller datasets consistently leads to state-of-the-art performance. Coralscapes will catalyze research on efficient, scalable, and standardized coral reef surveying methods based on computer vision, and holds the potential to streamline the development of underwater ecological robotics.
\end{abstract}

%%%%%%%%% BODY TEXT
\vspace{-5pt}
\section{Introduction}
\vspace{-5pt}

Coral reefs are under unprecedented stress from both global anthropogenic climate change and local human activities such as overtourism, pollution, or destructive fishing practices \cite{ ipcc21}. Coral reefs, which host more than a third of all marine biodiversity on less than 0.1\% of the world's oceans'
surface \cite{blabla}, are among the most vulnerable ecosystems on the planet: under the current greenhouse gas emission trajectory, more than 99\% of warm-water coral reefs are projected to lose significant area or suffer local extinction \cite{ipcc15, future}. This could cascade into catastrophic impacts for the more than 500 million people that rely on coral reef-related ecosystem services through food security, coastal protection, or tourism \cite{halfabillion}.

However, there are regions, species, and genotypes within species that are more heat-resistant, holding promise to withstand the projected ocean warming induced by climate change \cite{beyer2018risk}. The reefs of the northern Red Sea, for example, are projected to resist up to 5°C of warming \cite{maoz, voolstra, kruegerredsea, osmanredsea, savaryredsea}. These heat-resistant reefs must be protected from local stressors to ensure the survival of coral reef ecosystems on the planet.
To inform effective conservation or restoration strategies, reefs must be monitored at the highest spatial and temporal resolution. As aerial and satellite remote sensing methods are limited in resolution, constrained to very shallow areas, and rely on good atmospheric, weather, and water conditions \cite{satellite1,satellite2,drone1,drone2}, \textit{in situ} surveying remains the de facto method for monitoring coral reefs. 

Computer vision emerges as a promising tool to scale coral reef monitoring, through tackling the bottleneck imposed by the needed expert time for identifying corals and estimating their abundance on images \cite{beijbom2012automated,beijbom2015towards}, as well as the development of visual simultaneous localization and mapping (SLAM) systems \cite{orbslam,mast3r-slam} that can be used on autonomous underwater vehicles (AUVs) and other robots \cite{AUV1,AUV2}. However, existing datasets in the coral reef domain are limited to narrow domains such as orthomosaics or photo quadrats \cite{alonso2019coralseg,edwards}, limited to sparse point labels \cite{beijbom2012automated,beijbom2015towards} or image classification \cite{raphael2020deep,bof}, or in their size and diversity, inhibiting the widespread deployment of computer vision in coral reef science and conservation. 

The main challenge to creating large-scale classification/segmentation datasets in coral reefs is the annotation process. 
Annotating coral reefs requires trained experts and cannot be done by crowd-sourced click-workers. Corals display high morphologic plasticity \cite{alonso2019coralseg}, and the appearance of corals can vary drastically between biogeographic regions.
In general, precise phylogenetic taxonomy of coral species or even genera can simply not be determined from images, even in close-up scenarios in good conditions, limiting the classification hierarchy to morphological attributes, such as growth forms (e.g. branching, meandering, massive). To add to the challenge, corals may appear bleached or dead, followed by various stages of degradation (e.g. overgrown by algae), or decomposing into rocky substrate or rubble.
Besides corals, reefs host a wide range of life, including fish, invertebrates (sea urchins, sea cucumbers, anemones, sponges etc.), plants (algae, seagrass), and many other organisms. Finally, the identifiability of benthic classes in reefs is strongly impacted by the degradation of color and induced blur from the water column. Tackling these challenges to create a consistently labeled large-scale dataset requires a concerted effort by trained expert annotators.

We propose the Coralscapes dataset for semantic scene understanding in coral reefs. Coralscapes serves two main purposes: i) to drive the development of computer vision applications in reef conservation, including automated surveying methods and underwater robotics, and ii) to allow benchmarking of state-of-the-art semantic segmentation on a domain that is underrepresented in large pre-training datasets. 

The Coralscapes dataset is:
\begin{itemize}
    \item The first general-purpose semantic segmentation dataset in the coral reef domain, including a wide range of reefs from thriving to heavily bleached/dead, as well as diverse camera angles and distances to the substrate. Unlike previous datasets, Coralscapes is not restricted to close-ups of corals, orthomosaics, or photo quadrats.
    \item The largest expert-annotated semantic segmentation dataset spanning 2075 images at 1024$\times$2048px resolution, containing 174k polygons over 39 classes, labeled in a consistent and speculation-free manner.
    \item The first dataset in the coral reef domain designed for fair evaluation of machine learning approaches, enforcing a spatial train/test split on images gathered from 35 dive sites in 5 countries in the Red Sea.
\end{itemize}

%\begin{enumerate}
 %   \item Also less present in pre-training datasets than urban scenes
%    \item Serves two purposes: benchmark semantic segmentation models in domain outside of `everyday domain` (outdoors or indoor scenes). COCO stuff has no underwater classes. Second purpose is to drive the applications for conservation technology
    
%\end{enumerate}

\begin{table*}[t!]
    \centering
    \resizebox{0.99\textwidth}{!}{\begin{tabular}{l|cccccccccc}
    \toprule
        Dataset & \# Images & Annotation Type & \# Annotations & \makecell{\# Annotated\\ Pixels} &  \# Classes & Comment & \makecell{Spatial\\Split} & \makecell{Openly\\ Available}\\
        \hline
        RSMAS \cite{rsmas} & 776 & Image & 776 & - &14 &  - & \red{No} & \green{Yes} \\
        Jamil et al. \cite{bof} & 1582 & Image & 1582 & - & 3 &  - & \red{No} & \green{Yes} \\
        Raphael et al. \cite{raphael2020deep} & 5000 & Image  & 5000 & - & 11 & Coral only & \red{No} & \red{No} \\  
        DeOhloNosCorais \cite{furtado} & 1411 & Image + Foreground Map & 1411 & 7.08M & 21 & - & \red{No} & \green{Yes} \\
        Eilat \cite{beijbom2016improving} & 212 & Sparse & 1123 & - & 8 & & \red{No} & \green{Yes} \\
        King et al. \cite{king2018comparison} &  1807 & Sparse & 9511 & - & 10 &  - & \red{No} & \red{No} \\
        Moorea Labeled Corals \cite{beijbom2012automated} & 2055 & Sparse & 400,000 & - & 9 & Only photo quadrats & \red{No} & \green{Yes} \\
        Pacific Labeled Corals \cite{beijbom2015towards} &  5090 & Sparse & 251,988 & - & 20 & Only photo quadrats & \red{No} & \green{Yes} \\
        Benthoz-2015 \cite{benthoz15} & 9874 & Sparse & 407,968 & - & 147 & Ortho-photos & \green{Yes} & \green{Yes} \\
        Seaview \cite{seaview} & 11387 & Sparse & 859,870 & - & 228 & Ortho-photos & \green{Yes} & \green{Yes} \\
        BenthicNet \cite{benthicnet} * & 29603 & Sparse & 1,390,262 & - & 262 & *Subset of coral sites & \green{Yes} & \green{Yes} \\
        Benthos \cite{benthos} & 4 & Semantic Segmentation & 4500 & Unknown & 8 & Orthomosaics & \red{No} & \green{Yes} \\
        UCSD Mosaics \cite{edwards} & 16 & Semantic Segmentation & 54055 & 1.29B &35 & Orthomosaics & \red{No} & \red{No} \\       
        CoralSCOP \cite{zheng2024coralscop} & 41297 & Segmentation Masks$^\dagger$ & 330,144 & 17.56B & 136 & $^\dagger$Coral masks only & \red{No} & Request \\
        \midrule
        Coralscapes (Ours) & 2075 & Semantic Segmentation & 174,077 & 3.36B & 39 & General purpose & \green{Yes} & \green{Yes}\\
        \bottomrule
    \end{tabular}}
    \caption{Comparison of existing datasets for computer vision in coral reefs in terms of annotation type, size, classes, availability of disjoint geographic locations allowing a spatially disjoint train/test split, and open availability. \vspace{4pt} }
    \label{tab:other_datasets}
\end{table*}

%-------------------------------------------------------------------------
\section{Related Work}

\subsection{Large Datasets for Semantic Segmentation}

Large benchmark datasets have played a key role in deep learning's rapid advancement in semantic segmentation many other computer vision tasks. Early datasets such as PASCAL VOC \cite{pascalvoc} provided pixel-wise annotations for object categories, even before deep learning was ubiquitous. The introduction of Cityscapes \cite{cityscapes}, a dataset of 5000 images at 1024$\times$2048px with fine-grained semantic annotations in urban driving scenarios, set a precedent for large-scale semantic segmentation datasets. Notably, Cityscapes became the standard benchmark for evaluating segmentation networks, influencing the design of widely used neural network architectures such as the DeepLab \cite{deeplab} family of models \cite{deeplabv3,deeplabv3plus} and Pyramid Scene Parsing Networks \cite{pspnet}.

Following the success of Cityscapes, several semantic segmentation datasets have been developed: ADE20K \cite{ade20k} includes a broad range of everyday environments, including street scenes, on many more labeled images than Cityscapes. Similarly, Mapillary Vistas \cite{mapillary} extended both the scale of urban scene segmentation and its scope by introducing images captured under diverse lighting, weather, and geographical conditions. Datasets like COCO-Stuff \cite{cocostuff} and LVIS \cite{LVIS} explore object segmentation in a more general variety of contexts.

A multitude of semantic segmentation datasets in domains that commonly require experts have been proposed, such as the BRATS \cite{brats}, LITS \cite{LITS} and ACDC \cite{ACDC} datasets in the medical imaging domain, and the DeepGlobe \cite{deepglobe}, Potsdam \cite{potsdam}, and Vaihingen \cite{vaihingen} datasets in remote sensing. Expert-domain datasets are significantly smaller than general-purpose semantic segmentation datasets due to the limited availability of expert time for annotation.

\subsection{Coral Reef Classification \& Segmentation}

\begin{figure*}
    \centering
    \includegraphics[width=0.99\textwidth]{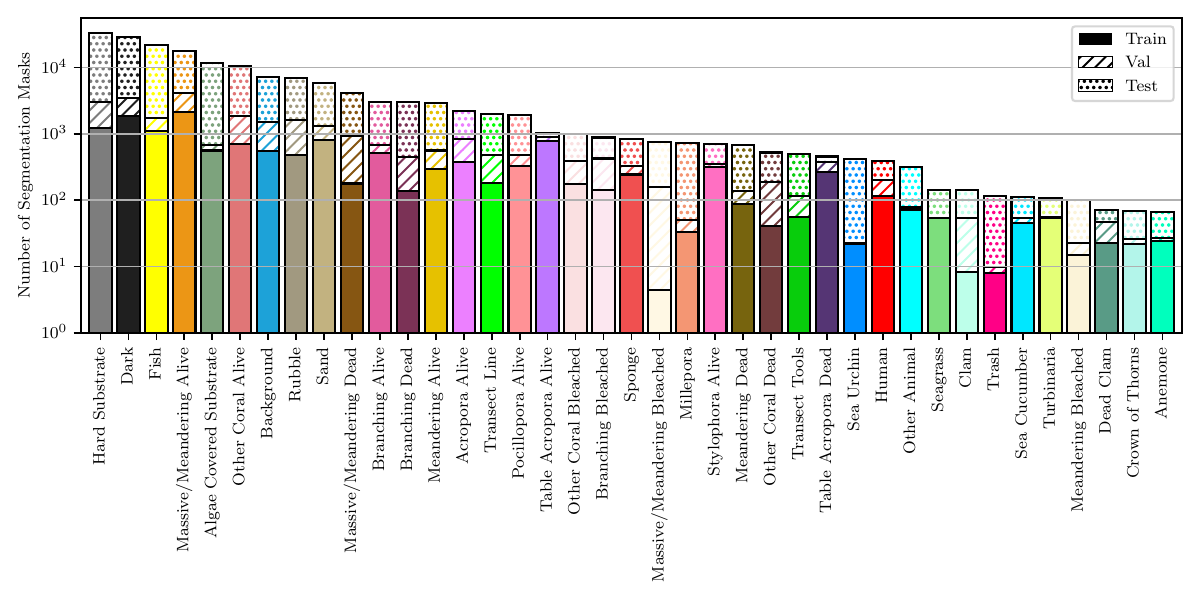}
    \caption{Number of annotated segmentation masks per class in the Coralscapes dataset splits for each of the 39 classes (shown with linear proportions on logarithmic scale).}
    \label{fig:polygon_counts}
\end{figure*}

In the context of coral reefs, %the creation of large datasets is impeded by the annotation effort requiring expert time. 
existing expert-labeled datasets provide either only image-level labels \cite{rsmas, bof, raphael2020deep, furtado} on close-up images of corals, a few sparse labeled pixels on entire images \cite{beijbom2016improving,king2018comparison,beijbom2012automated,beijbom2015towards}, or focus on segmentation in restricted domains such as photo quadrats or orthomosaics \cite{benthos,edwards}. Table~\ref{tab:other_datasets} provides a comparison of existing datasets, including the BenthicNet \cite{benthicnet} dataset (subselected for coral images, i.e. all sites with at least one hard coral label and shallower than 60m), and CoralSCOP \cite{zheng2024coralscop}, which provides a model similar to Segment Anything \cite{sam} trained on annotated masks of corals (but no other benthic classes). 

Most existing datasets do not split the training, validation, and testing data according to different geographic sites, leading to a likely overestimation of the classification performance compared to real-world applications in reef sites outside the training data. Furthermore, the larger, but sparsely annotated datasets \cite{benthicnet,benthoz15, seaview} are essentially crowd-sourced from various annotation teams, leading to a large and noisy label set, that is challenging to transfer to new biogeographic regions. Lastly, except for the SUIM dataset \cite{suim}, which provides dense segmentation labels focusing on robot-human interaction on diving missions and has no dedicated classes for live corals, no existing segmentation datasets are \textit{general-purpose} in the sense of diverse camera angles, distances to the substrate, environmental conditions (water turbidity, color, lighting), and with all relevant classes visible on the images annotated.

The absence of a high-quality general purpose segmentation dataset means that widely used machine learning applications in coral reefs are restricted to point- or patch-wise classification \cite{beijbom2012automated,coralnetengine,mermaid,reefcloud,hopkinson2020automated}, and raised interest in automatic expansion of sparse labels into dense masks \cite{alonso2019coralseg, benthos, raine2022point}. The Coralscapes dataset aims to fill this gap, providing a dataset built for dense segmentation in general purpose reef scenarios.

\FloatBarrier
\section{Dataset}

\subsection{Data Collection}

All imagery was collected during scuba dives at 35 sites in coral reefs of Djibouti, Eritrea, Sudan, Jordan, and Israel using GoPro Hero 10 cameras. The annotated images are video frames originally taken at 1080 $\times$ 1920px resolution and 30 frames per second in the linear lens setting. The videos were taken in a diverse set of reefs: from thriving reefs very high live coral cover to devastated environments with severe bleaching and coral mortality, from high visibility in good lighting conditions to turbid scenes in low-light settings. As the videos were acquired during transect surveying campaigns of the Transnational Red Sea Center hosted at EPFL, Coralscapes includes images including transect tools, transect lines, divers, and some human-made structures commonly encountered in reef monitoring settings. Additional information about the data collection (location, depth, reef types) is found in Appendix~\ref{appendix:additional_data}. To protect the coral reefs in Coralscapes from overtourism or illegal poaching activities, and in accordance with local research permit authorities, image sites are withheld and instead replaced by a site ID.

\subsection{Annotation}

The main challenge in creating a high-quality dataset for coral reef segmentation lies in the complexity of the annotations. Besides the visual degradation induced by the water column, many benthic classes are difficult to differentiate even in good conditions and by experts: classifying corals to a high taxonomic level from images is complex because of their strong morphological plasticity \cite{plasticity, alonso2019coralseg}. Although some genera can be identified by visually distinct features, many genera can simply not be discerned from casual imagery, even by domain experts familiar with the biogeographic area. In such cases, the growth form of the coral (e.g. branching, massive) is commonly reported, as this determines the main ecological function of the coral within the reef ecosystem. Live corals are not the only source of ambiguity: fine algal mats only become apparent when viewed closely and are indistinguishable from non-algae-covered substrates at a larger distance. Additionally, when hard corals die, their calcium carbonate skeletons start degrading over time, and transform into hard substrate that may become itself covered in algae or break down into rubble. As a result, these substrates often fit multiple label classes depending on their stage of degradation.
In spite of the inherent ambiguities, these classes remain of real interest to coral surveyors that monitor how a reef's state changes with time, and should be identified by an automatic tool. At the same time, there should be no \textit{speculative} assignment of polygons to classes that may be incorrect.
	
To align to all those requirements, the annotation was done in a speculation-free manner, enforcing a conservative semantic segmentation in terms of taxonomic depth. For example, when a branching coral was too far away from the camera to decide with certainty whether it is alive or dead, it was labeled as `background'. Viewed closer in another video frame, the same coral may be assigned to its growth form (branching, massive, etc.) and whether it is alive or dead. A precise genus label was assigned only when the genus was clearly distinguishable. In-depth details of the annotated classes is given in Appendix~\ref{appendix:class_guide}.

\begin{figure}
    \centering
    \includegraphics[width=\columnwidth]{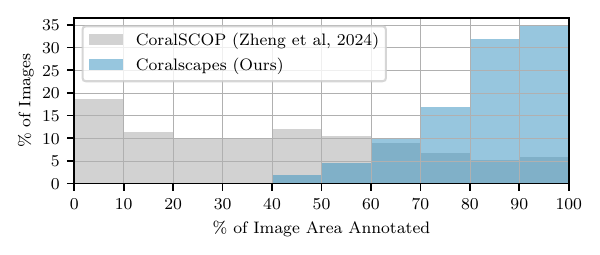}
    \vspace{-20pt}
    \includegraphics[width=\columnwidth]{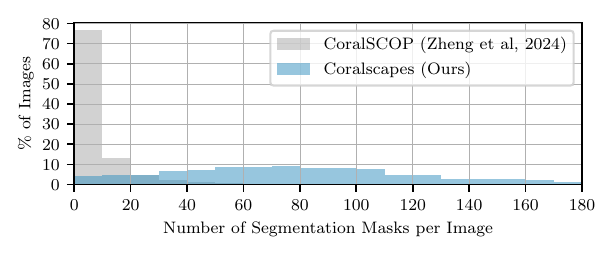}
    \vspace{-8pt}
    \caption{Histogram of the image area covered by annotations (top) and of the number of polygons present per image (bottom), highlighting the complexity of semantic segmentation in Coralscapes, compared to the CoralSCOP dataset~\cite{zheng2024coralscop}. Further comparison to the CoralSCOP dataset can be found in Appendix~\ref{appendix:coralscop}.}\vspace{-10pt}
    \label{fig:comparison_histogram}
\end{figure}

A set of 39 classes was chosen to cover the most prominent and most important benthic classes present in the images (Figure~\ref{fig:polygon_counts}). This label set includes 10 visually distinct live coral classes, 5 dead coral classes, 4 bleached coral classes, 6 non-coral invertebrates, 4 substrate classes, and classes `human', `transect line', `transect tools, `fish', and `trash'. The `dark' and `background' classes are assigned to pixels that can not be classified due to either poor lighting or being too far from the camera (and thus blurred or colorless).  While the exact delineation of `dark' and `background' is subjective, their inclusion is necessary to reduce speculative labels.

Annotation of the selected 39 classes with segmentation masks was done by four experts on coral reefs who were also present during data collection, and six additional trained annotators. The specific image frames for annotation were selected from over 200 hours of video material in order to capture the diversity of the scenes, while obtaining a good coverage of the 39 classes. 
The CVAT annotation interface \cite{cvat} was used, which allowed annotators to interact with the Segment Anything Model \cite{sam}. While this helped to segment some objects with salient boundaries, it proved to be largely unhelpful for classes with less apparent boundaries and those appearing as small polygons, where polygons where drawn by hand. The annotation process emphasized high-quality annotations, meaning that there was a focus on annotating each frame with many polygons and on covering much of the image area, instead of annotating many images sparsely, as shown in Fig.~\ref{fig:comparison_histogram}. All polygons were visually verified to be correctly delineated and assigned the correct class.  

\begin{figure}
    \centering
    \includegraphics[width=\columnwidth]{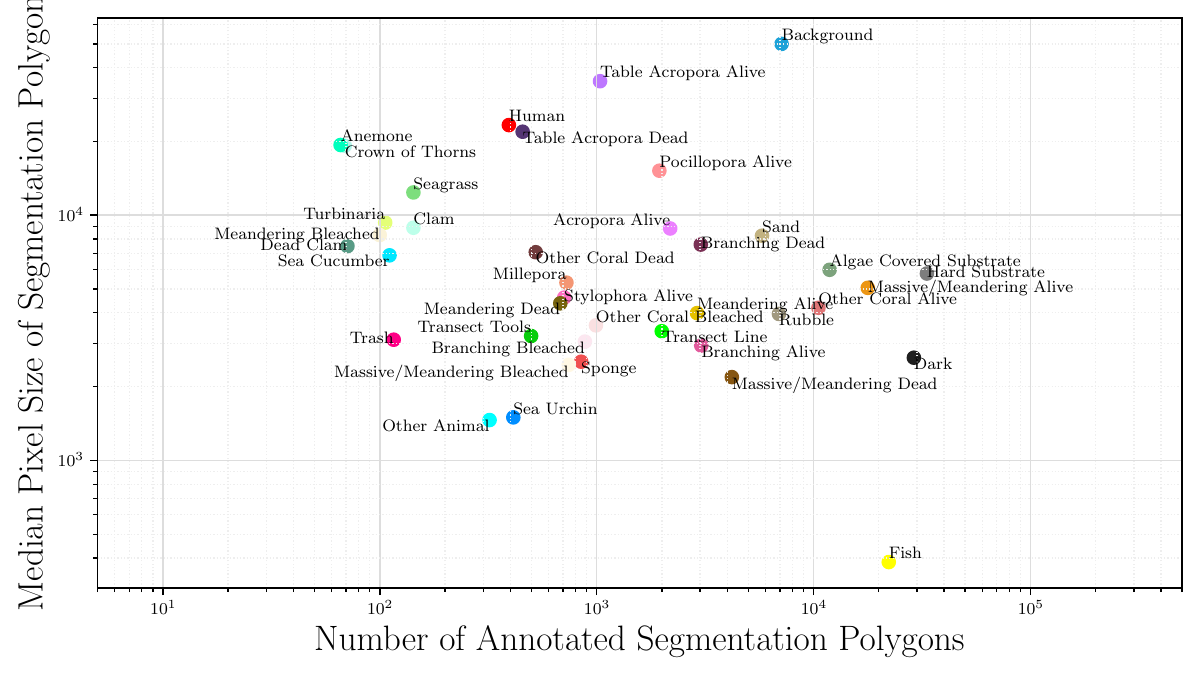}
    \caption{Median size (in pixels) of an annotated polygon for the classes of Coralscapes, plotted against the number of annotated polygons. This highlights the challenge of segmenting classes that require the global image structure to segment correctly, as well as small fine-grained classes in the same dataset.}
    \label{fig:median_dist}
\end{figure}

In terms of class presence, the annotations of Coralscapes capture the strong imbalance in reef scenes: despite making an effort to prioritize the selection of frames with rare classes, the most common class has over 400 times more annotated polygons than the rarest class, as shown in Figure~\ref{fig:polygon_counts}. Similarly, the granularity and size of the visible benthic classes varies dramatically: even though a class may appear very often, its total number of annotated pixels may still be low because it often appears in small polygons, and vice versa, as highlighted in Figure~\ref{fig:median_dist}. For example, the median size of a `fish' polygon is more than 120 times smaller than the median `background' polygon. This imbalance in terms of representation and scale contributes to making Coralscapes a challenging benchmark for semantic segmentation. Additional statistics of Coralscapes (pixel counts, class frequency at image level) are provided in Appendix~\ref{appendix:additional_data}.

\subsection{Dataset Structure}

We split the dataset spatially by reef site to allow a fair evaluation, resulting in a training set of 1517 images (27 sites), a validation set of 166 images (3 sites), and a test set of 392 images (5 sites). These splits were selected so that the classes are well represented in all three splits, which is shown in the proportions in Figure~\ref{fig:polygon_counts} (`Turbinaria' and `Seagrass' are only present in two sites each, so they are omitted from the validation set). 

To facilitate the usability of Coralscapes, the structure of the dataset mimics that of the widely used Cityscapes benchmark: images are provided as 8-bit PNG images resized to 1024$\times$2048px resolution, and the 19 preceding and 10 trailing video frames are available for download (at 30 FPS). We make Coralscapes easily available: frames and their segmentation masks are available on Huggingface\footnote{\tiny{\url{https://huggingface.co/datasets/EPFL-ECEO/coralscapes}}}, from where they can be loaded to Python in one line, and on Zenodo\footnote{\tiny{\url{https://zenodo.org/records/15061505}}}, where we also provide the preceding and trailing frames, the frames at original 1080$\times$1920px resolution, and benchmarked model checkpoints.

\section{Benchmarking}

\begin{table}[b!]
    \centering
    \resizebox{\columnwidth}{!}{
    \begin{tabular}{l|cccccccccc}
    \toprule
    Method & Test Accuracy  & Test mIoU \\ 
    \midrule
    UNet++ - ResNet50 & 75.593 & 42.906 \\
    DeepLabV3+ - ResNet50 & 78.171 & 45.720 \\
    SegFormer - MiT-b2  & 80.904 & {54.682} \\
    SegFormer - MiT-b2 (LoRA) & 80.987 & 51.911 \\
    SegFormer - MiT-b5 & 80.939 & \underline{55.031} \\
    SegFormer - MiT-b5 (LoRA) & 80.863 & 52.775 \\
    Linear - DINOv2-Base & {81.339} & 52.478 \\
    DPT - DINOv2-Base  & 78.124 & 44.203 \\
    DPT - DINOv2-Base (LoRA)  & 80.053 & 47.233 \\
    DPT - DINOv2-Giant & 80.691 & 50.643 \\
    DPT - DINOv2-Giant (LoRA)  & \underline{81.701} & 54.531 \\
    \midrule
    SegFormer - MiT-b5 (Stronger Augmentations) & \textbf{82.761 }& \textbf{57.800}  \\

    \bottomrule
   \end{tabular}}
    \caption{Quantitative results on Coralscapes segmentation, best shown in \textbf{bold}, second best \underline{underlined}.}
    \label{tab:benchmarking}
\end{table}

\begin{figure*}
    \centering
    \includegraphics[width=\linewidth]{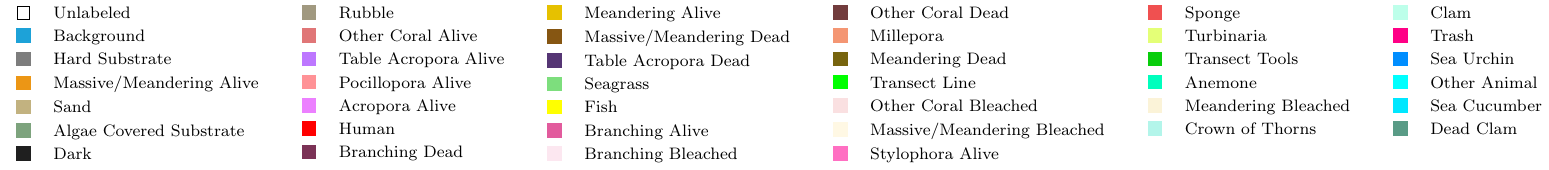}
    \begin{subfigure}{0.195\textwidth}
        \centering
        \includegraphics[width=\linewidth]{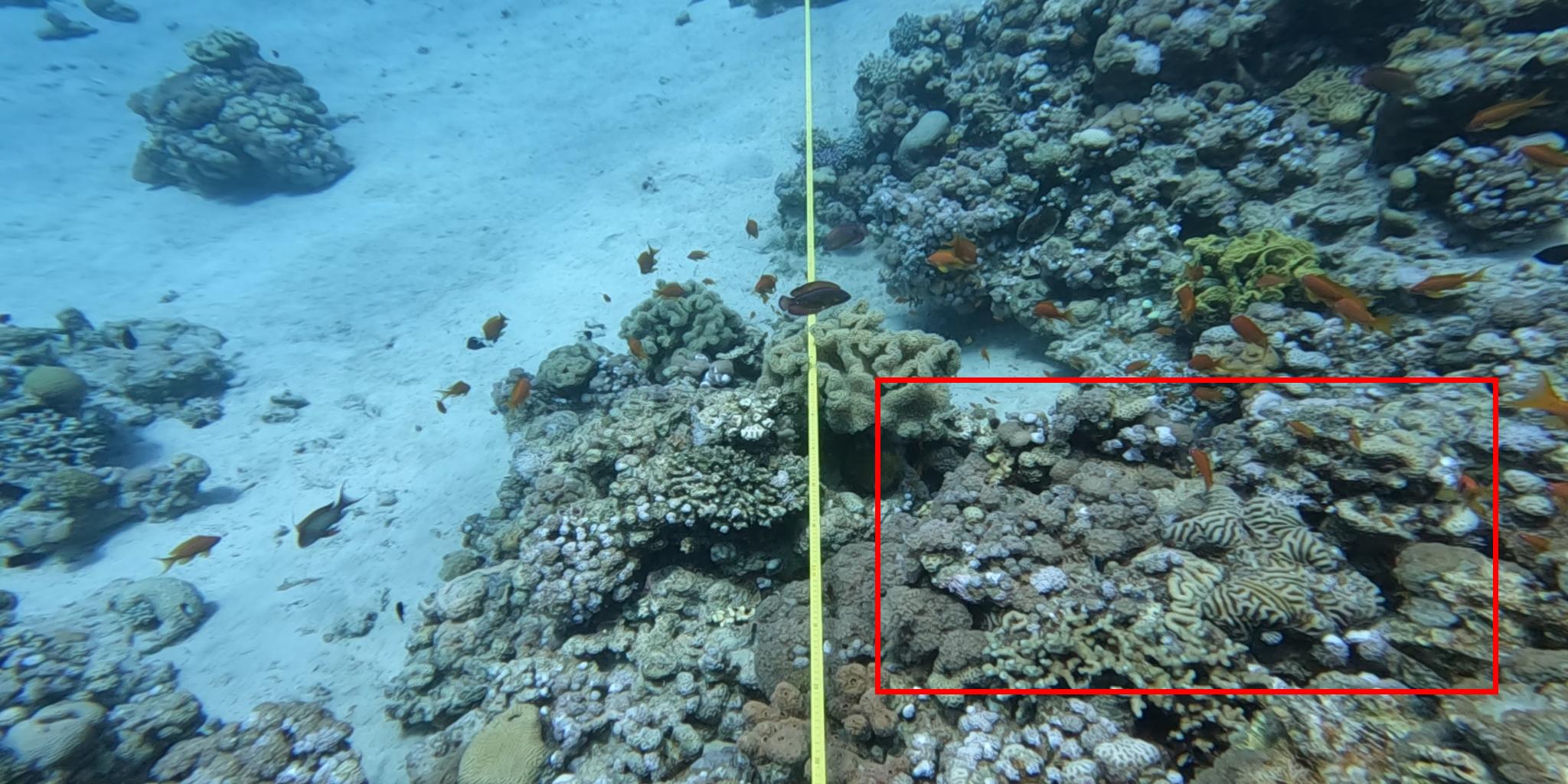} 
        \includegraphics[width=\linewidth]{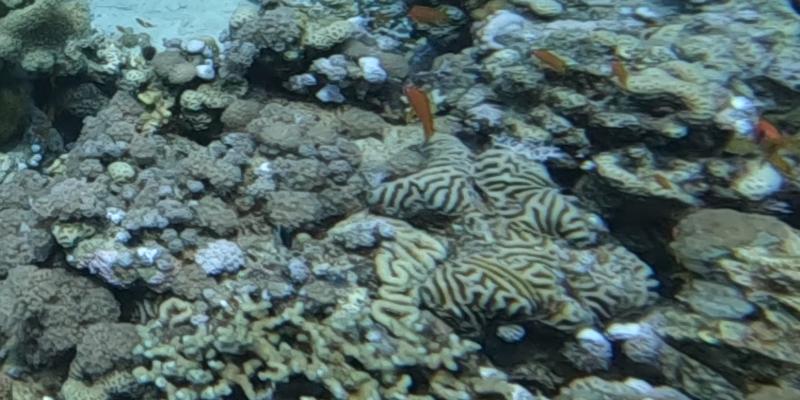}
        \includegraphics[width=\linewidth]{figures/benchmarking_examples/0167image.rgb.jpg} 
        \includegraphics[width=\linewidth]{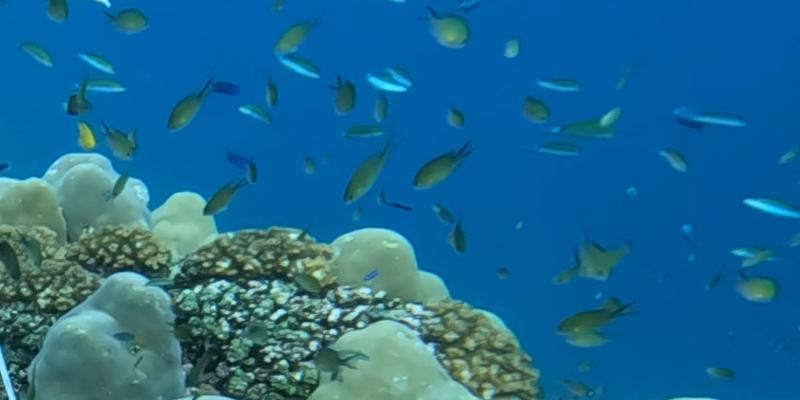} 
        \includegraphics[width=\linewidth]{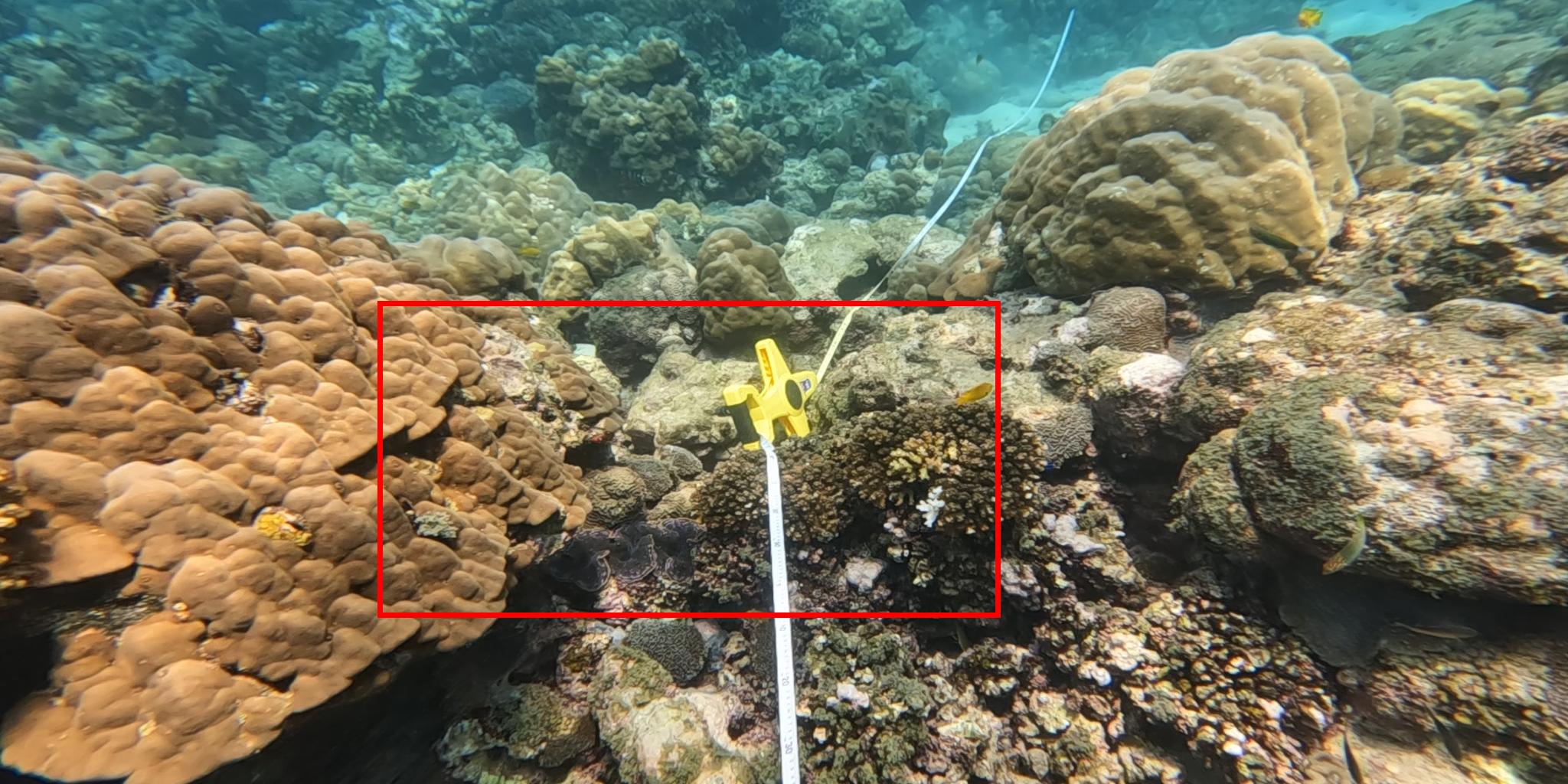} 
        \includegraphics[width=\linewidth]{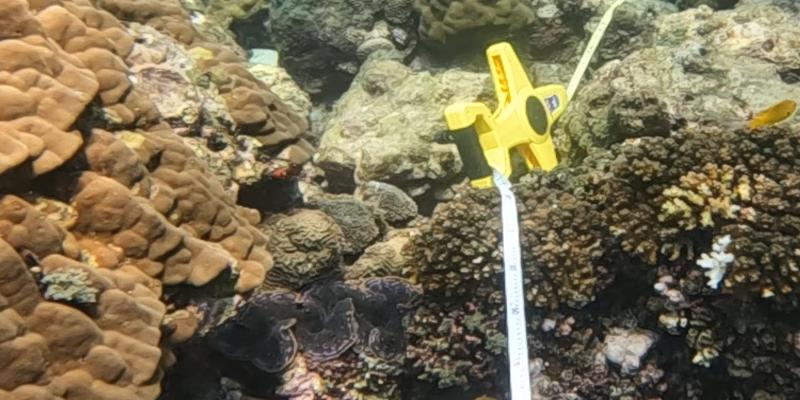} 
        \includegraphics[width=\linewidth]{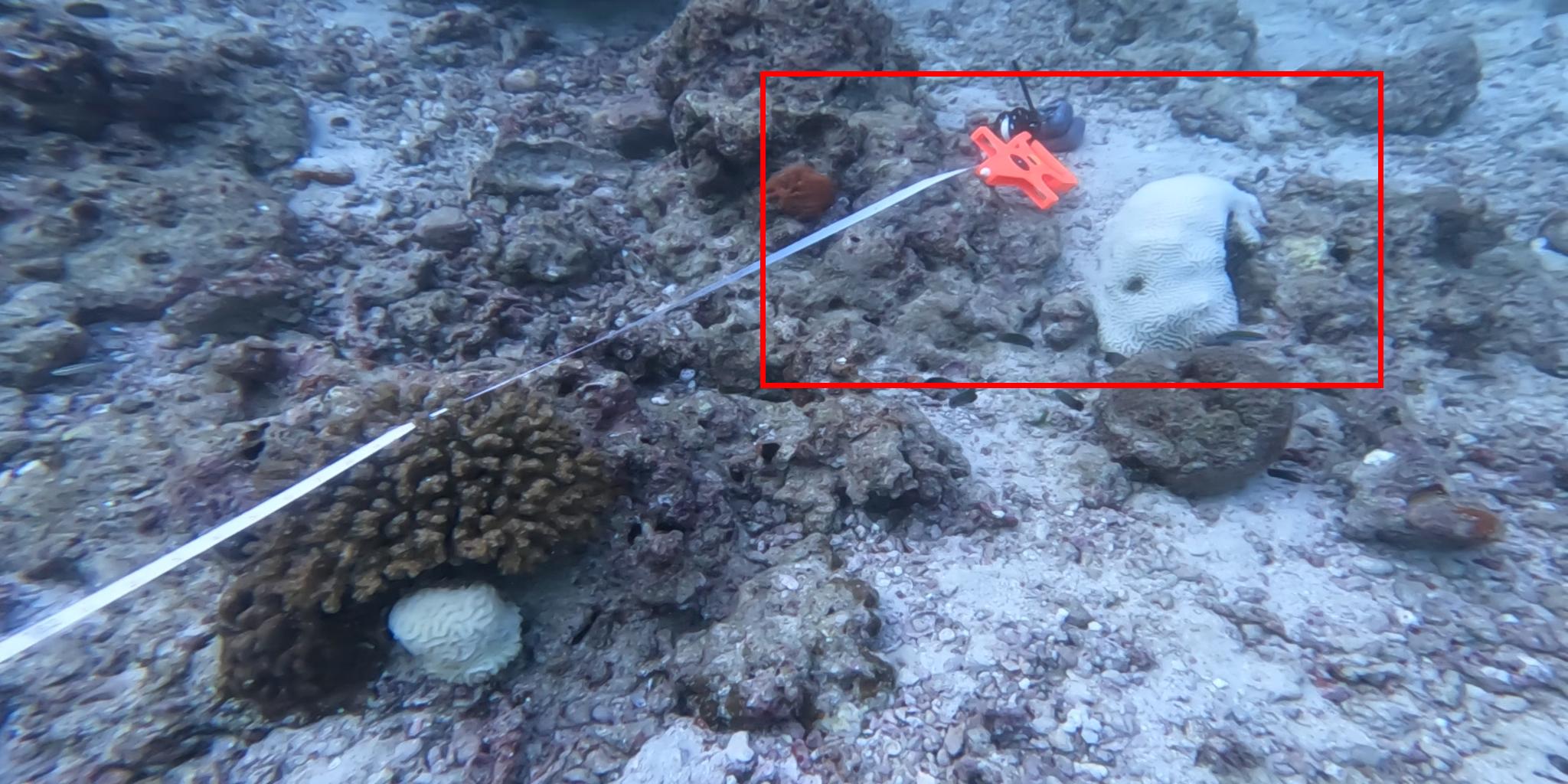} 
        \includegraphics[width=\linewidth]{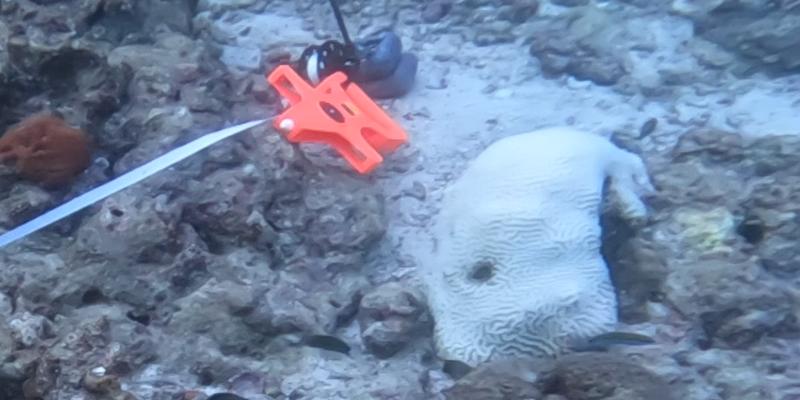} 
        \caption{\\Image}
        \label{fig:sub1}
    \end{subfigure}
    \begin{subfigure}{0.195\textwidth}
        \centering
        \includegraphics[width=\linewidth]{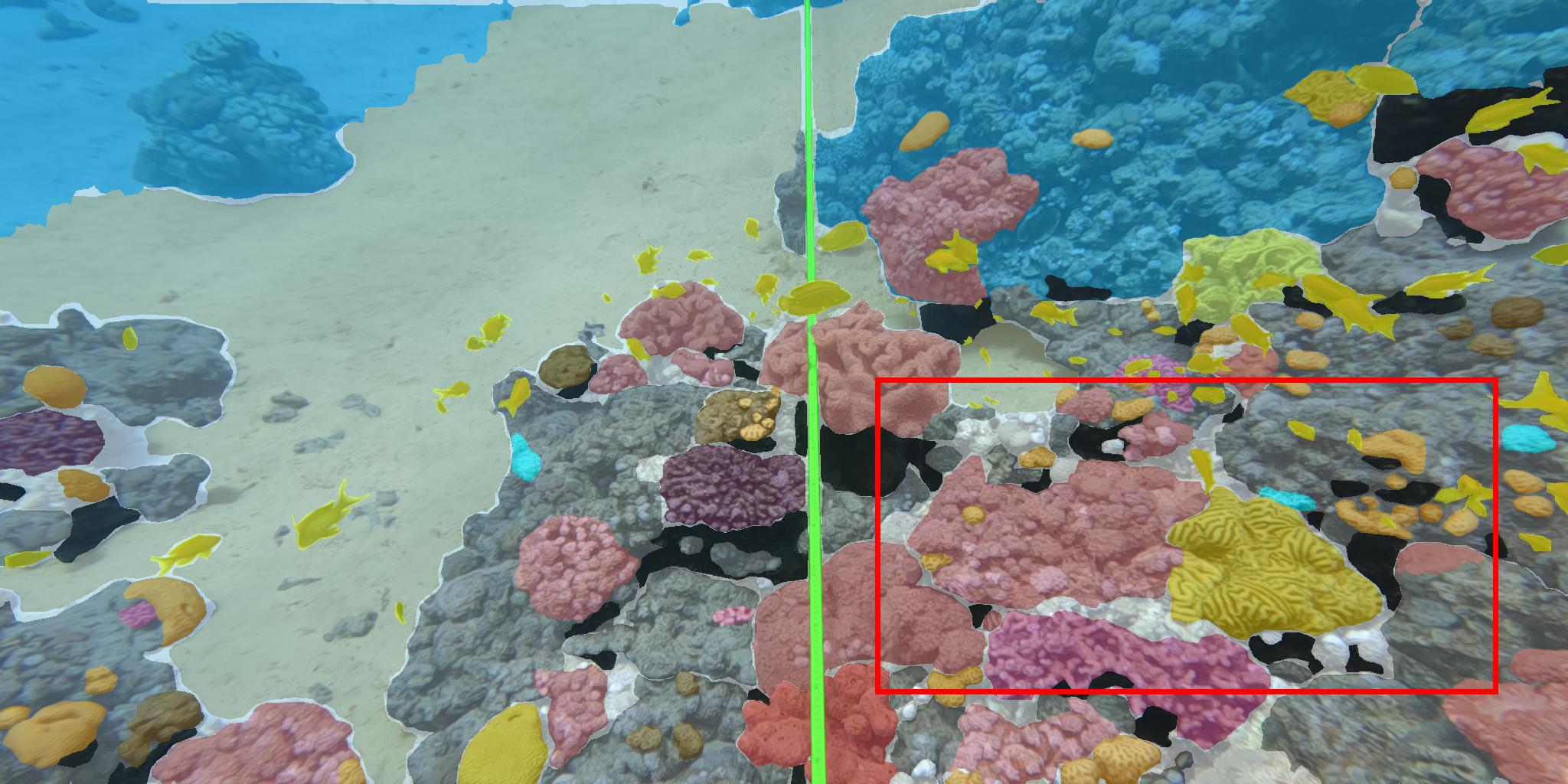} 
        \includegraphics[width=\linewidth]{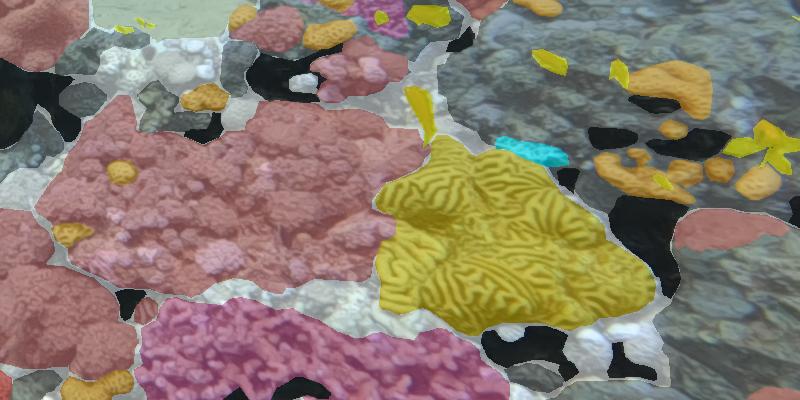} 
        \includegraphics[width=\linewidth]{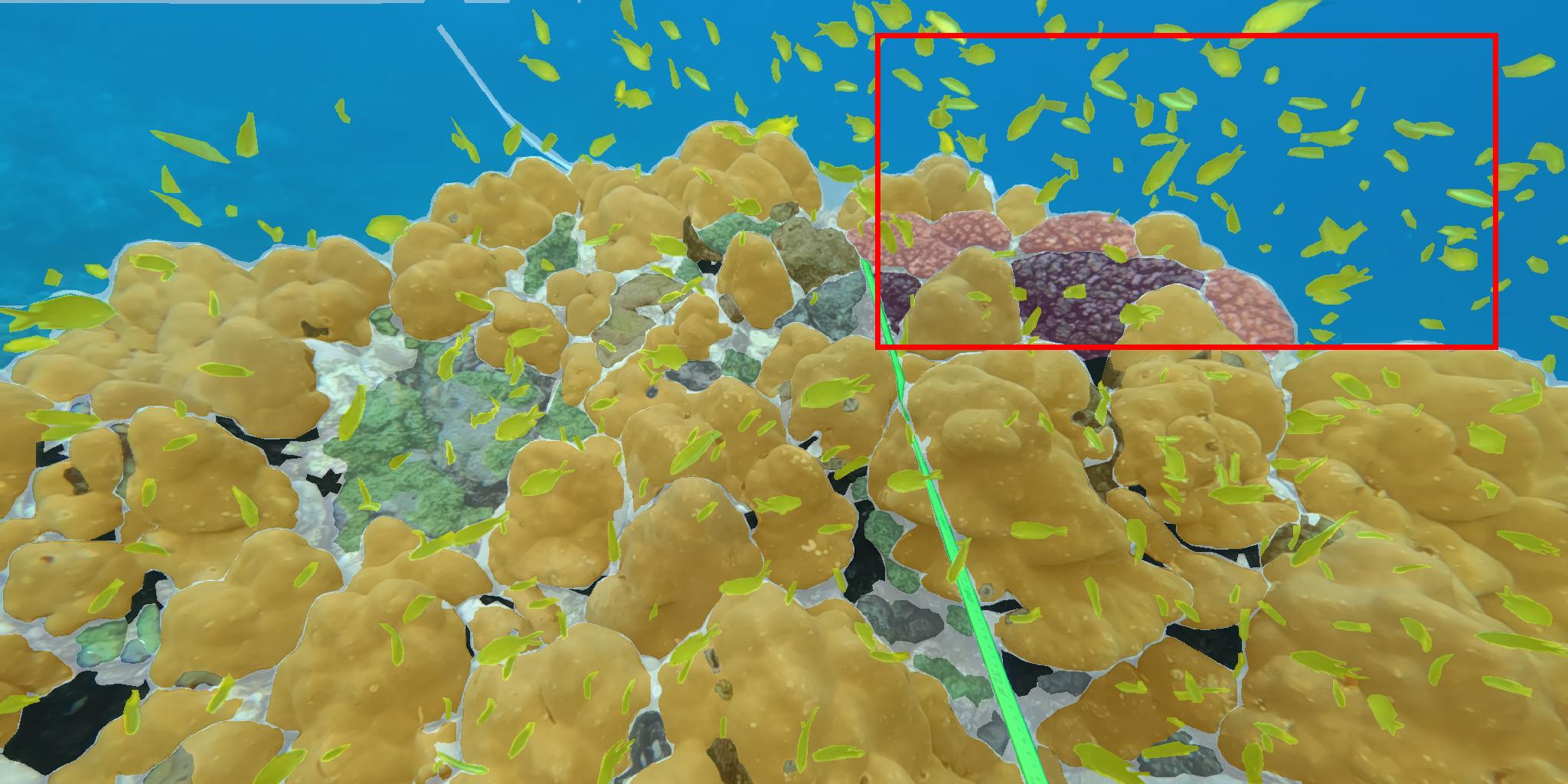} 
        \includegraphics[width=\linewidth]{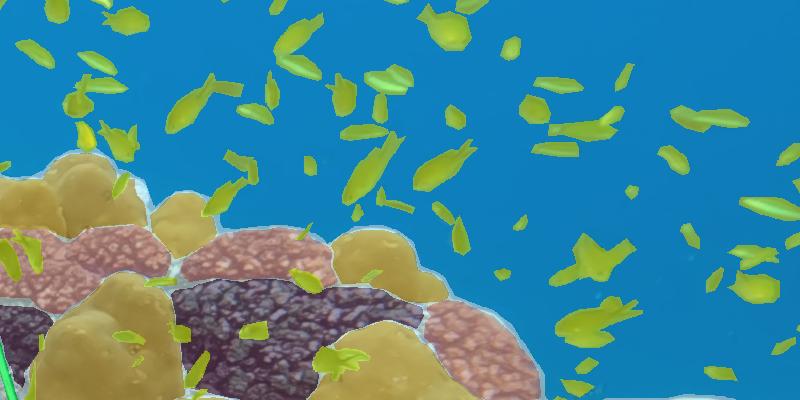}
        \includegraphics[width=\linewidth]{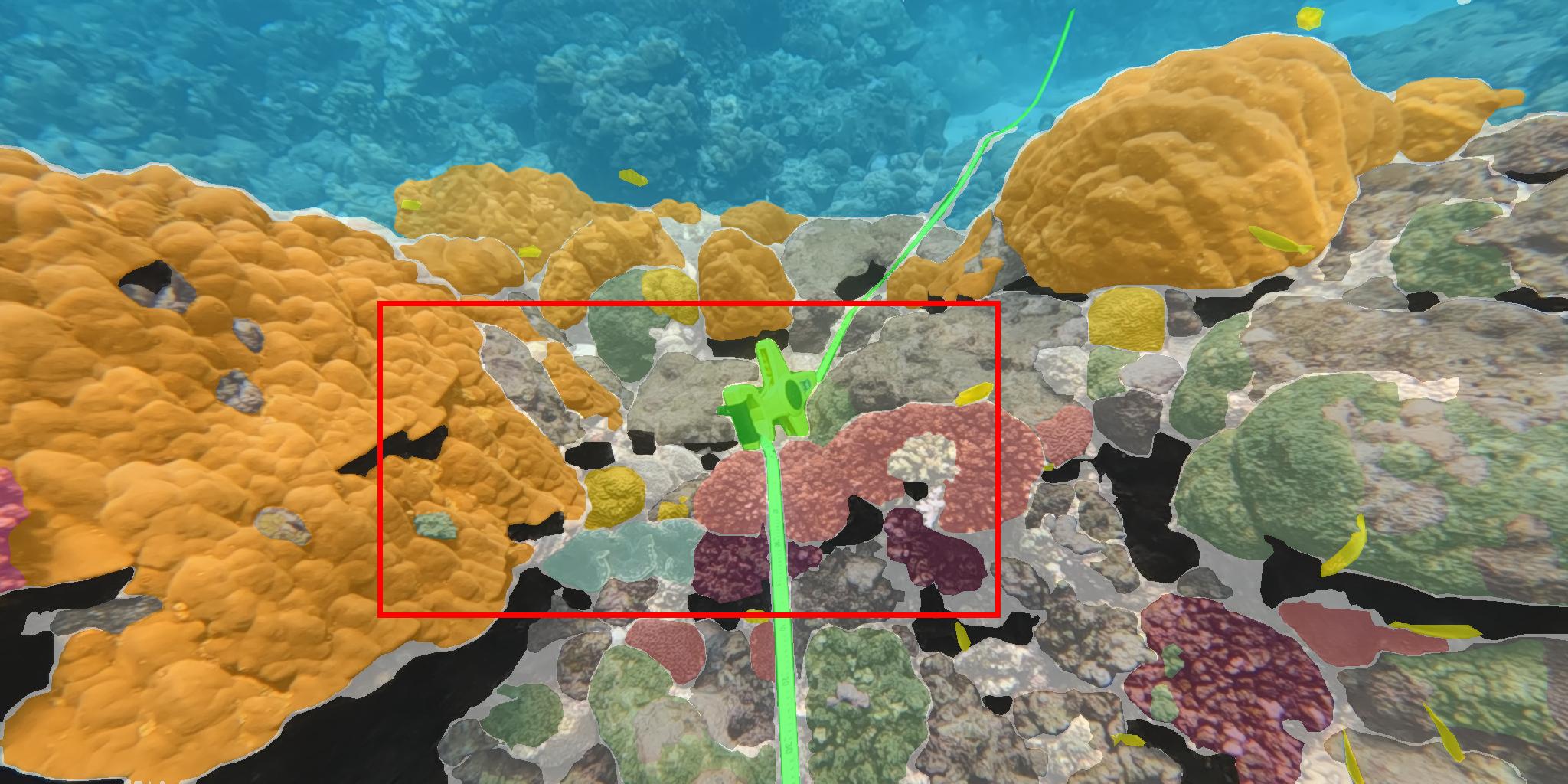} 
        \includegraphics[width=\linewidth]{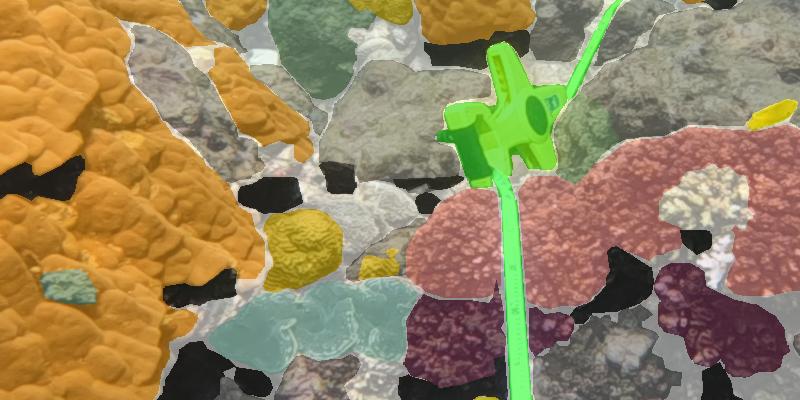} 
        \includegraphics[width=\linewidth]{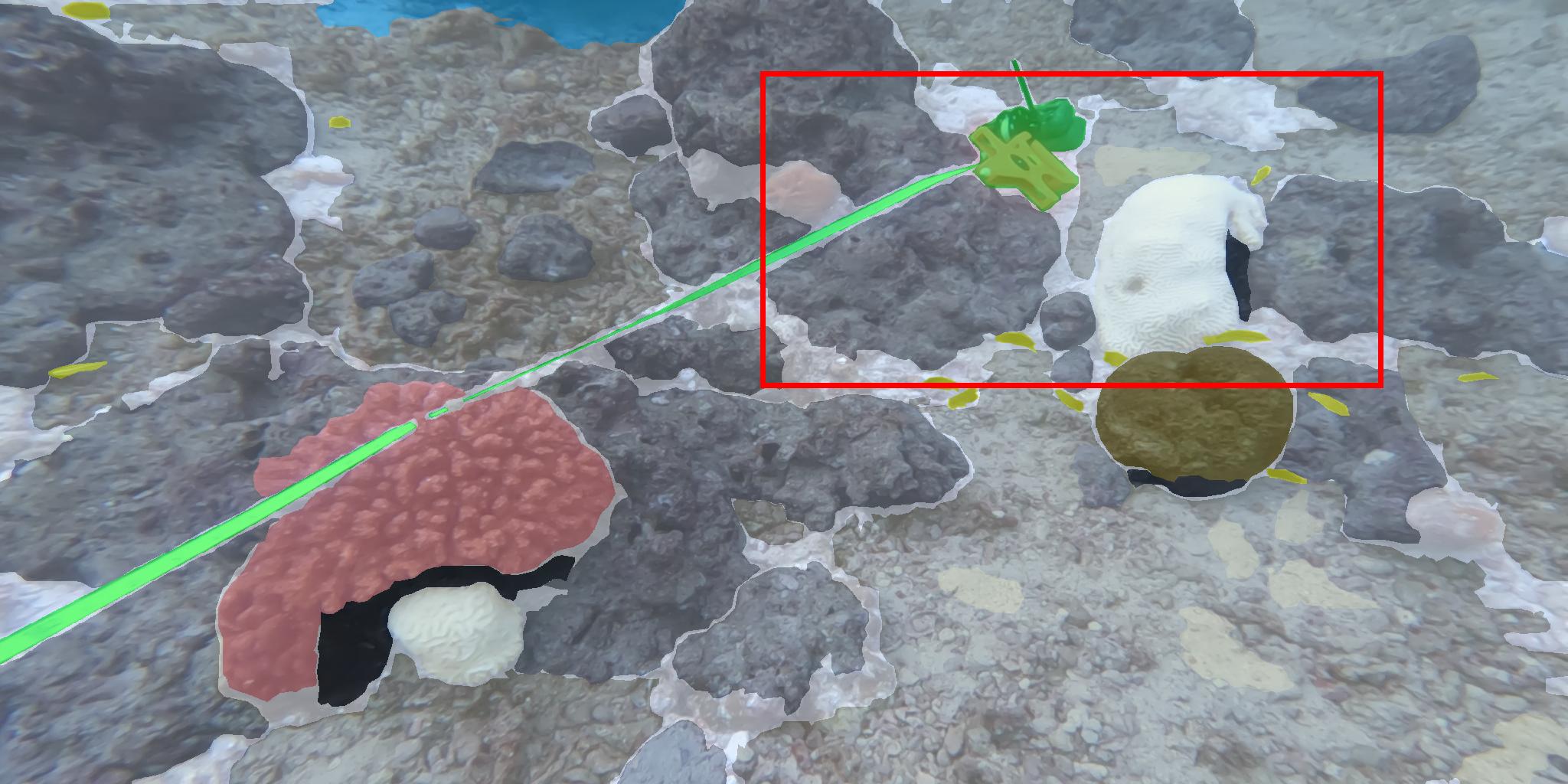} 
        \includegraphics[width=\linewidth]{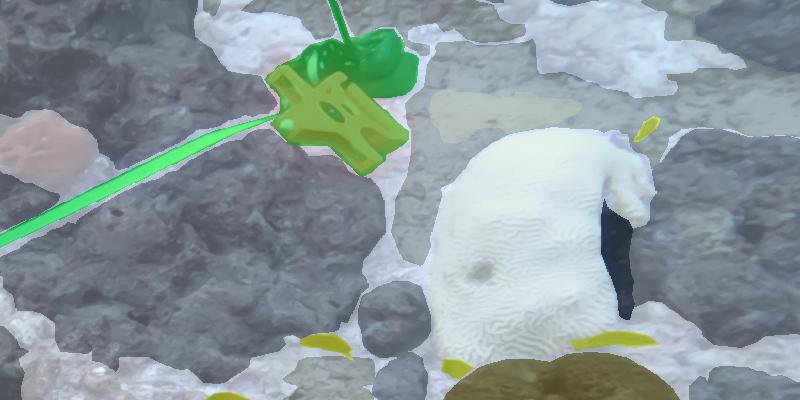}         
        \caption{\\Ground Truth}
        \label{fig:sub2}
    \end{subfigure}
    \begin{subfigure}{0.195\textwidth}
        \centering
        \includegraphics[width=\linewidth]{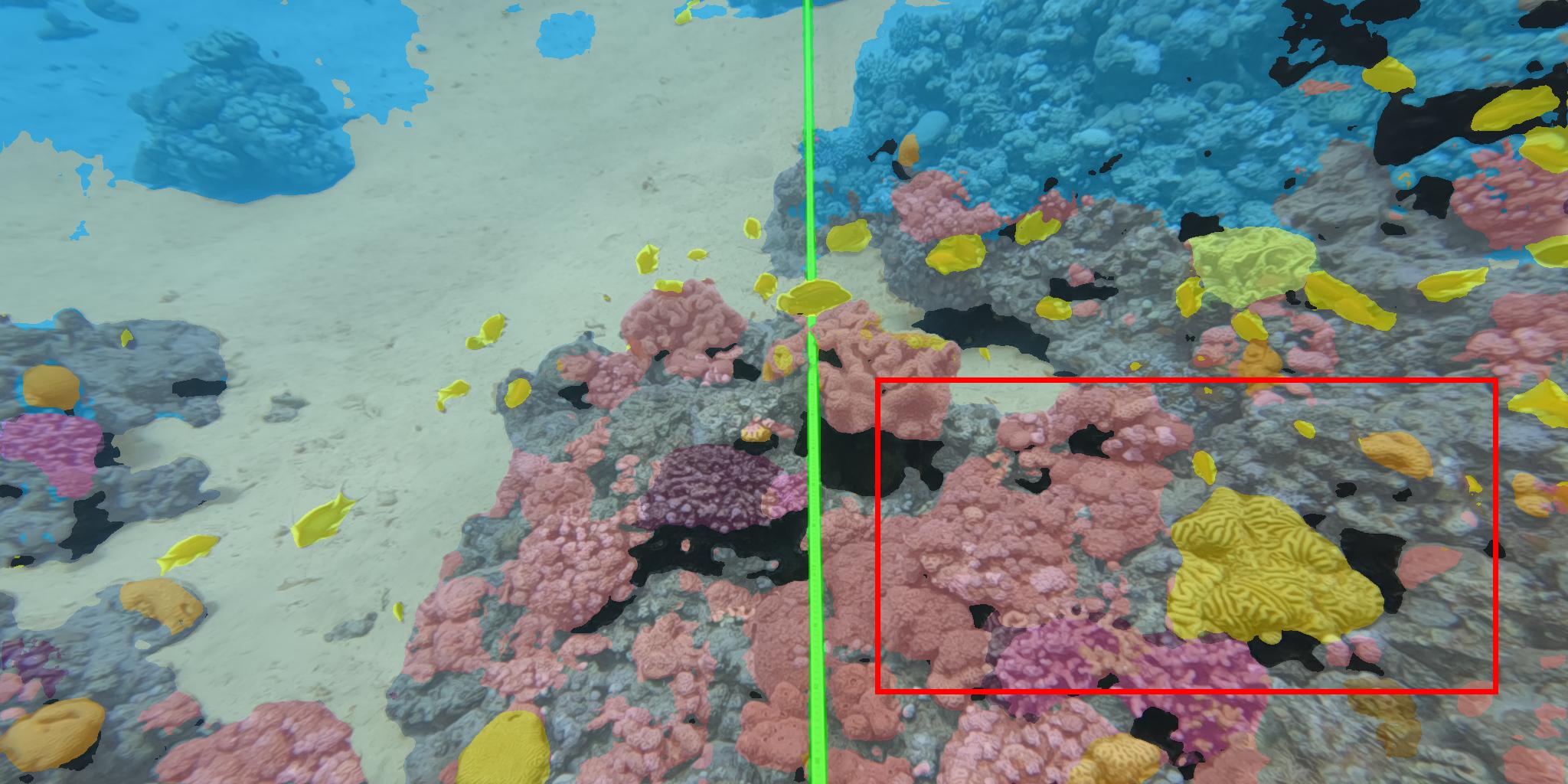} 
        \includegraphics[width=\linewidth]{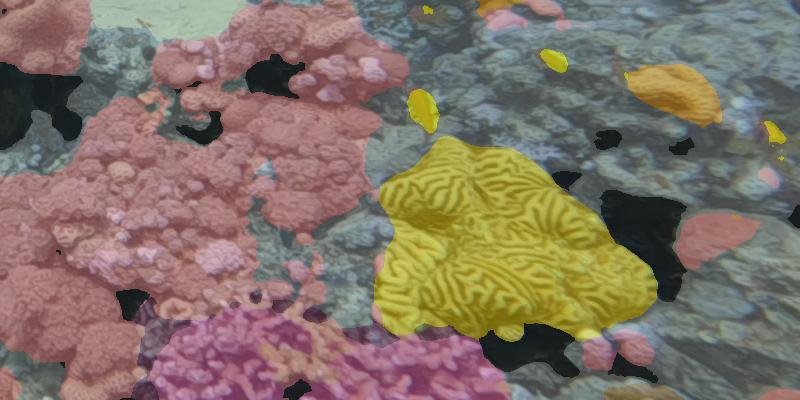} 
        \includegraphics[width=\linewidth]{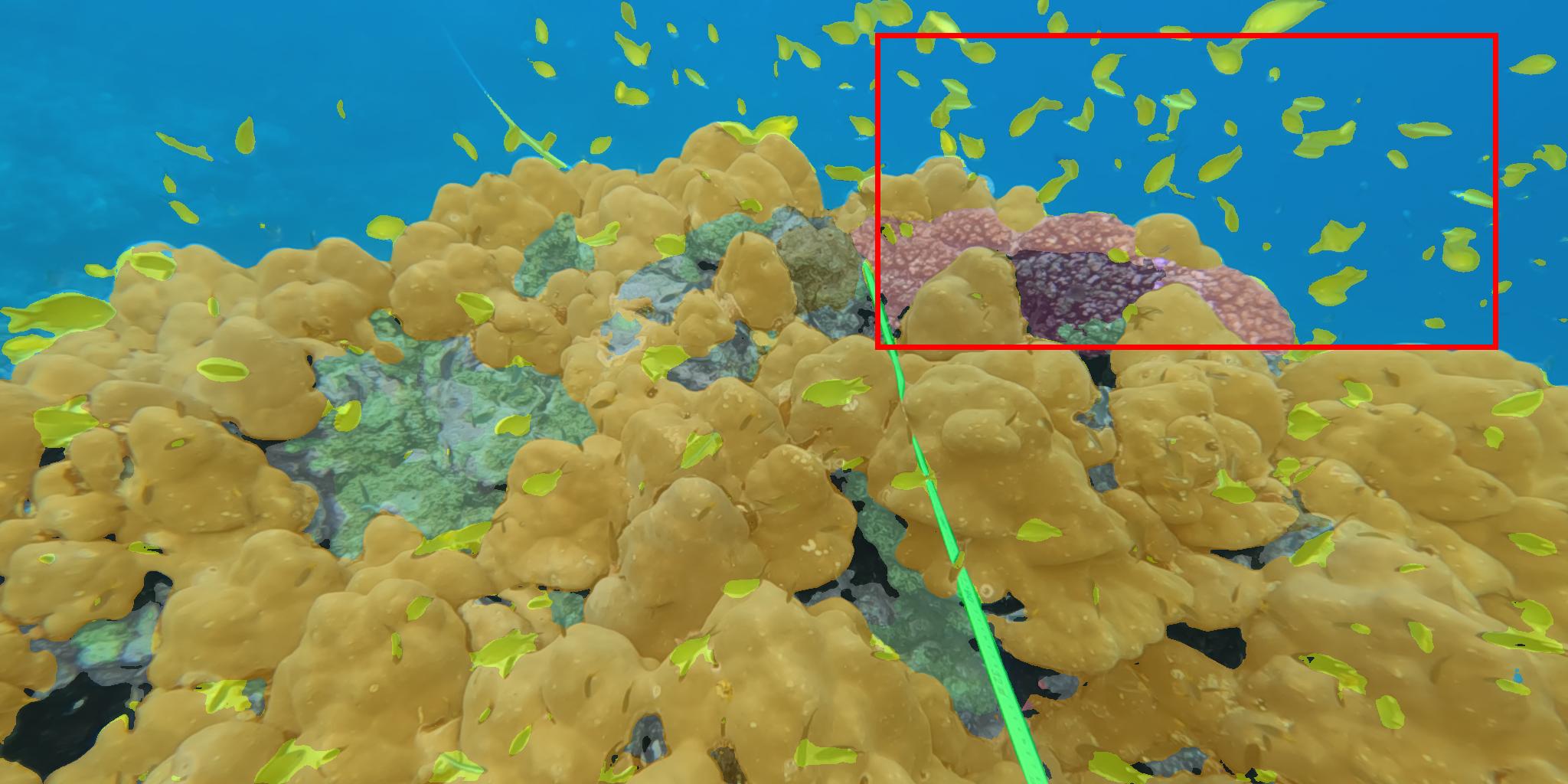} 
        \includegraphics[width=\linewidth]{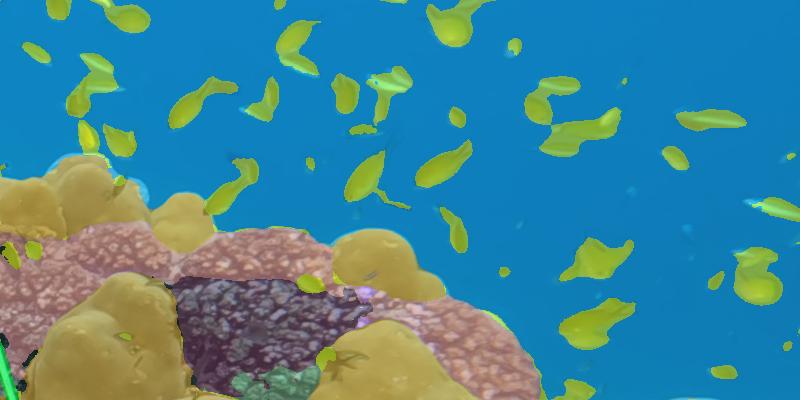} 
        \includegraphics[width=\linewidth]{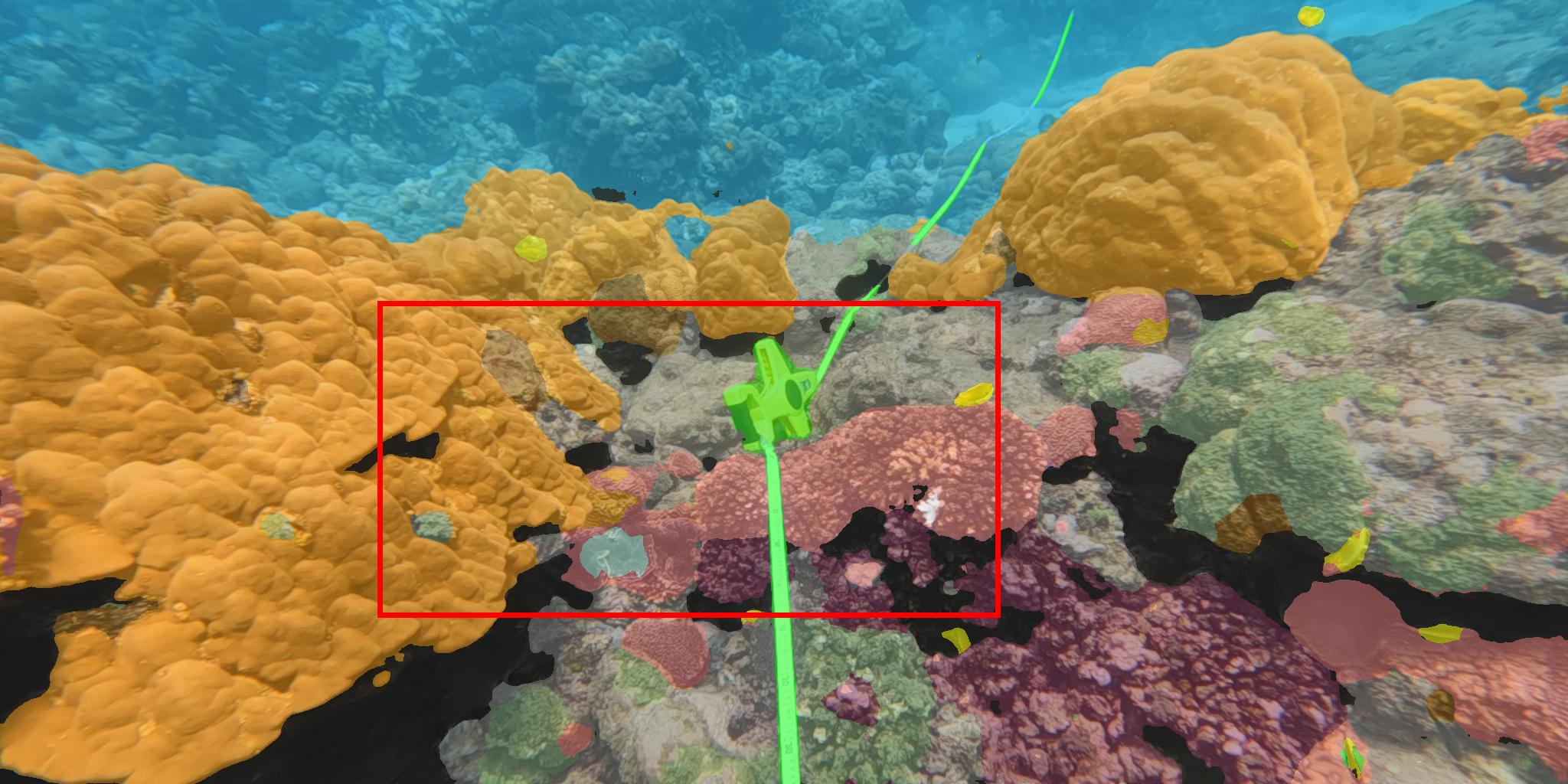} 
        \includegraphics[width=\linewidth]{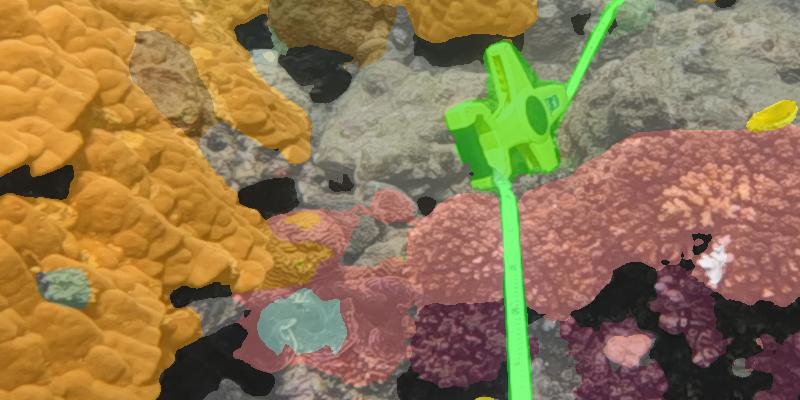} 
        \includegraphics[width=\linewidth]{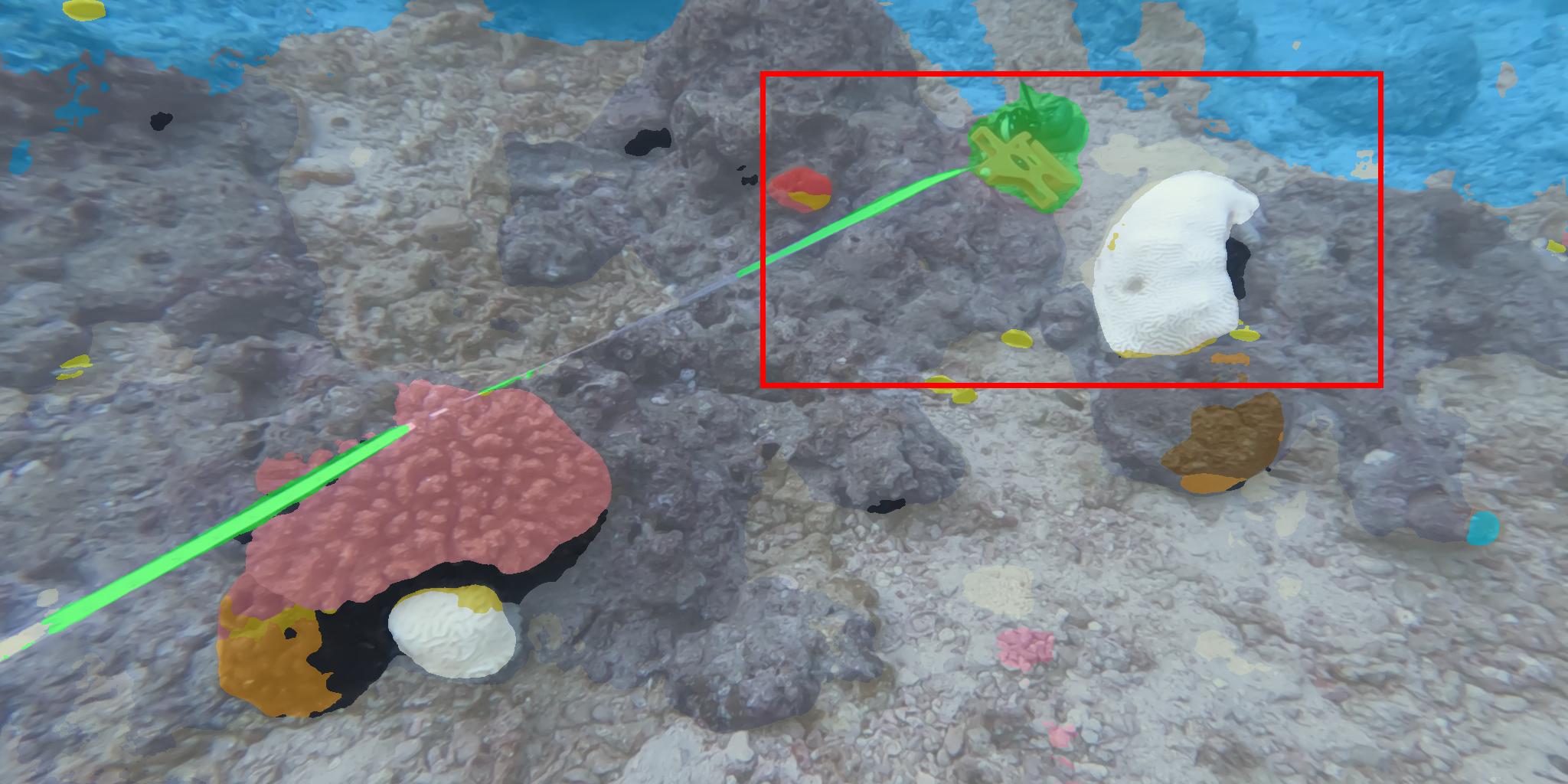} 
        \includegraphics[width=\linewidth]{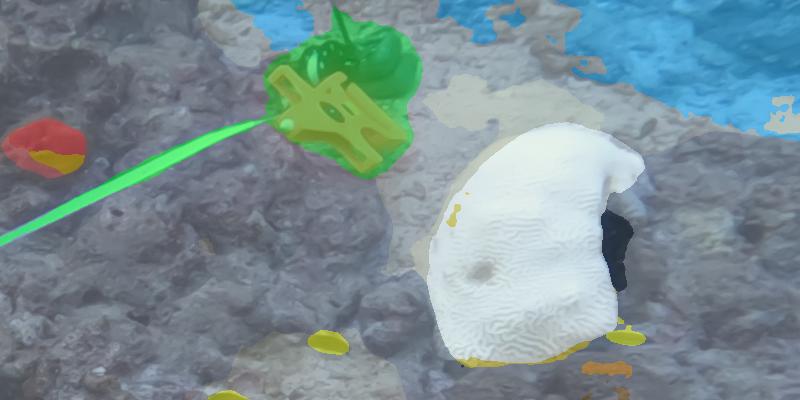}           
        \caption{\\DeepLabV3+}
        \label{fig:sub3}
    \end{subfigure}
    \begin{subfigure}{0.195\textwidth}
        \centering
        \includegraphics[width=\linewidth]{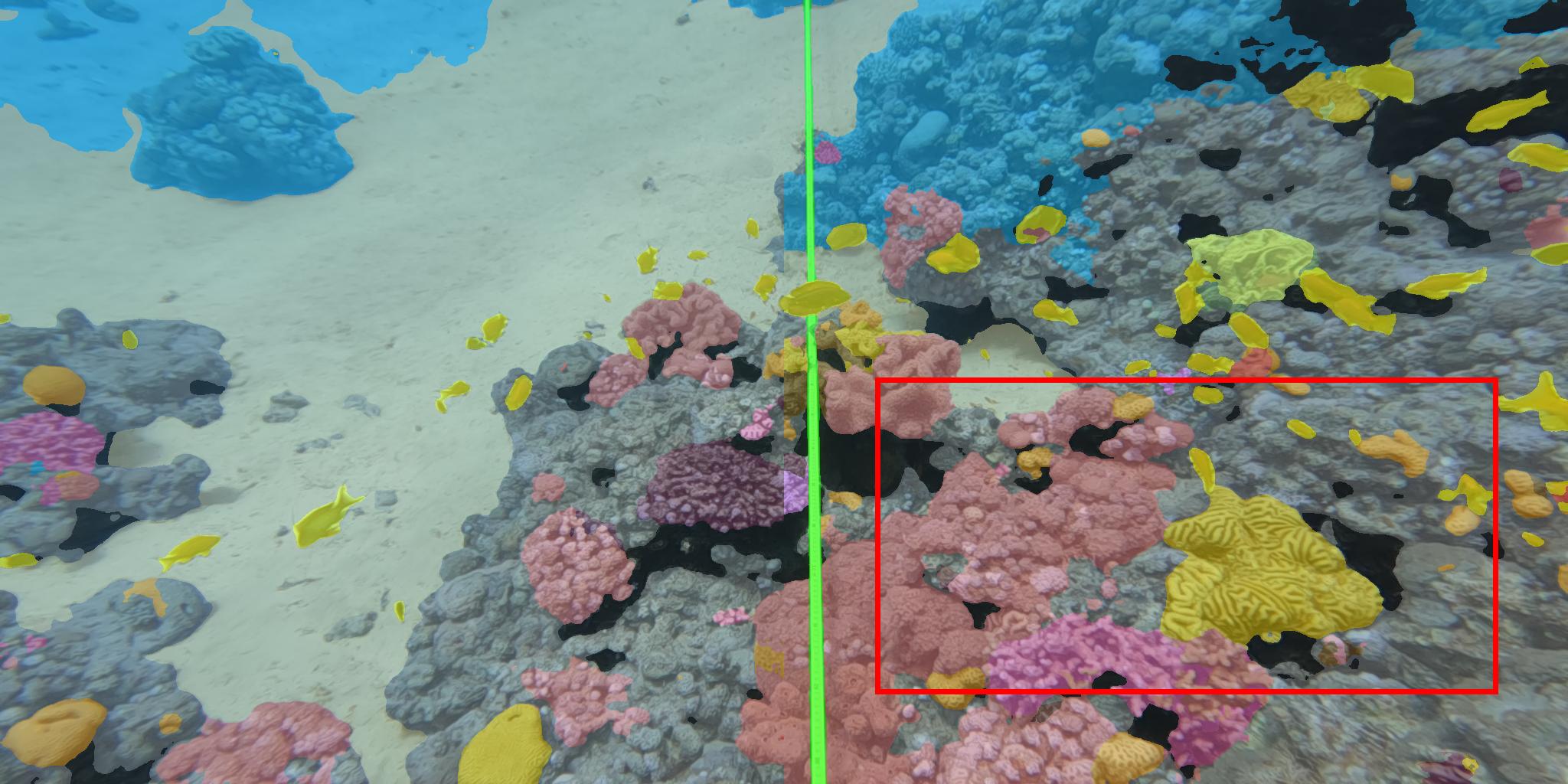} 
        \includegraphics[width=\linewidth]{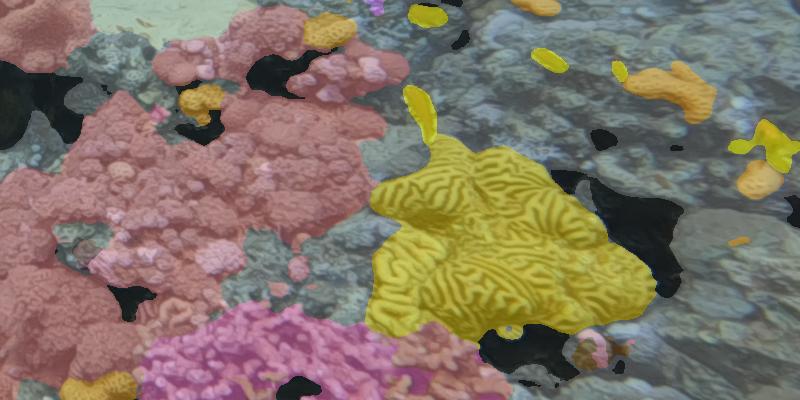} 
        \includegraphics[width=\linewidth]{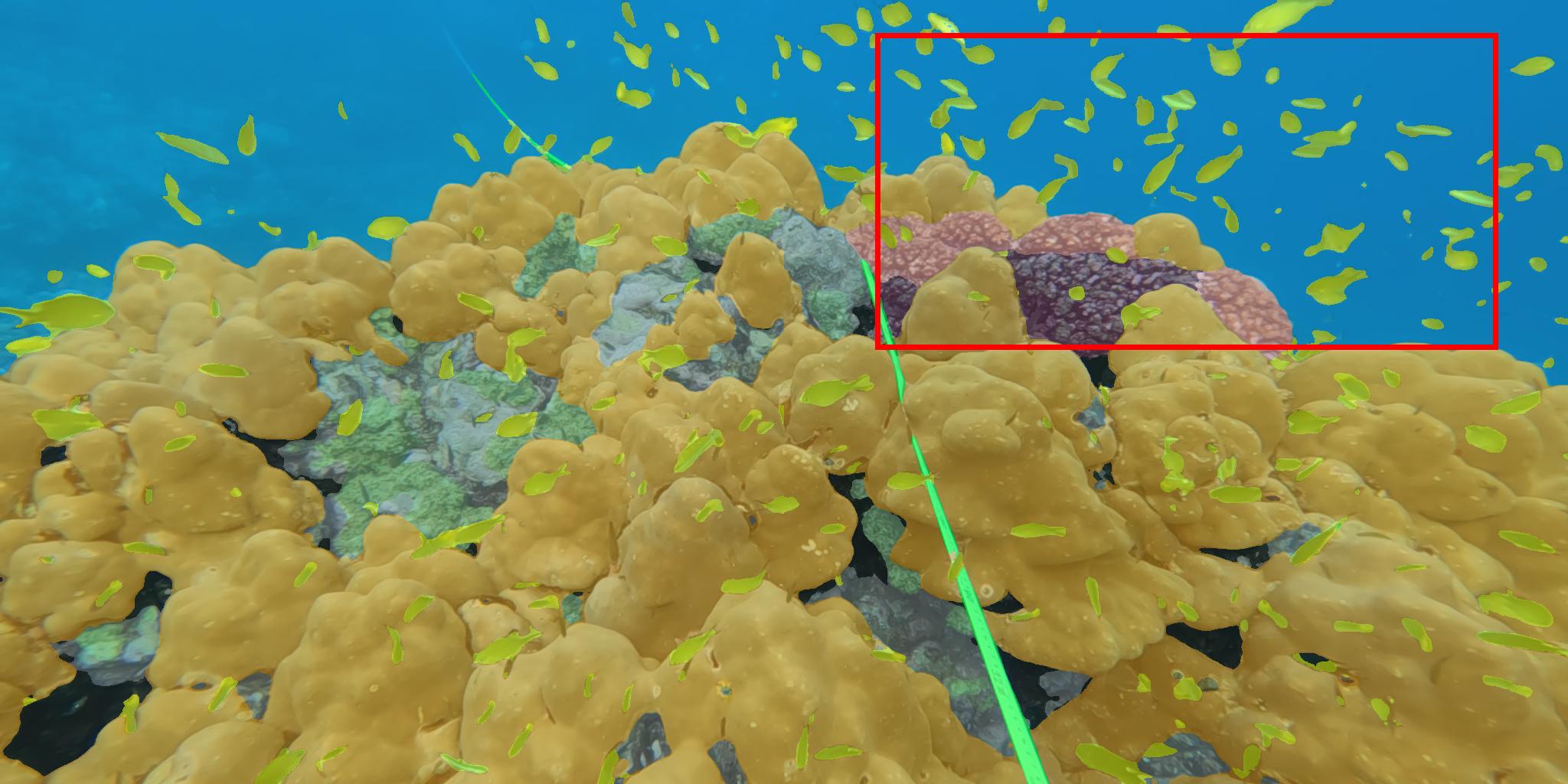}
        \includegraphics[width=\linewidth]{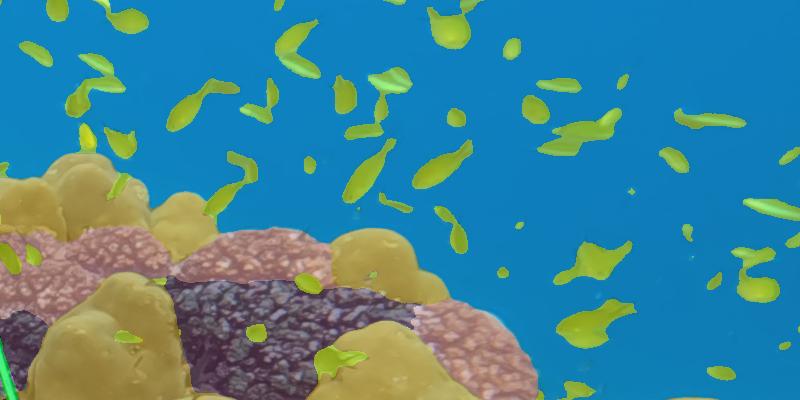} 
        \includegraphics[width=\linewidth]{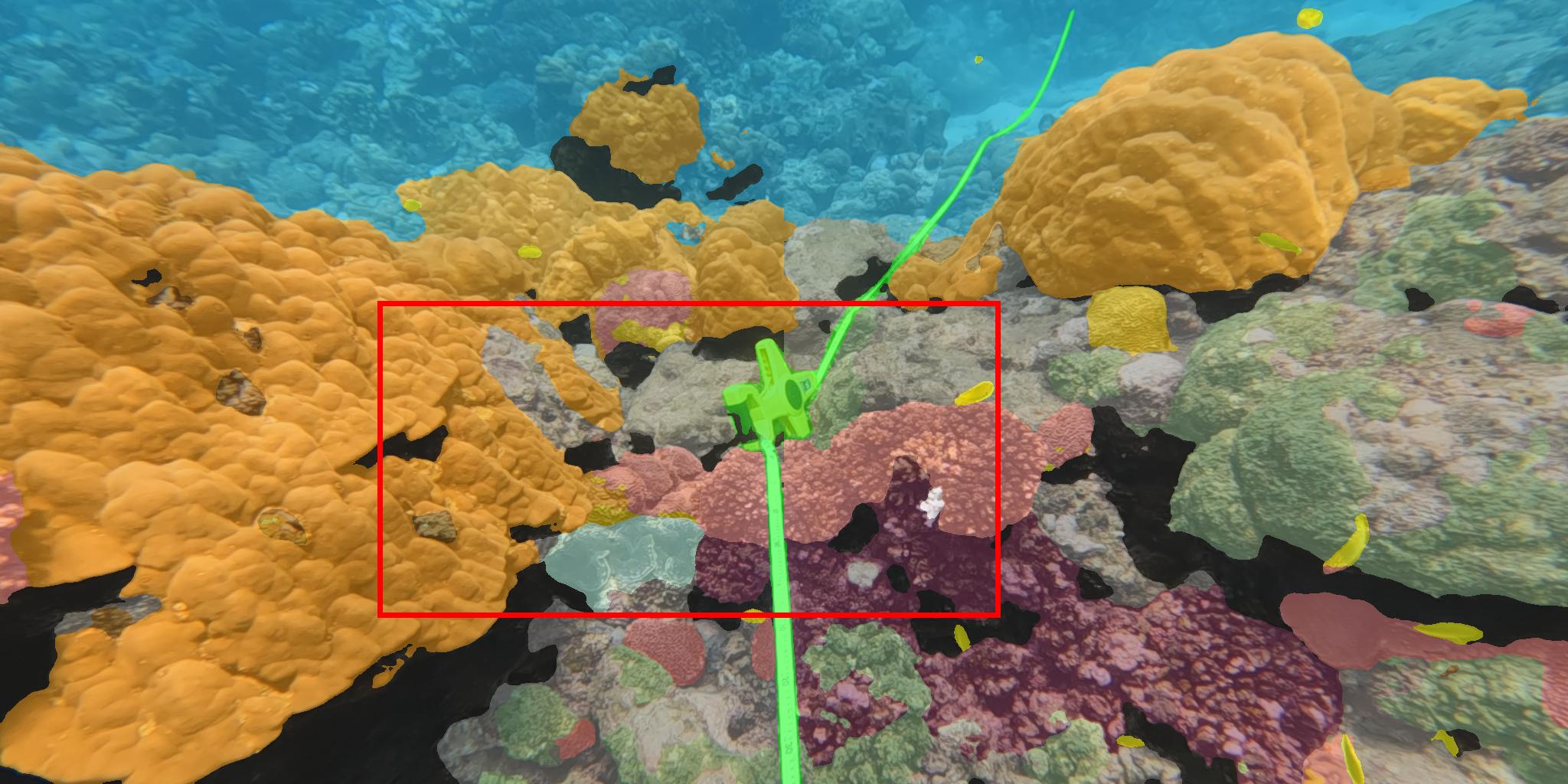} 
        \includegraphics[width=\linewidth]{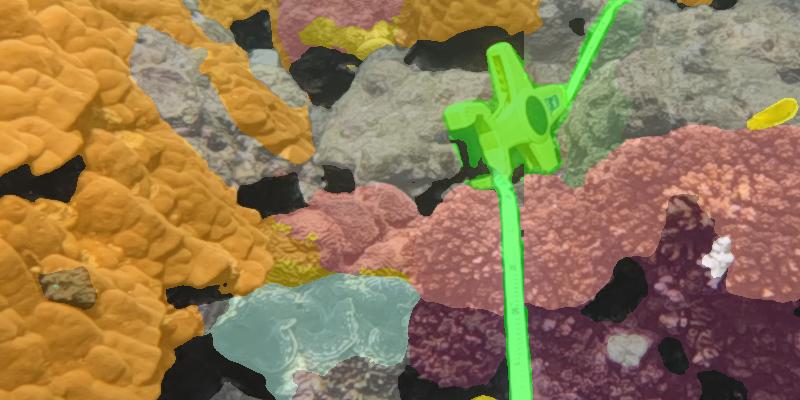} 
        \includegraphics[width=\linewidth]{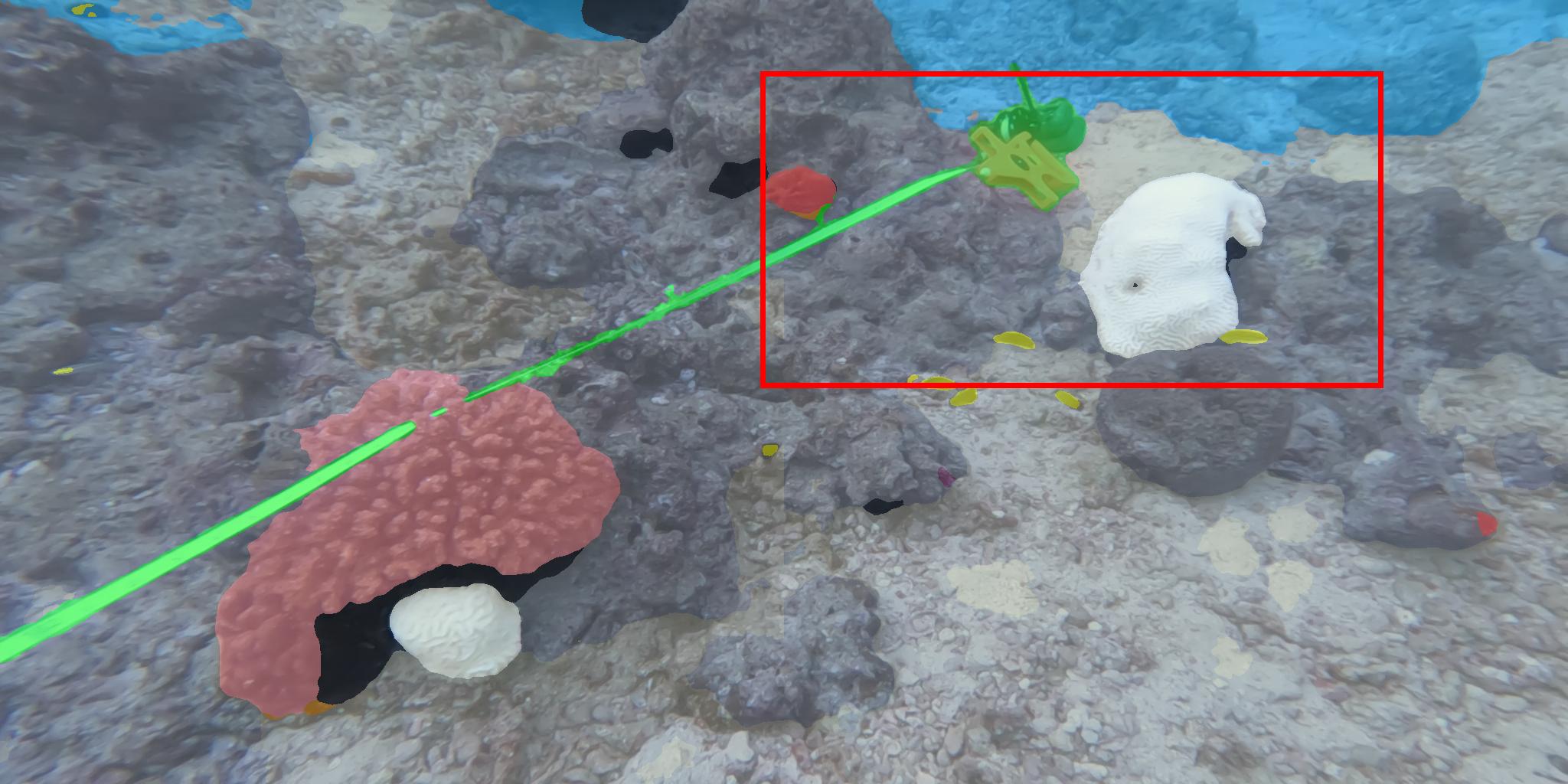} 
        \includegraphics[width=\linewidth]{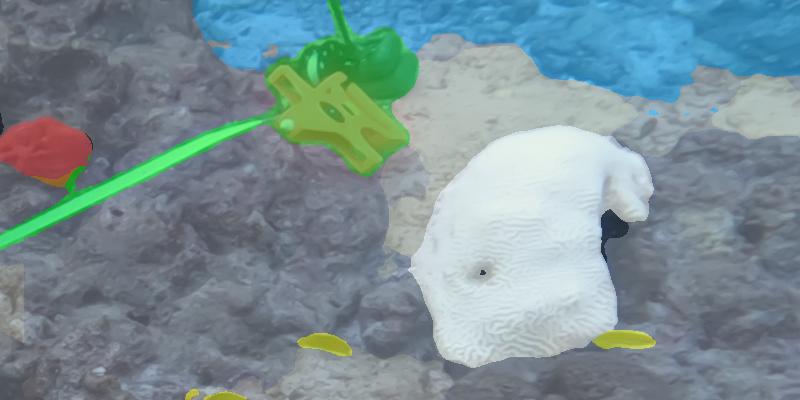}           
        \caption{\\SegFormer - MiT-b2}
        \label{fig:sub4}
    \end{subfigure}
    \begin{subfigure}{0.195\textwidth}
        \centering
        \includegraphics[width=\linewidth]{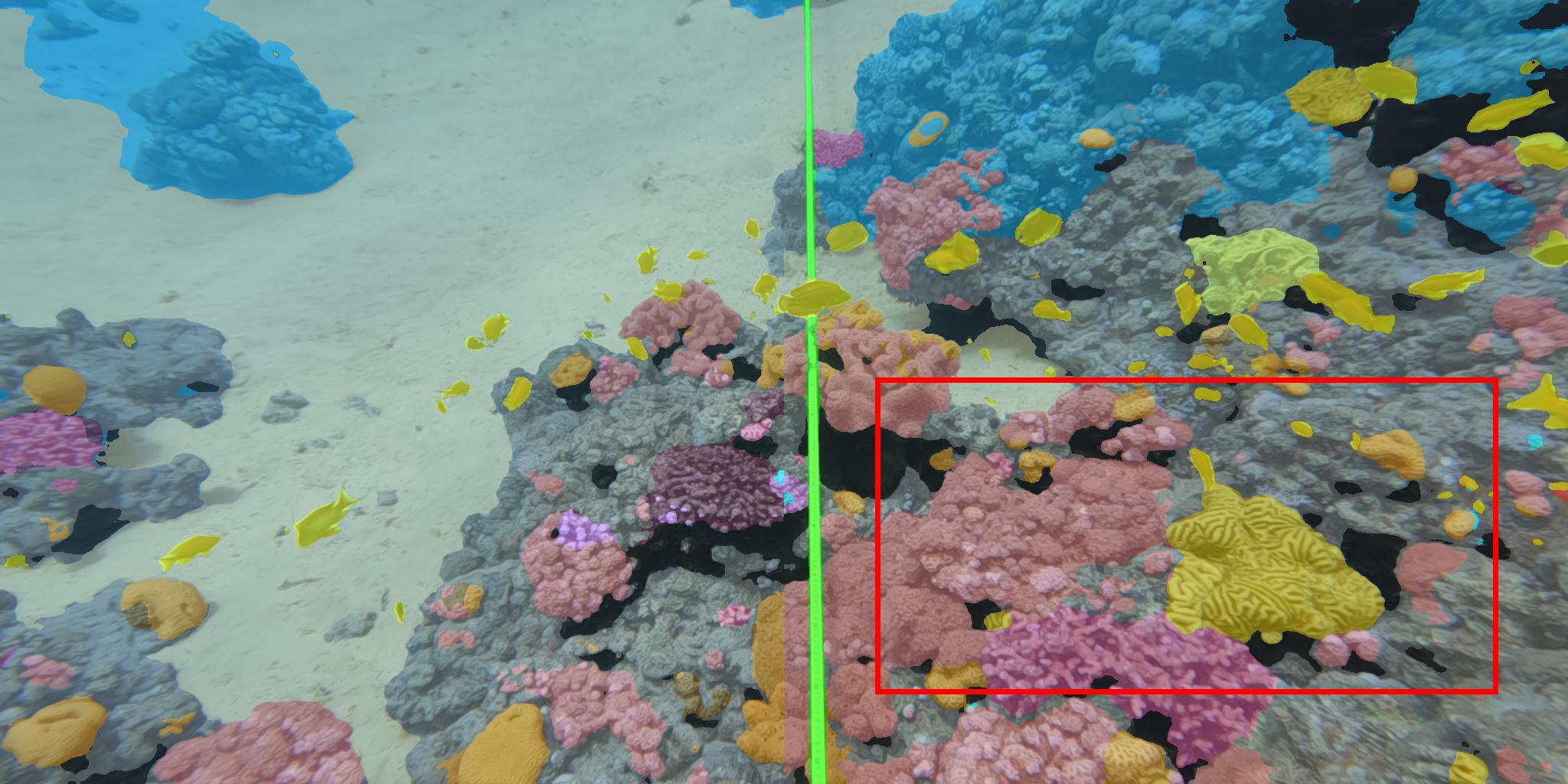} 
        \includegraphics[width=\linewidth]{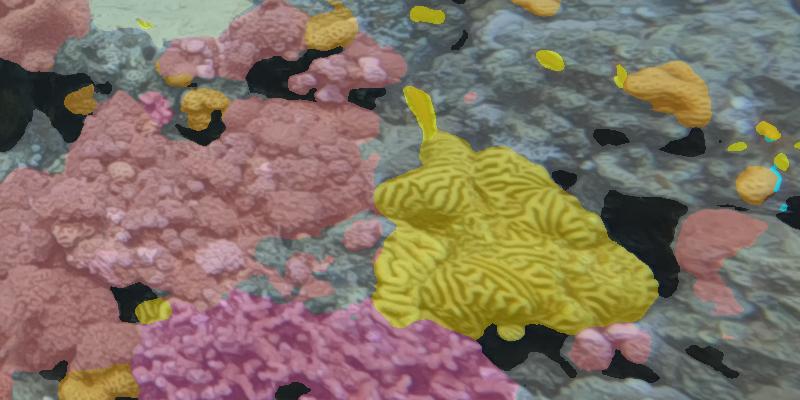} 
        \includegraphics[width=\linewidth]{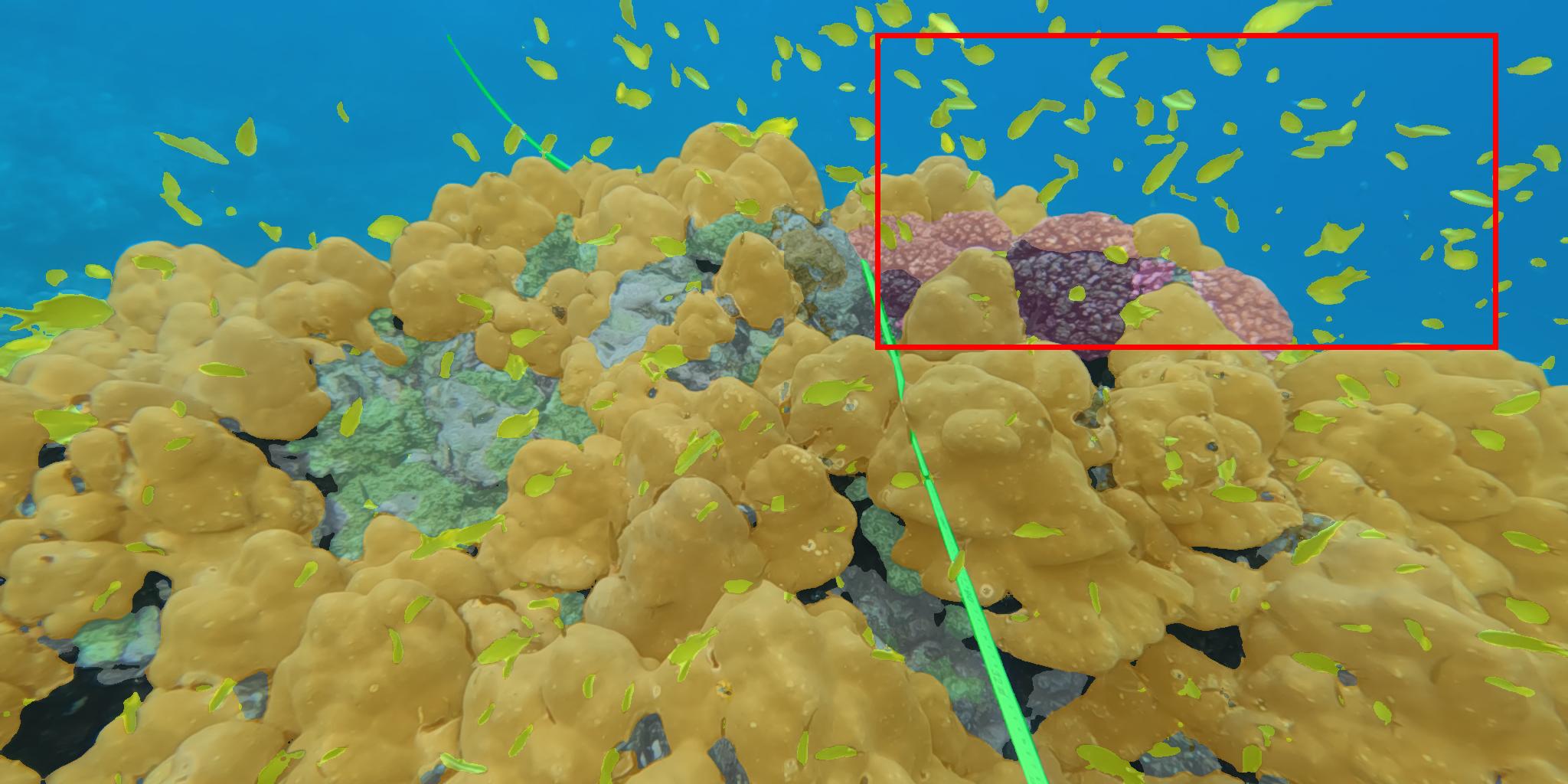}
        \includegraphics[width=\linewidth]{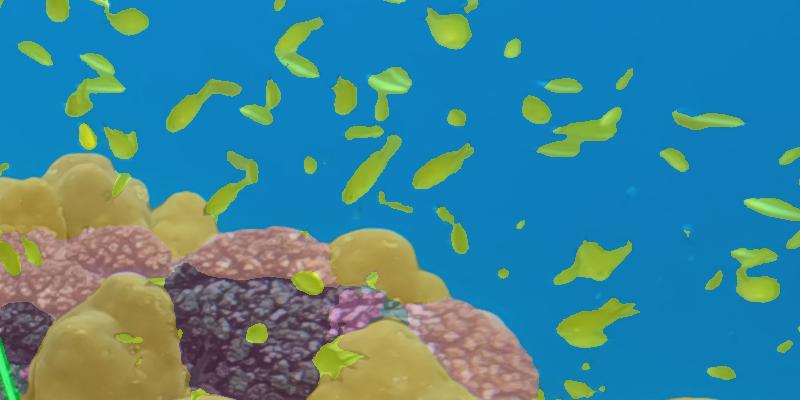} 
        \includegraphics[width=\linewidth]{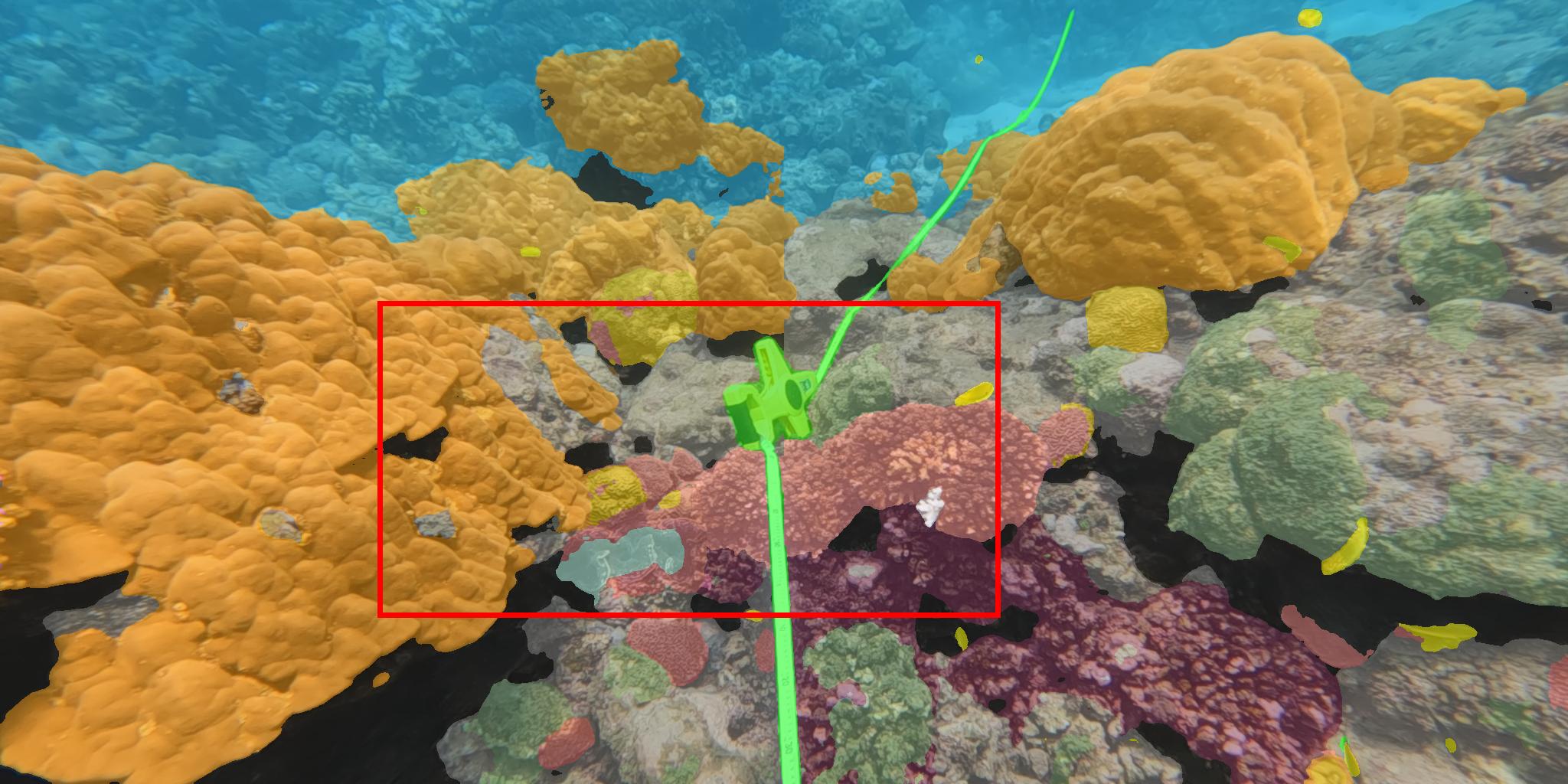} 
        \includegraphics[width=\linewidth]{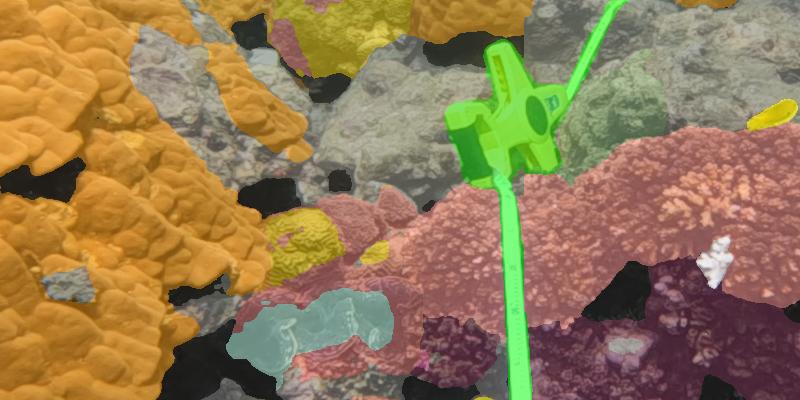} 
        \includegraphics[width=\linewidth]{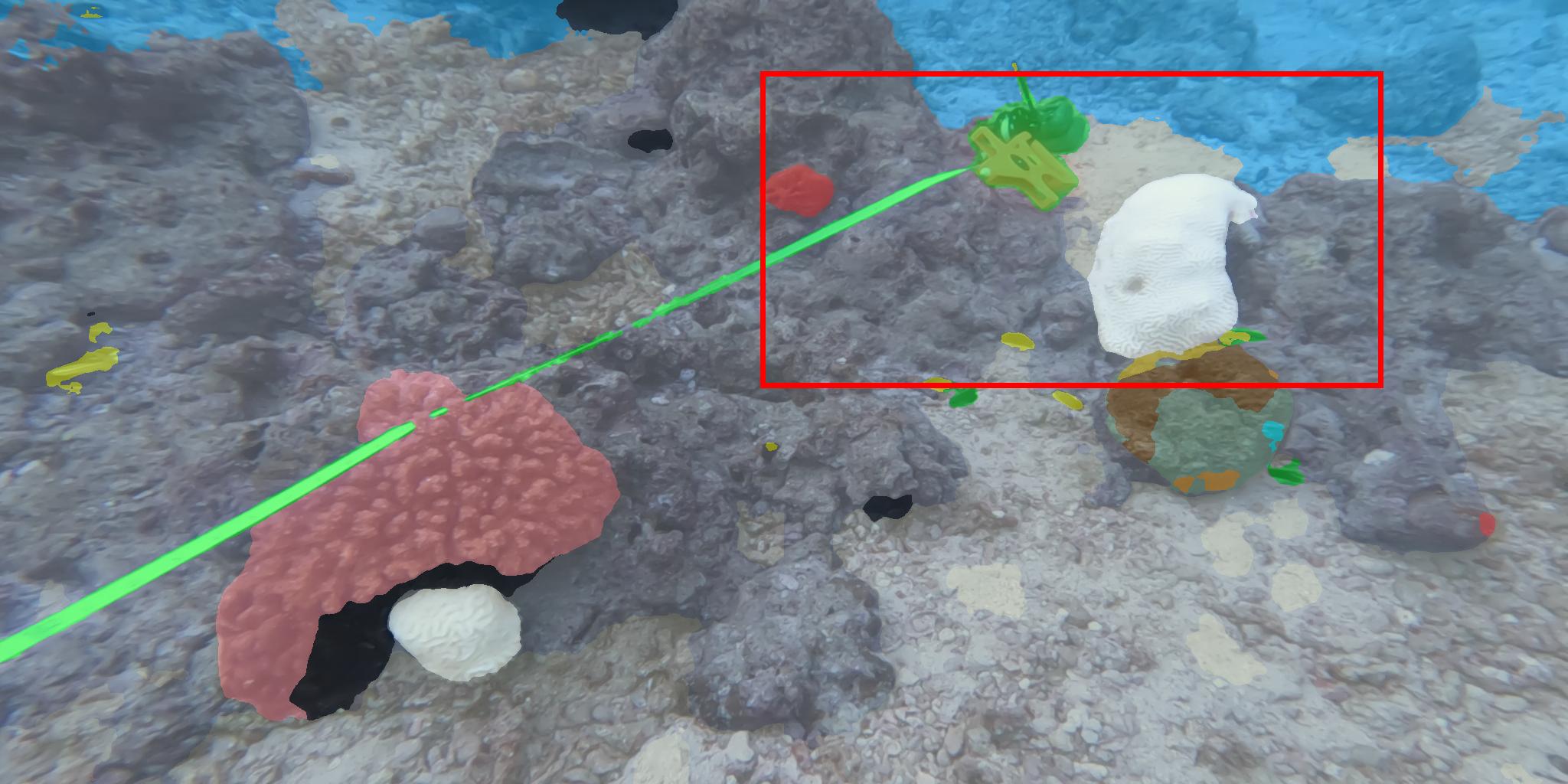} 
        \includegraphics[width=\linewidth]{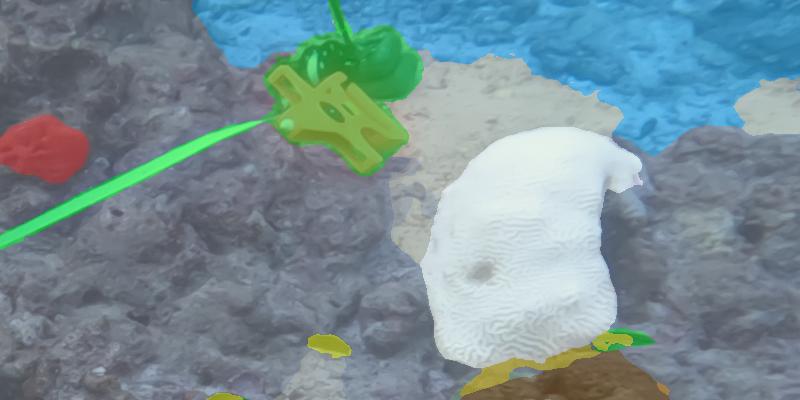}           

        \caption{\\SegFormer - MiT-b5}
        \label{fig:sub5}
    \end{subfigure}
    \vspace{-8pt}
    \caption{Qualitative samples from the Coralscapes test set. Additional samples provided in Appendix~\ref{appendix:additional_samples}.\vspace{-10pt}}
    \label{fig:main}
\end{figure*}

\begin{table*}[h!]
    \centering
    \resizebox{\textwidth}{!}{
    \begin{tabular}{l|ccccccccccc}
    \toprule
     & \textbf{Label} & \textbf{Accuracy} & \textbf{mIoU}\\
    \textbf{Method} & \textbf{Style} & \textbf{5 / 10 / 25 / 300 / All}  &\textbf{5 / 10 / 25 / 300 / All} \\
    \midrule
    PLAS \cite{raine2022point} & Grid & 65.73 / 70.18 / 73.60 / 83.72 / \hspace{9pt}-\hspace{9pt} & 19.48 / 26.01 / 36.27 / 52.83 / \hspace{9pt}-\hspace{9pt} \\
    HIL-S (DinoV2+K-NN) \cite{raine2024human} & Grid & 72.24 / 77.64 / 85.93 / 85.93 / \hspace{9pt}-\hspace{9pt} & 25.66 / 34.80 / 43.41 / 54.07 / \hspace{9pt}-\hspace{9pt} \\
    HIL-S (DinoV2+K-NN) \cite{raine2024human} & Human-in-the-Loop & 74.53 / 71.04 / 81.69 / 86.29 / \hspace{9pt}-\hspace{9pt} & 32.96 / 38.21 / 45.46 / 54.62 / \hspace{9pt}-\hspace{9pt} \\
    \midrule
    DeeplabV3+ - ResNet50 & Random & 80.29 / 81.21 / 83.18 / 86.54 / 87.09  &  36.39 / 38.54 / 43.24 / 51.59 / 52.65 \\
    (Ours) DeeplabV3+ - ResNet50 & Random & 81.56 / 82.70 / 83.87 / 86.66 / 87.56 & 40.27 / 42.53 / 46.14 / 53.22 / 55.68 \\
    SegFormer - MiT-B2 & Random & \underline{84.85} / \underline{85.79} / \underline{87.27} / \textbf{89.99} / \textbf{90.03}  &  \underline{47.59} / \underline{52.08} / \textbf{56.27} / \textbf{63.03} / \textbf{65.84} & \\
    (Ours) SegFormer - MiT-B2 & Random & \textbf{84.98} / \textbf{86.18} / \textbf{87.44} / \underline{89.04} / \underline{89.92}  & \textbf{48.36} / \textbf{52.41} / \underline{55.80} / \underline{62.01} / \underline{65.42}  \\
    SegFormer - MiT-B5 (Decoder) & Random & 76.41 / 77.58 / 79.15 / 81.59 / 81.98 & 27.10 / 30.22 / 32.98 / 39.32 / 40.70 \\
    (Ours) SegFormer - MiT-B5 (Decoder) & Random & 81.35 / 82.05 / 82.89 / 84.85 / 85.26 & 36.04 / 37.62 / 40.31 / 44.33 / 45.62 \\
    \bottomrule
   \end{tabular}}
    \caption{Transfer learning on UCSD Mosaics using 5 / 10 / 25 / 300 / all labeled pixels per image. Models marked with (Ours) are pre-trained on Coralscapes.\vspace{-10pt}}
    \label{tab:mosaics}
\end{table*}

We benchmark a diverse set of commonly used semantic segmentation architectures to establish baseline performances for Coralscapes.  Our evaluation includes both convolution-based and transformer-based models, covering a wide range of network complexities and training techniques. For convolutional architectures, we evaluate UNet++\cite{zhou2018unet++} and DeepLabV3+\cite{deeplabv3plus}, both using a ResNet-50 backbone. For transformer-based architectures, we benchmark SegFormer \cite{xie2021segformer} with both MiT-b2 and MiT-b5 backbones, each trained with and without Low-Rank Adaptation (LoRA) \cite{hu2022lora}. Similarly, we use DPT \cite{ranftl2021vision} with DINOv2 \cite{oquab2023dinov2} backbones at two encoder sizes (Base and Giant), again with and without LoRA fine-tuning. To further explore the capabilities of self-supervised vision transformers, we also include a linear segmentation head trained on top of DINOv2-Base features. The ResNet and MiT backbones are initialized with ImageNet-pretrained weights, while the DINOv2 backbones are pretrained on a large collection of images in a self-supervised fashion. 

Whenever available, we follow the original authors' training procedures on Cityscapes in terms of hyperparameters, augmentations, and inference/evaluation strategy, unless otherwise specified. Implementation details for all benchmarked models are provided in Appendix~\ref{appendix:implementation_details}. To choose the number of training epochs, we train each model on the training set only and evaluate on the validation set. We then re-train each model on the train+validation sets with the number of epochs leading to the highest validation mIoU (smoothed across 10 evaluation epochs in order to reduce the effect of noise).  Quantitative results are shown in Table~\ref{tab:benchmarking}, where we report performance of the models averaged over last three evaluated epochs, in order to reduce noise without training an ensemble of classifiers or cherry-picking results.

While the SegFormer models showed the best performance, DINOv2 models lead to competitive pixel accuracy despite lower mIoU. This likely stems from the fact that the DINOv2 models predict the segmentation maps for the image at once, where the global structure of the image helps determine the slightly subjective delineation of `background', as opposed to the strided patch predictions from SegFormer, which processes the image at higher resolution and can predict the fine-grained classes better. We find that deviating from the training procedure of SegFormer on the Cityscapes dataset by increasing the strength of data augmentations, the performance can be further improved, yielding the best model.
\vspace{-3pt}

\section{Transfer Learning \& Applications}
%\vspace{-2pt}
%In this section, we evaluate the potential of transferring models trained on Coralscapes to three downstream tasks.
\vspace{-3pt}

 \paragraph{UCSD Mosaics Transfer Learning}

We evaluate models trained on Coralscapes in a transfer learning setting on the UCSD Mosaics dataset \cite{edwards}, which consists of 16 orthomosaics densely annotated with 33 benthic classes and one class for the remaining substrate. We follow \cite{alonso2019coralseg,raine2022point,raine2024human} in dividing the orthomosaics into patches of size 512$\times$512px, and remove patches with corrupted ground-truth masks \cite{raine2022point}, leaving 3974 training and 696 test patches. We evaluate both in a dense segmentation setting, as well as in a commonly performed sparse-to-dense setting, in which the training annotations are sparsified, but evaluation is performed on the unchanged dense labels. In existing works \cite{alonso2019coralseg,raine2022point, raine2024human}, this setting is tackled by a two-step procedure, where in a first step, algorithms propagate the sparse labels into denser masks, and in a second step, a segmentation neural network is trained on the obtained denser masks. Our approach is conceptually simpler: we train segmentation models directly on the sparsified labels, ignoring the unlabeled pixels. We select 5, 10, 25, and 300 pixel locations per orthomosaic patch uniformly at random, comparing models pre-trained on Coralscapes (with the last layer re-initialized) against off-the-shelf models with a pre-trained backbone. The results, shown in Table~\ref{tab:mosaics}, show that pre-training on Coralscapes substantially improves segmentation with DeepLabV3+, particularly in highly sparse label regimes (5, 10, 25). When fine-tuning all parameters of a SegFormer MiT-b2, pre-training on Coralscapes improves performance in the sparse regimes, but not in the less sparse or dense (``All'' in Table~\ref{tab:mosaics}) settings. When training only the lightweight decoder on the MiT-b5 encoder weights obtained by pre-training on Coralscapes, we observe a consistent performance increase compared to an ImageNet pre-trained encoder. Interestingly, even an off-the-shelf DeepLabV3+ trained on sparse randomly located labels outperforms existing two-step procedures in sparse label regimes, despite \cite{raine2024human} showing that selecting pixels on a grid or with a human in the loop instead of randomly is beneficial. %TODO explain more clearly!
\vspace{-5pt}

\paragraph{Crown of Thorns Starfish Survey} 

\begin{table}[b!]
\vspace{-7pt}
    \centering
    \resizebox{0.99\columnwidth}{!}{\begin{tabular}{l|cccccc}
    \toprule
                & \multicolumn{2}{c}{\textbf{Video 0}} & \multicolumn{2}{c}{\textbf{Video 1}} & \multicolumn{2}{c}{\textbf{Video 2}} \\
        \textbf{Method} & \makecell{\textbf{mAP}\\\textbf{@50}} & \makecell{\textbf{mAP}\\\textbf{@50-95}}&  \makecell{\textbf{mAP}\\\textbf{@50}} & \makecell{\textbf{mAP}\\\textbf{@50-95}} &  \makecell{\textbf{mAP}\\\textbf{@50}} & \makecell{\textbf{mAP}\\\textbf{@50-95}} \\
    \midrule
        YoloV8-L  & 0.496 & \textbf{0.295} & 0.384 & \textbf{0.226} & 0.413 & \textbf{0.222} \\
        SegFormer - MiT-b5 & 0.443 & 0.199 &   0.359 & 0.141 & 0.364 & 0.082 \\
        (Ours) SegFormer - MiT-b5 & \textbf{0.542} & 0.288 & \textbf{0.393} & 0.196 & \textbf{0.512} & 0.194\\
        \bottomrule
    \end{tabular}}
    \caption{COTS detection results in a leave-one-out test-set setting for each of the three videos.\vspace{-5pt}
}
    \label{tab:cots_table}
\end{table}

\begin{figure*}[t!]
\centering
\begin{subfigure}{0.147\textwidth}
\captionsetup{font=small,justification=centering, singlelinecheck=off}
\includegraphics[width=\linewidth,height=41px]{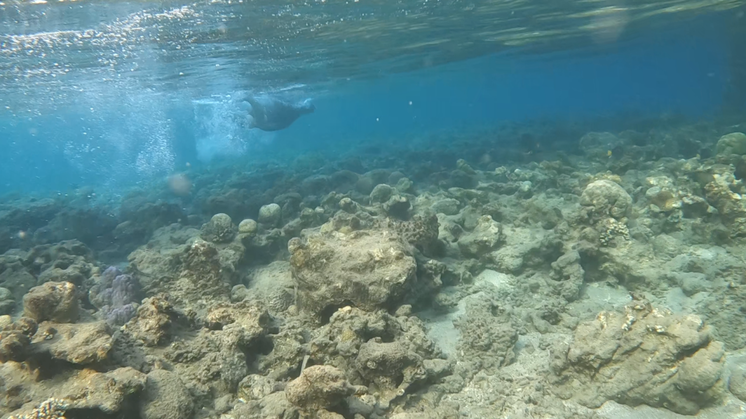}    
\includegraphics[width=\linewidth,height=41px]{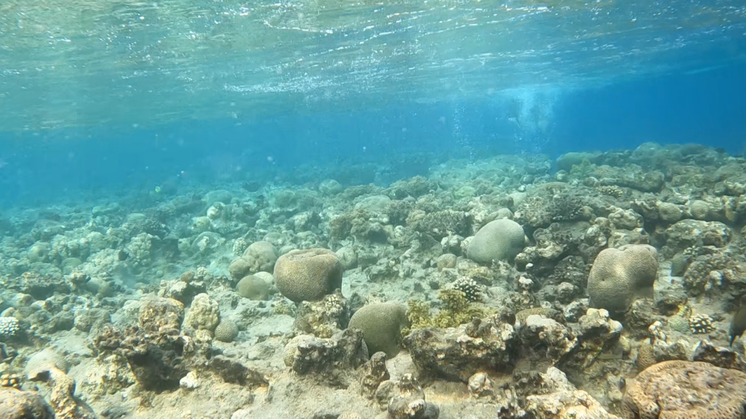}    \\
\vspace{-9pt}

\includegraphics[width=\linewidth,height=41px]{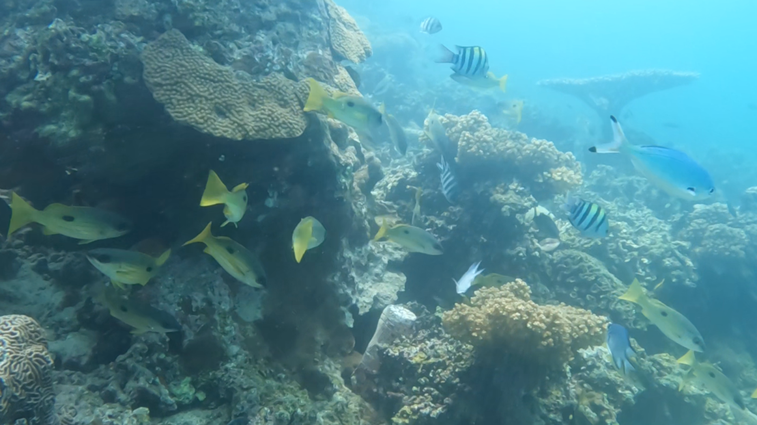}    
\includegraphics[width=\linewidth,height=41px]{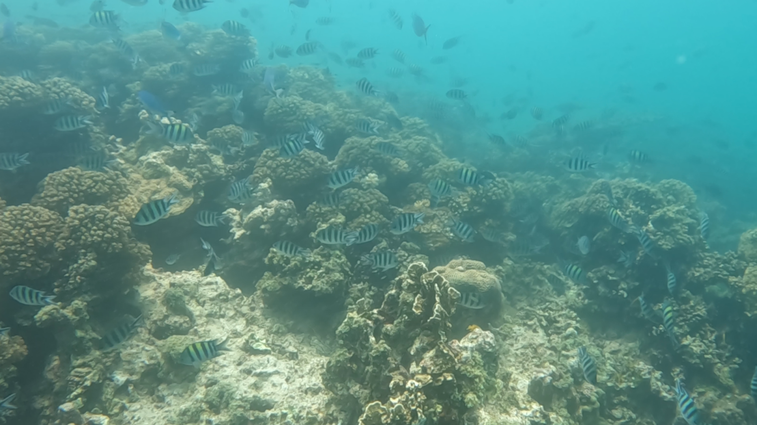}    \\
\vspace{-9pt}

\includegraphics[width=\linewidth,height=41px]{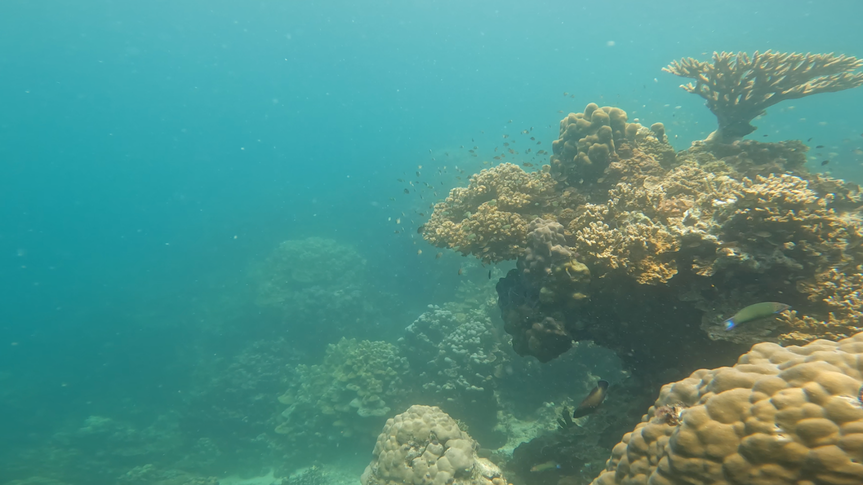}    
\includegraphics[width=\linewidth,height=41px]{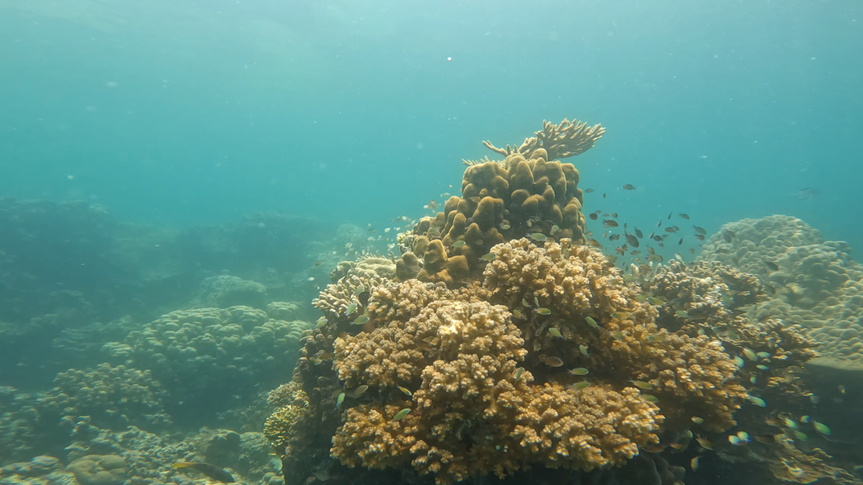}   
\caption{\phantom{x}\\Images\\\phantom{x}}
\end{subfigure}
\begin{subfigure}{0.147\textwidth}
\captionsetup{font=small,justification=centering, singlelinecheck=off}
\includegraphics[width=\linewidth,height=41px]{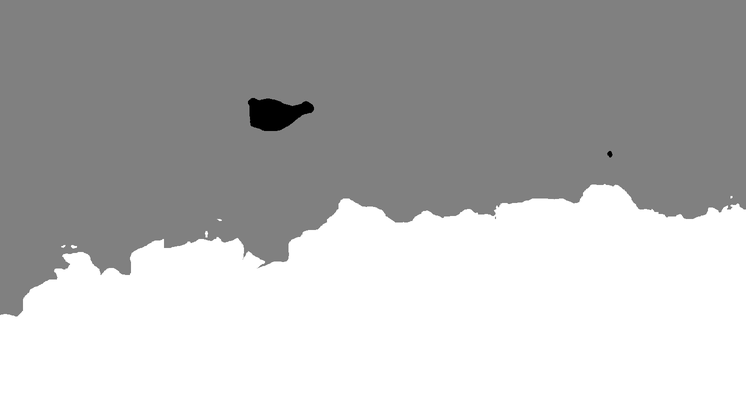}    
\includegraphics[width=\linewidth,height=41px]{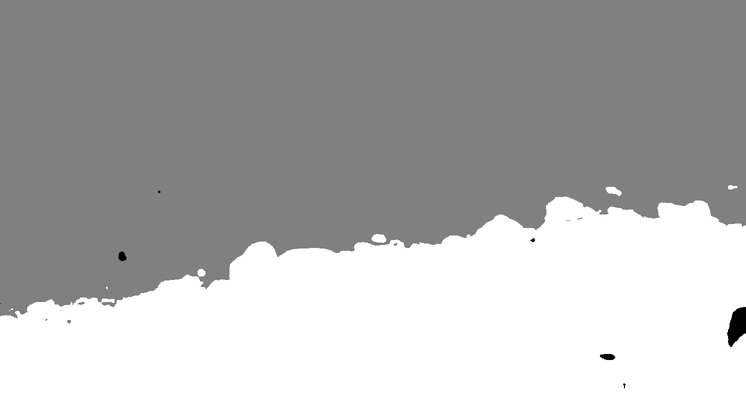}    \\
\vspace{-9pt}

\includegraphics[width=\linewidth,height=41px]{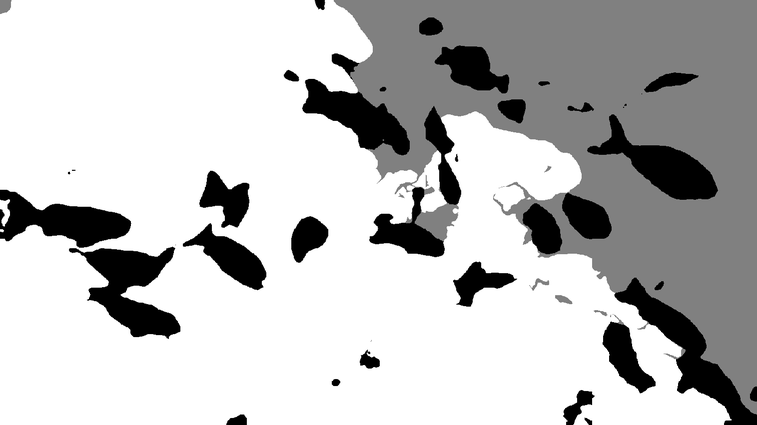}    
\includegraphics[width=\linewidth,height=41px]{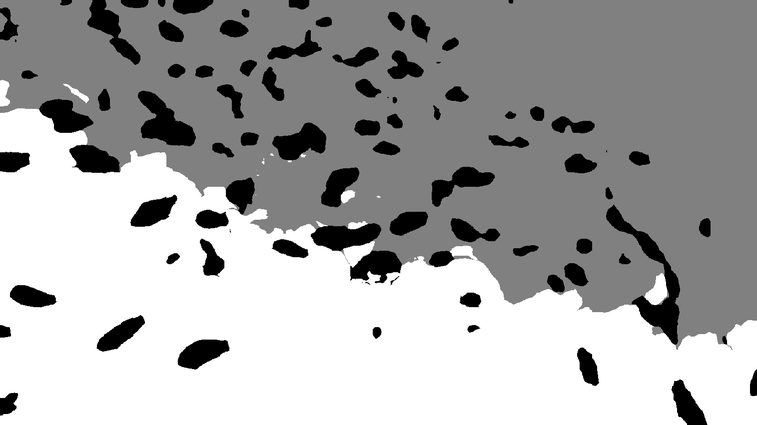}    \\
\vspace{-9pt}

\includegraphics[width=\linewidth,height=41px]{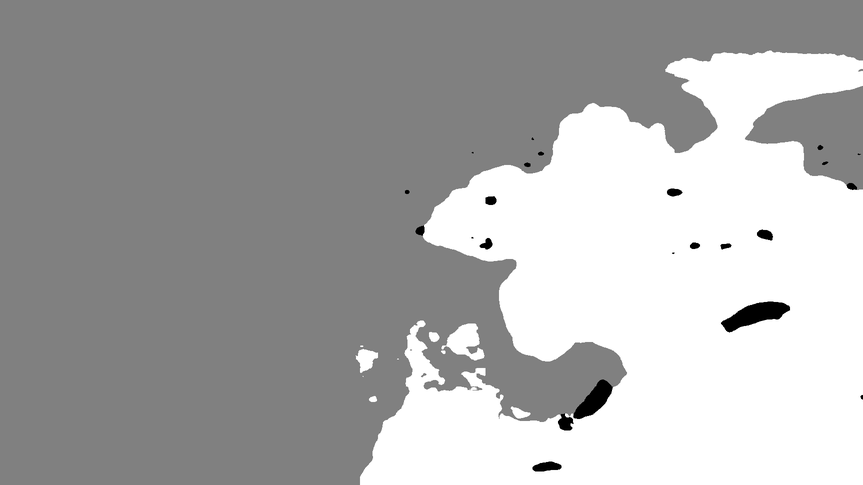}    
\includegraphics[width=\linewidth,height=41px]{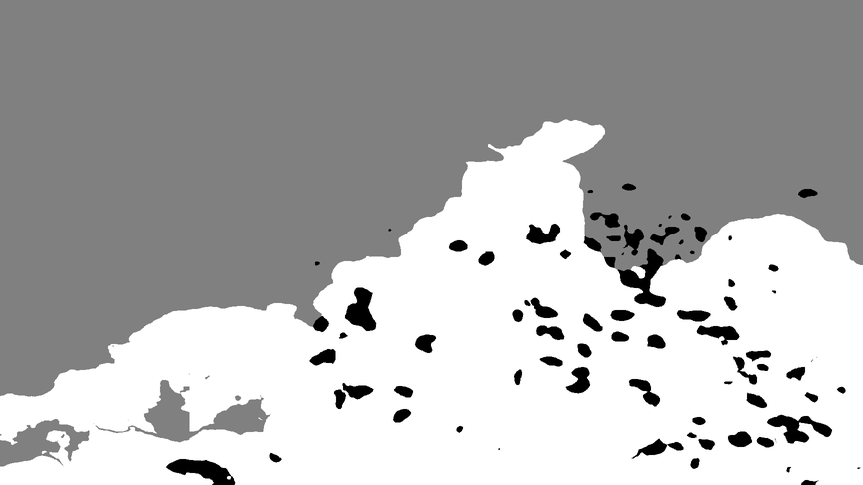}   
\caption{\phantom{x}\\Masks\\\phantom{x}}
\end{subfigure}
\begin{subfigure}{0.177\textwidth}
\captionsetup{font=small,justification=centering, singlelinecheck=off}
\includegraphics[width=\linewidth,height=83px]{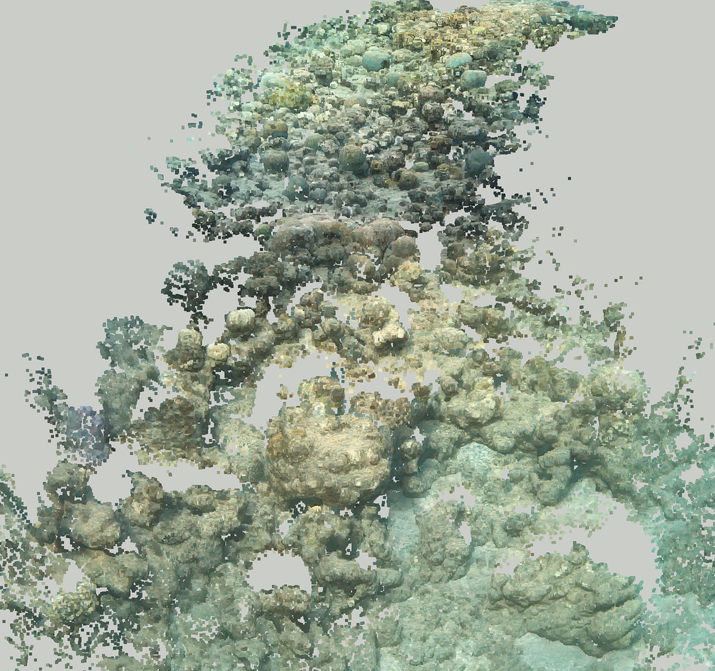}    \\
\vspace{-9pt}

\includegraphics[width=\linewidth,height=83px]{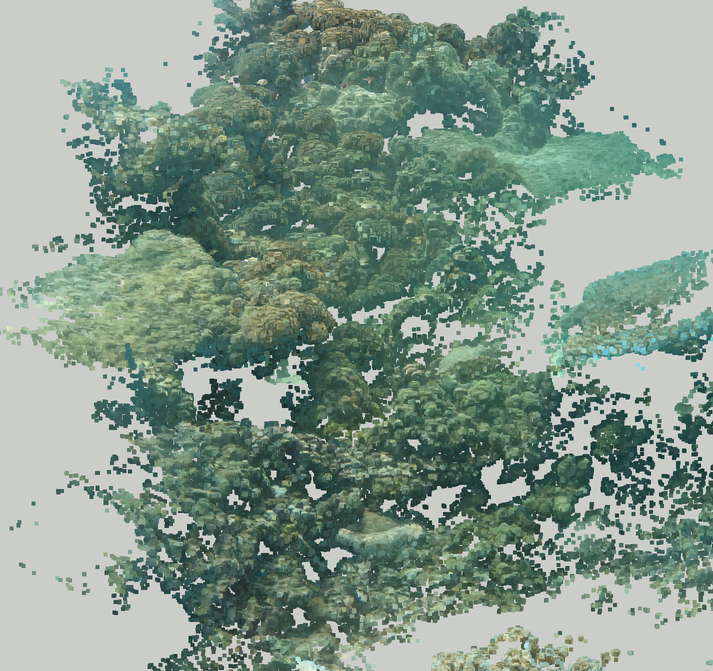}    \\
\vspace{-9pt}

\includegraphics[width=\linewidth,height=83px]{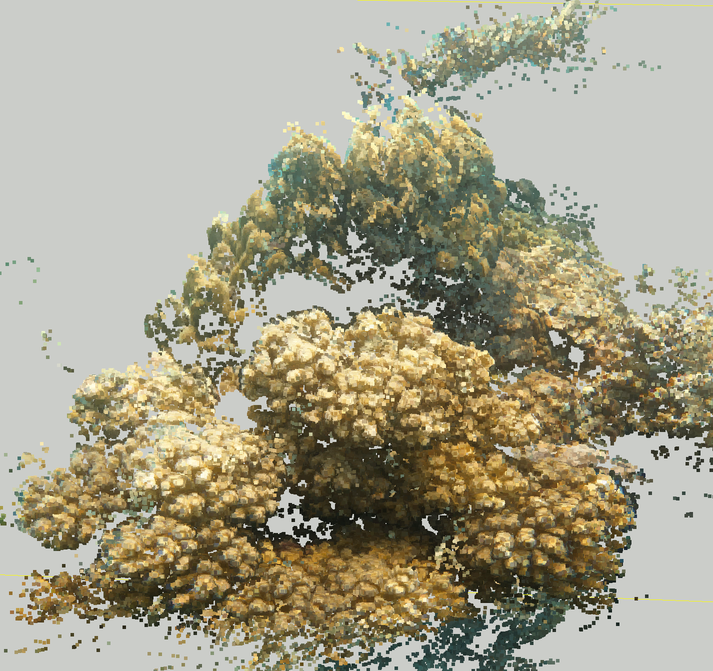}  
\caption{\phantom{x}\\3D Reconstruction\\Ours}
\end{subfigure}
\begin{subfigure}{0.177\textwidth}
\captionsetup{font=small,justification=centering, singlelinecheck=off}
\includegraphics[width=\linewidth,height=83px]{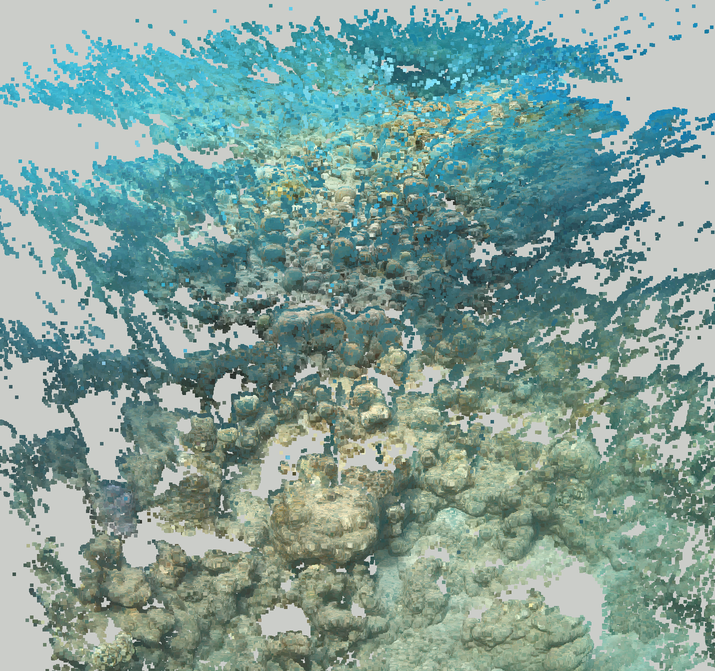}   \\
\vspace{-9pt}

\includegraphics[width=\linewidth,height=83px]{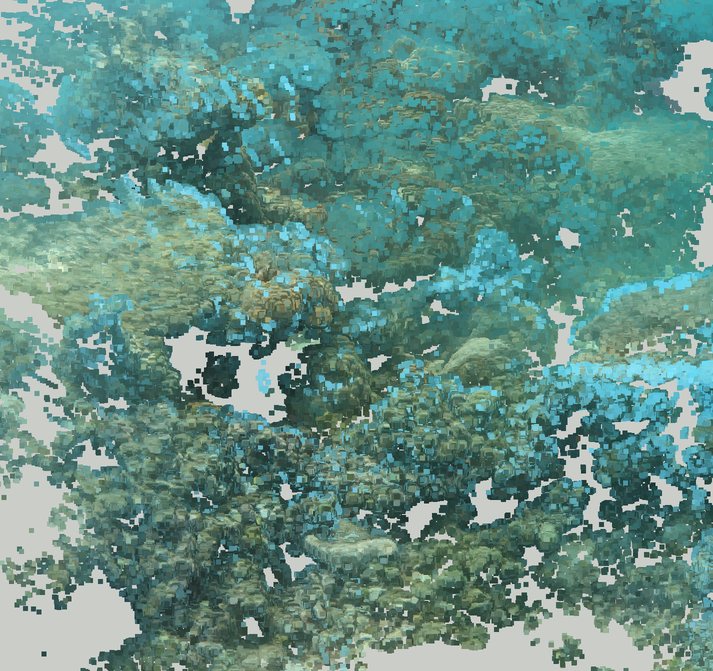}    \\
\vspace{-9pt}

\includegraphics[width=\linewidth,height=83px]{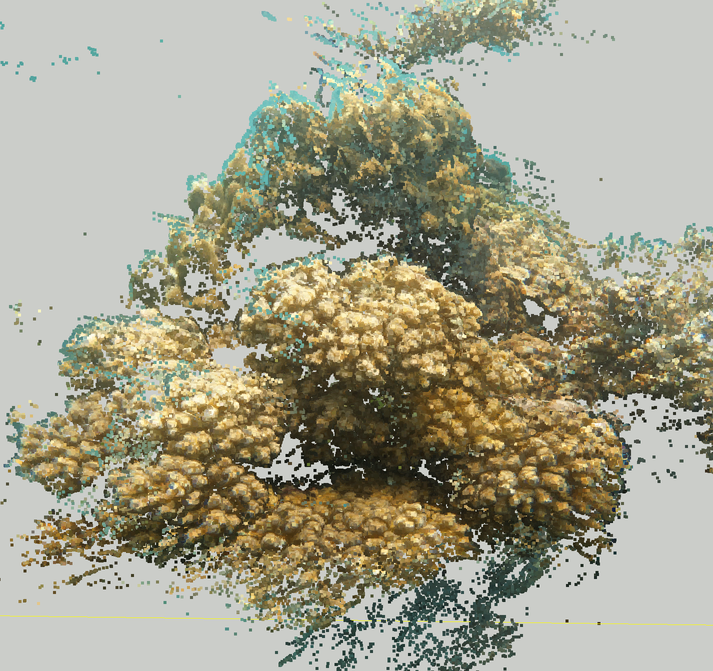}  
\caption{\phantom{x}\\3D Reconstruction\\No Masks}
\end{subfigure}
\begin{subfigure}{0.157\textwidth}
\captionsetup{font=small,justification=centering, singlelinecheck=off}
\includegraphics[width=\linewidth,height=41px]{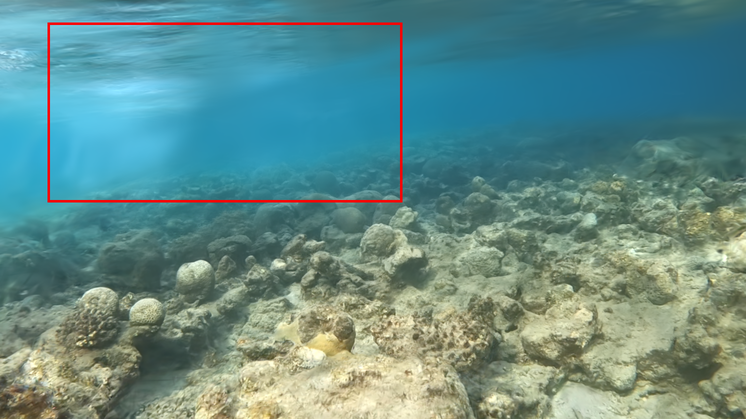}    
\includegraphics[width=\linewidth,height=41px]{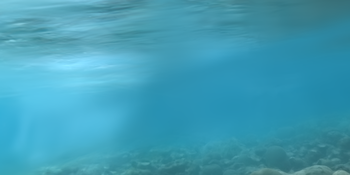}  \\ 
\vspace{-9pt}

\includegraphics[width=\linewidth,height=41px]{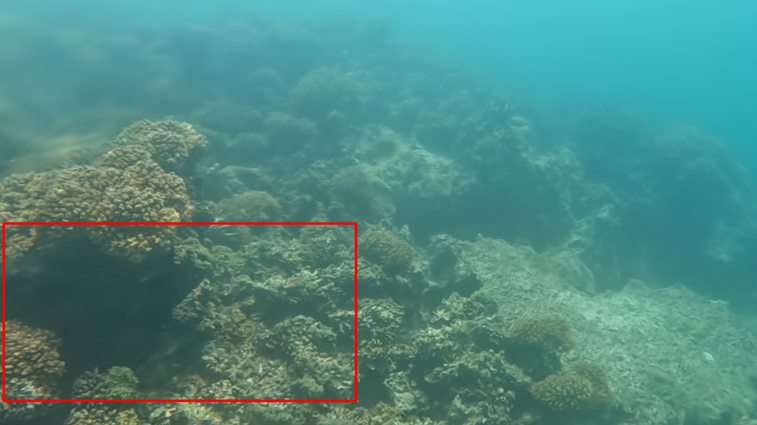}    
\includegraphics[width=\linewidth,height=41px]{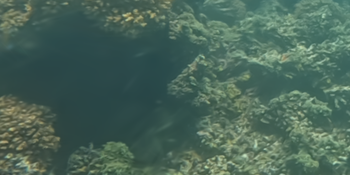}  \\ 
\vspace{-9pt}

\includegraphics[width=\linewidth,height=41px]{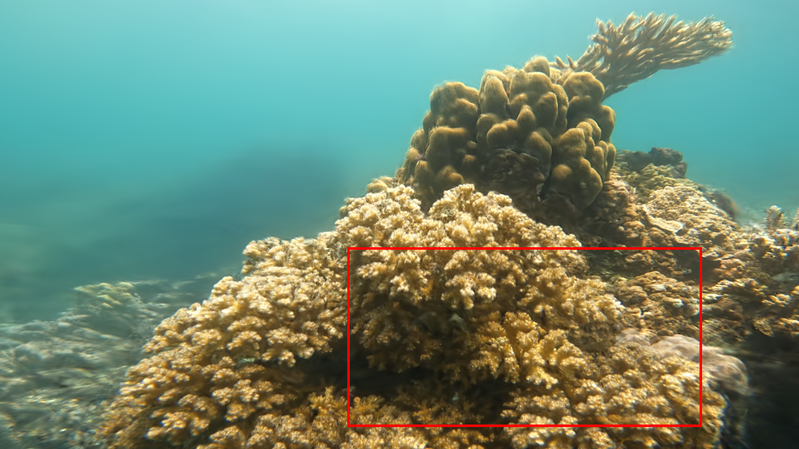}    
\includegraphics[width=\linewidth,height=41px]{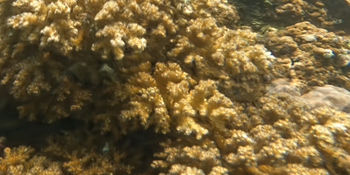} 
\caption{\phantom{x}\\3DGS Novel View\\Ours}
\end{subfigure}
\begin{subfigure}{0.157\textwidth}
\captionsetup{font=small,justification=centering, singlelinecheck=off}
\includegraphics[width=\linewidth,height=41px]{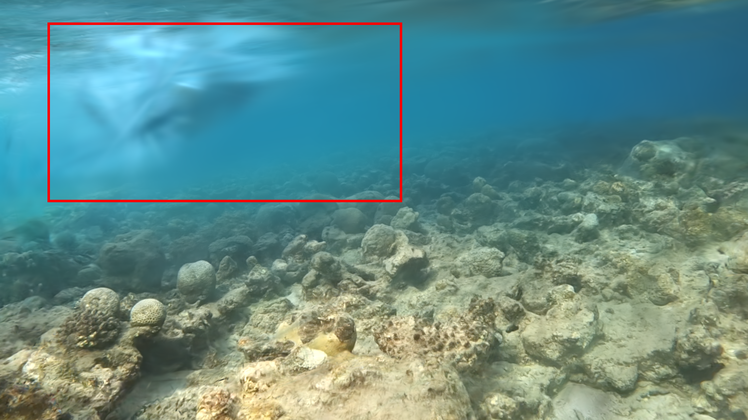} 
\includegraphics[width=\linewidth,height=41px]{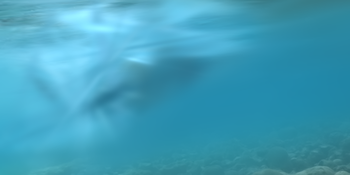} \\
\vspace{-9pt}

\includegraphics[width=\linewidth,height=41px]{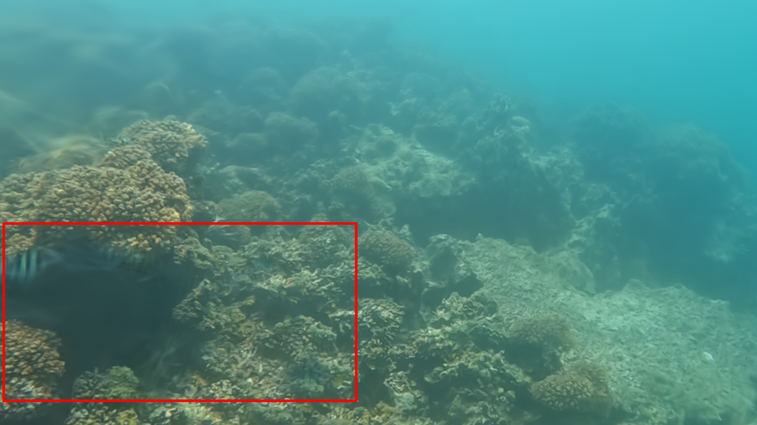} 
\includegraphics[width=\linewidth,height=41px]{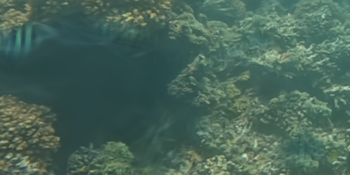} \\
\vspace{-9pt}

\includegraphics[width=\linewidth,height=41px]{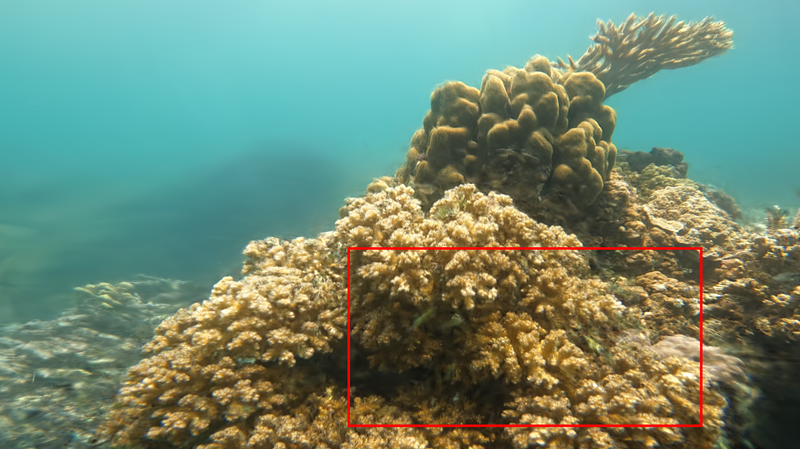} 
\includegraphics[width=\linewidth,height=41px]{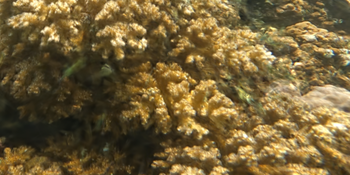}
\caption{\phantom{x}\\3DGS Novel View\\No Masks}
\end{subfigure}
\vspace{-8pt}
\caption{Using masks obtained from Coralscapes to mask out unwanted classes can alleviate artifacts from dense 3D reconstruction and from novel view synthesis such as 3DGS.\vspace{-8pt}}
\label{fig:3d}
\end{figure*}

\textit{Acanthaster planci}, colloquially known as the Crown of Thorns starfish (COTS), is a common reef inhabitant in the entire Indo-Pacific, where it preys on hard corals. While in normal population sizes, COTS are a healthy part of the ecosystem, there are outbreaks of COTS, in which their population increases dramatically and the hard coral cover declines, which can lead to devastating consequences for entire reefs \cite{cots1,cots2}. Interventions against COTS outbreaks are most effective in the early stages of an outbreak \cite{cots3,cots4}: computer vision can help to rapidly detect such outbreaks by automatically analyzing imagery from reefs for COTS abundance. In \cite{cots}, a dataset of three videos from manta tows (a snorkeler with a camera, towed behind a boat) in Australia were annotated with bounding boxes of COTS.
The videos contain 12347, 11374, and 10759 frames at 720 $\times$ 1280 px with 3065, 6384, and 2449 bounding-box annotations of COTS respectively. We use the bounding-boxes as segmentation labels to fine-tune a SegFormer-MiT-b5 pre-trained on Coralscapes for 10 epochs, with bounding boxes being drawn around connected components of the prediction for evaluation. The results (Table~\ref{tab:cots_table}) show that pre-training on Coralscapes leads to a substantial improvement compared to using an ImageNet pre-trained backbone. Compared to a Yolo-V8-L baseline, the transfer-learned segmentation model consistently improves the mAP@50, but the mAP@50-95, which is more sensitive to the bounding box locations, is lower. This is likely due to the naive approach to transforming segmentations into bounding boxes.

\vspace{-5pt}
\paragraph{Underwater 3D Mapping from Videos}
\vspace{-2pt}

Even though 3D photogrammetry is widely used to map coral reefs  \cite{coralsfm1,coralsfm2,coralsfm3,coralsfm4,coralsfm5,coralsfm6,reefscapegenomics}, commonly used pipelines are designed for in-air mapping and often struggle in underwater environments. The detrimental effects of the water column include changes in the color of objects relative to the distance to the camera and induced blur \cite{seathru,backscatternet}. Dynamic objects such as moving fish or divers also lead to artifacts in the 3D reconstructions or even complete failure. Therefore, image collections are usually highly curated \cite{sfminstructions} to exclude images with such objects or with the water column visible. 

Instead of constraining the input data, general purpose semantic segmentation from Coralscapes can be used to mask out unwanted classes during 3D reconstruction pipelines, expanding the range of useful images for 3D reconstruction. We qualitatively demonstrate this on three reef videos, using GLOMAP \cite{glomap} (at 2FPS in sequential matching mode) for sparse SfM, on which we run dense reconstruction with COLMAP \cite{COLMAP} as well as 3D Gaussian Splatting (3DGS) \cite{3dgs} for novel view synthesis. Example images, masks and results are shown in Figure~\ref{fig:3d}, where detrimental artifacts from the background are alleviated in dense reconstruction, and moving objects (fish and divers) are removed in 3DGS novel view synthesis. Similarly, masks obtained from models trained on Coralscapes can be used in other neural rendering techniques \cite{seathrunerf, uwgs, watersplat}, or neural monocular SLAM systems \cite{deepreefmap}.
\vspace{-5pt}

\section{\vspace{-6pt}Conclusion and Future Work}
%\vspace{-4pt}

We present the Coralscapes dataset for semantic understanding in reef scenes, which is the first general purpose dataset of coral scenes that are densely and consistently annotated by trained experts. 
Coralscapes is representative of the challenges of semantic segmentation in reef scenes, including complex scenes with challenging environmental conditions.
By benchmarking state-of-the-art computer vision models, we confirm that Coralscapes is a challenging benchmark due to the imbalance in class occurrence and feature details. The Coralscapes dataset aims to catalyze the development of computer vision applications in coral reef science and conservation. Future work will extend Coralscapes to further increase the diversity (e.g. new biogeographic regions, different camera models, onboard lighting), and add more fine-grained classes, also including fish genera or species, and include instance annotations for fish.

%- labeling corals extremely challening
%- future dataset: include more acquisition diversity, e.g. with on-board lighting, different camera models \& intrinsics. Include fish \& instance segmentations
\FloatBarrier
\newpage
{\small
\bibliographystyle{ieee_fullname}
\bibliography{egbib}
}

\newpage
\appendix
\onecolumn
\section{Appendix: Additional Dataset Information}
\label{appendix:additional_data}
\begin{figure}[h]
    \centering
\includegraphics[width=0.99\textwidth]{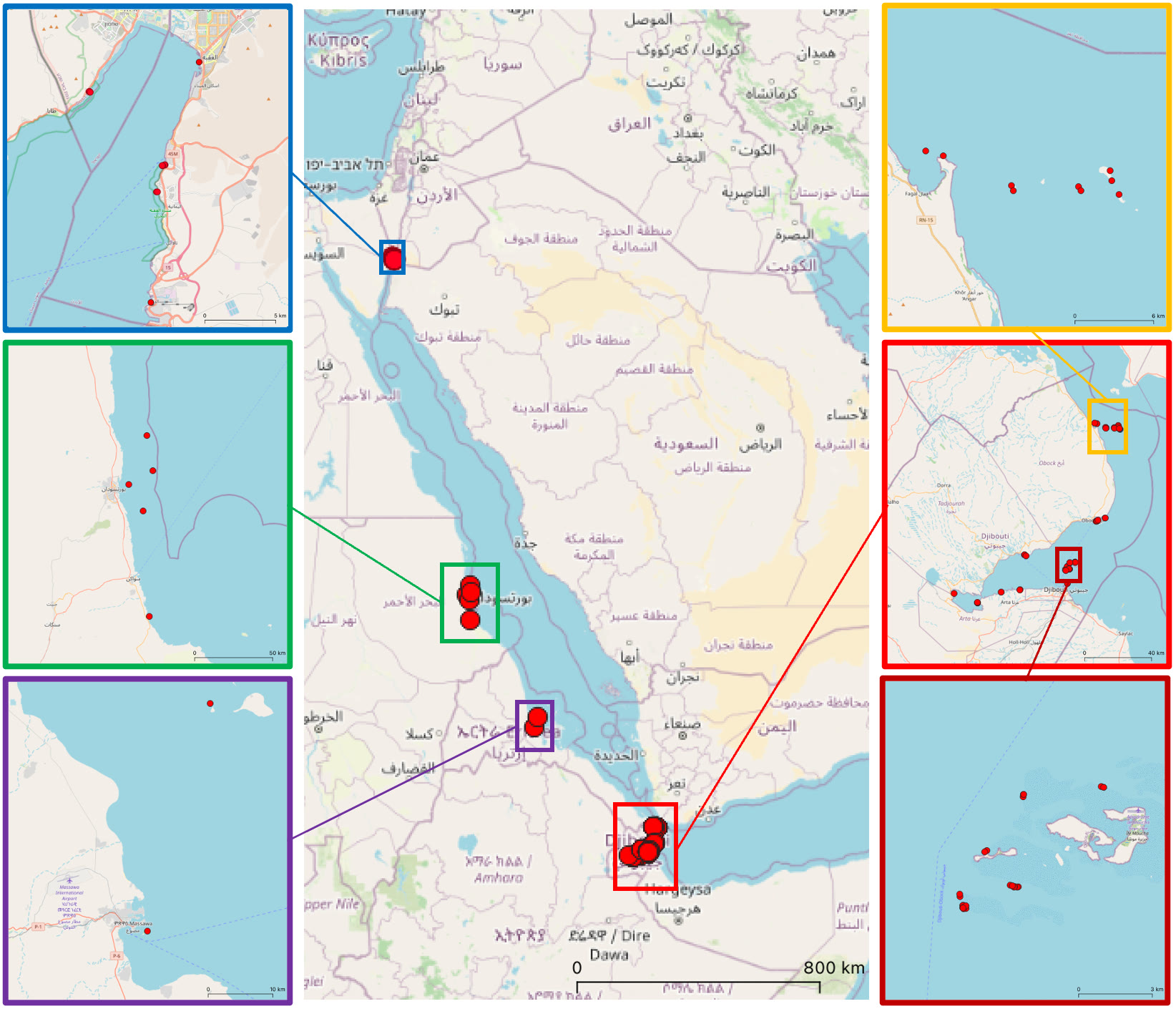}\\
    \caption{Map of all reefs from which data was collected and annotated as part of Coralscapes.}
    \label{fig:map_sites}
\end{figure}

Here, we give further information about the Coralscapes dataset and the annotated classes. A map of all reefs from which annotated data was collected is shown in Fig.~\ref{fig:map_sites}. All videos were collected in reefs that are shallower than 15m, with the vast majority of images taken at a depth of around 5-7m. The distance from the seabed varies between roughly 50cm and 5m, with the vast majority of images taken at a surface distance of 1-2 meters. The images include diverse scenes from healthy reefs (approx. 70\% live coral cover) to devastated reefs with 0\% live coral cover, and include varying turbidity conditions (visibility from approx. 20m to 1.5m). Imagery was collected at all times of the day between just after sunrise to just before sunset in sunny and cloudy conditions. No artificial illumination or physical red filter was included. 

\begin{figure}[h]
    \centering
\includegraphics[width=0.99\textwidth]{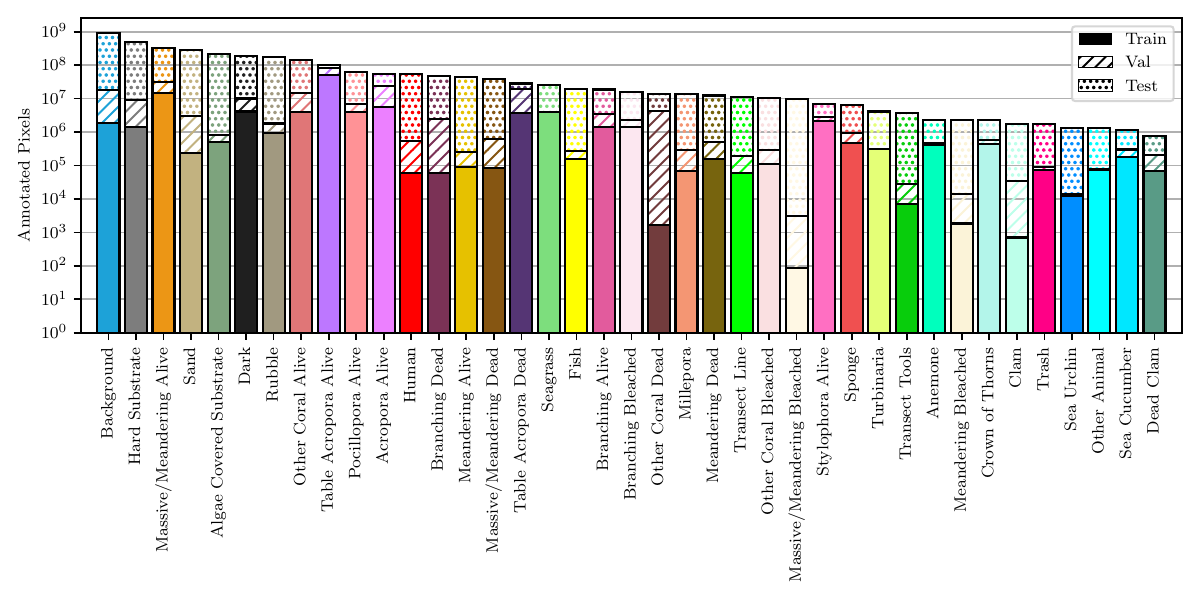}\\
    \caption{Number of annotated pixels by class on a log-scale (with linear proportions for train/val/test).}
    \label{fig:pixel_counts}
\end{figure}

\begin{figure}[h]
    \centering
\includegraphics[width=0.99\textwidth]{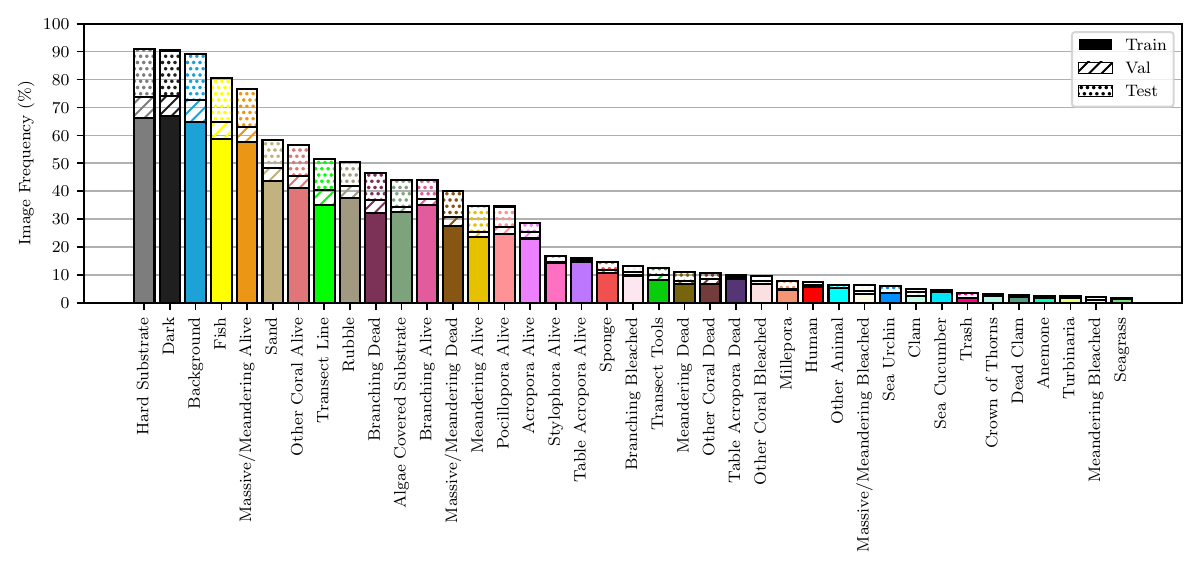}
    \caption{Percentage of images with each class present.}
    \label{fig:image_freq}
\end{figure}

We show the number of annotated pixels per class in Fig.~\ref{fig:pixel_counts}, and the image frequency of each classes (on what percentage of images this class is present) in Fig.~\ref{fig:image_freq}. All main reef-building coral types found in the Red Sea are represented in Coralscapes. Even though all the classes of Coralscapes can be found across the entire Indo-Pacific, there are morphotypes, species or genera falling into each class that are not adequately represented within Coralscapes as they do not (or rarely) appear in the Red Sea (e.g. Pocillopora grandis within the `pocillopora' class or Xestospongia testudinaria within `sponge'). 

\clearpage

\begin{figure}[h]
    \centering
\includegraphics[trim={150px 20px 150px 20px},clip, width=0.99\textwidth]{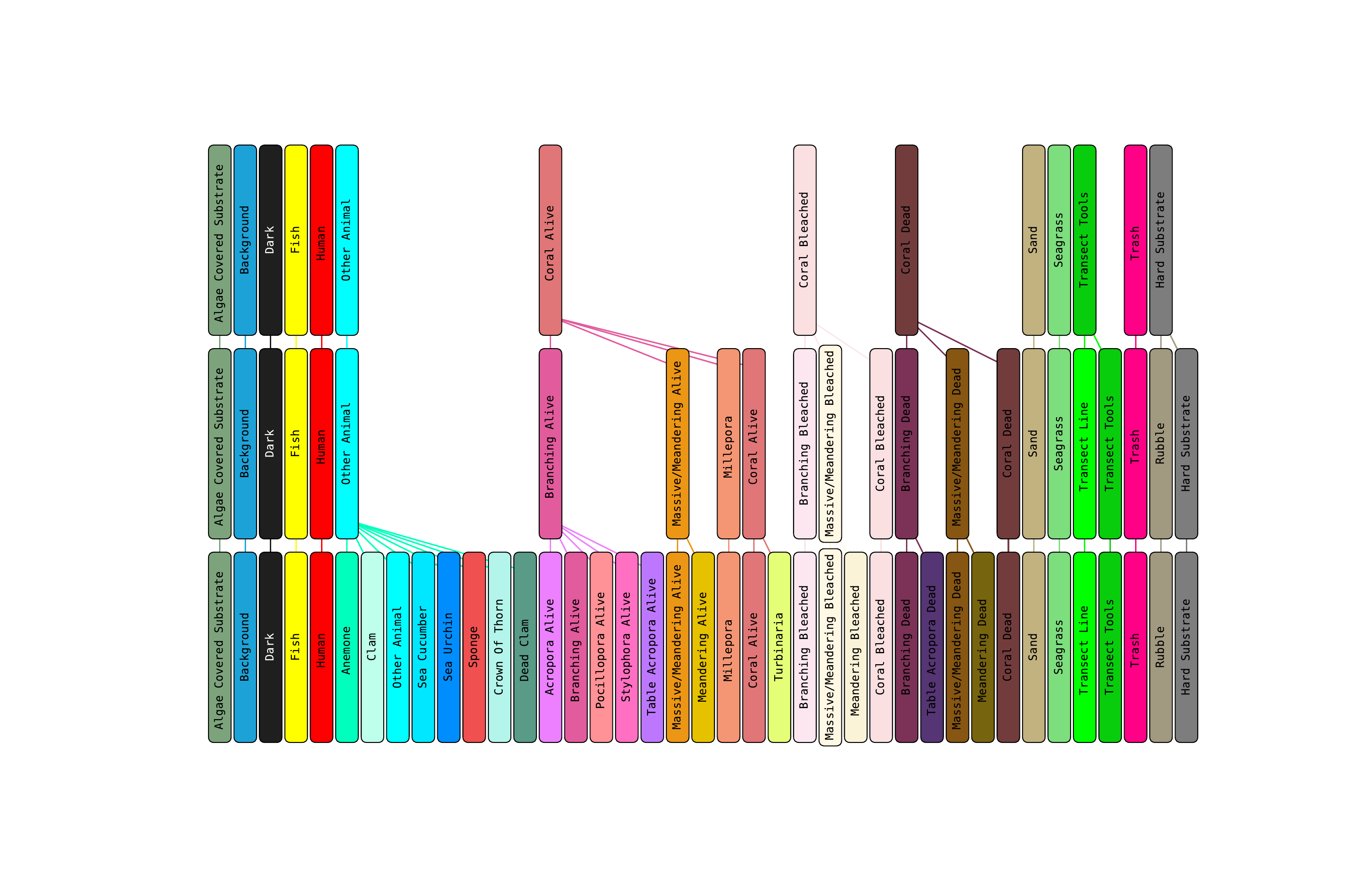}\\
    \caption{Proposed label hierarchy for summarization.}
    \label{fig:tree}
\end{figure}

Choosing the best resolution for a semantic segmentation dataset is challenging, as the label set should capture all classes of interest while at the same time having enough samples of each class to be able to accurately assess the performance of a machine learning system. In Figure~\ref{fig:tree}, we propose a hierarchical scheme for summarizing the label set into coarser classes.

\clearpage

\subsection{Annotation Class Guide}
\label{appendix:class_guide}
\paragraph{Table Acropora Alive} \fcolorbox{black}{table_acropora_alive_color}{\phantom{X}} Acropora that grow in a tabular growth form.\\
    \includegraphics[trim={150px 20px 150px 20px},clip,width=0.99\columnwidth]{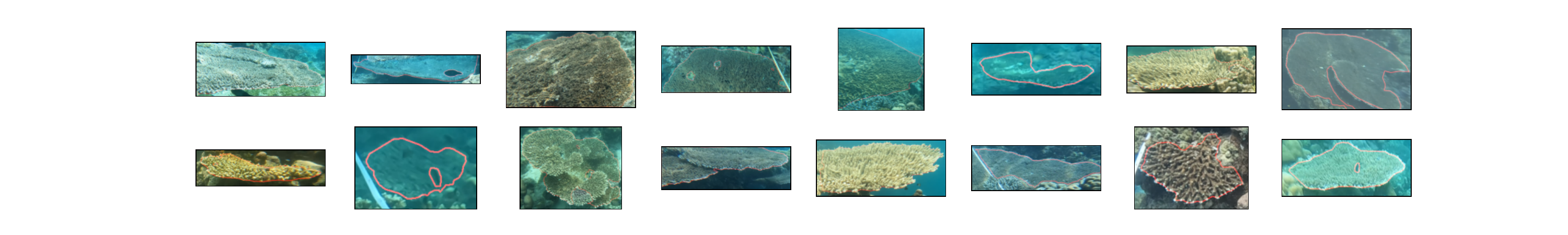}\\
\paragraph{Acropora Alive} \fcolorbox{black}{acropora_alive_color}{\phantom{X}} Acropora that do not grow in tabular growth form.\\
\includegraphics[trim={150px 20px 150px 20px},clip,width=0.99\columnwidth]{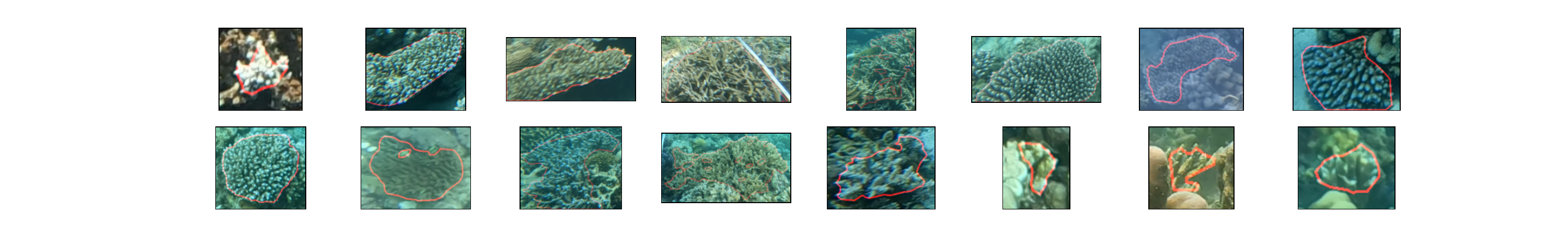}\\
\paragraph{Stylophora Alive} \fcolorbox{black}{stylophora_alive_color}{\phantom{X}} Clearly identifiable Stylophora. \\
\includegraphics[trim={150px 20px 150px 20px},clip,width=0.99\columnwidth]{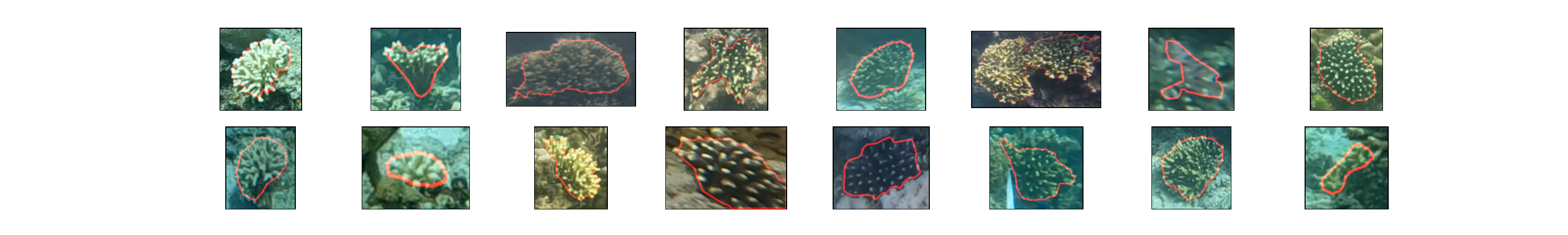}\\
\paragraph{Pocillopora Alive} \fcolorbox{black}{pocillopora_alive_color}{\phantom{X}} Clearly identifiable Pocillopora\\
\includegraphics[trim={150px 20px 150px 20px},clip,width=0.99\columnwidth]{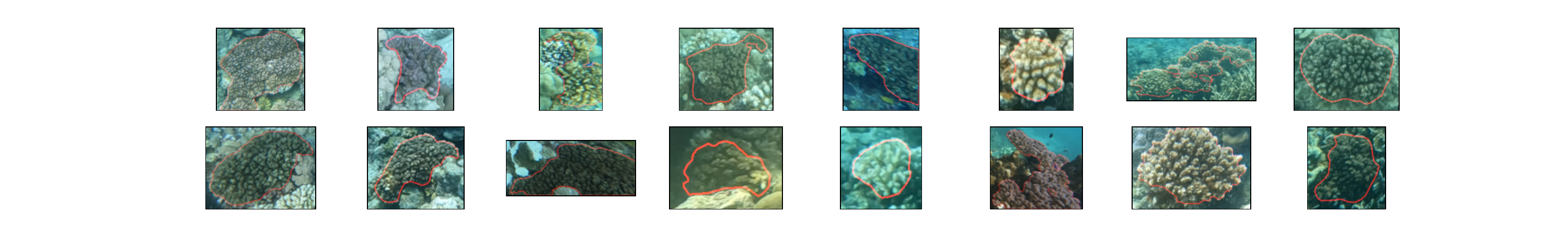}\\
\paragraph{Branching Alive} \fcolorbox{black}{branching_alive_color}{\phantom{X}} Branching corals that can surely be determined to be alive, but do not fit in to `table acropora alive`, `acropora`, `stylophora', or `pocillopora' because they are from a different genus or appear slightly blurred in the context. \\
\includegraphics[trim={150px 20px 150px 20px},clip,width=0.99\columnwidth]{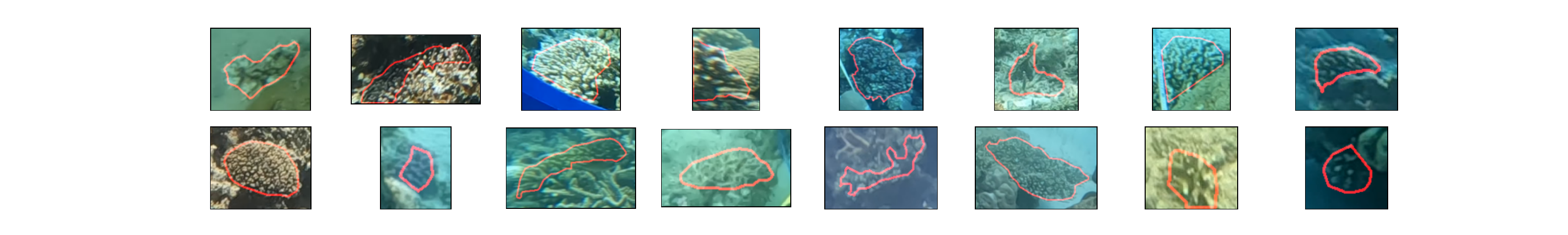}\\
\paragraph{Turbinaria} \fcolorbox{black}{turbinaria_color}{\phantom{X}} Colloqually the `scroll' coral. Does not include the macroalgae genus Turbinaria.\\
    \includegraphics[trim={150px 20px 150px 20px},clip,width=0.99\columnwidth]{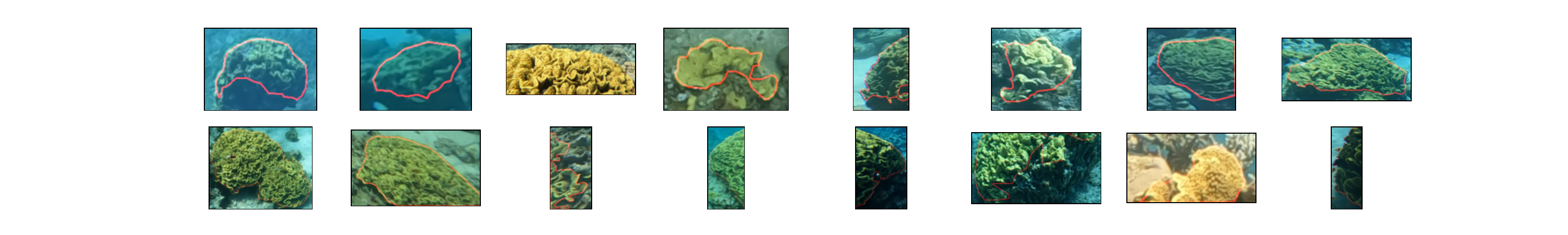}\\
\paragraph{Meandering Alive} \fcolorbox{black}{meandering_alive_color}{\phantom{X}} Corals of a meandering growth form. Includes Platygyra, Lobophyllia, Symphyllia. \\
    \includegraphics[trim={150px 20px 150px 20px},clip,width=0.99\columnwidth]{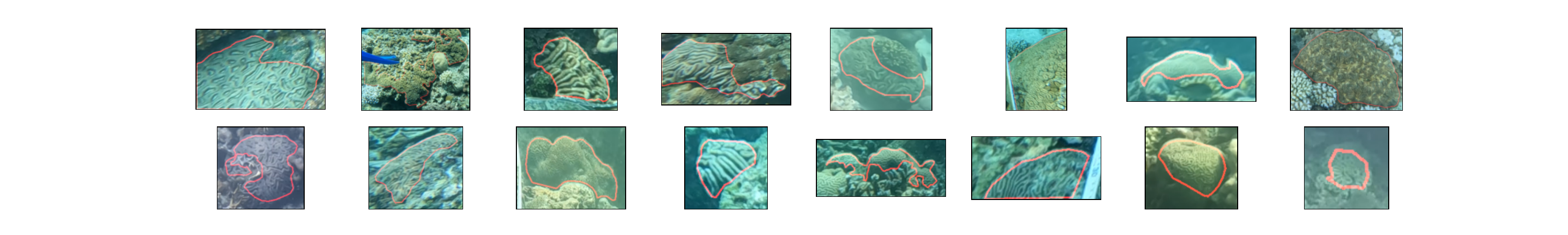}\\
\paragraph{Massive/Meandering Alive} \fcolorbox{black}{massive_meandering_alive_color}{\phantom{X}} Corals in a massive growth form. Prominently includes Porites, Favia, Favites, and many others. Includes some likely meandering corals (like Platygyra) where the meandering structure can not be clearly identified.\\
    \includegraphics[trim={150px 20px 150px 20px},clip,width=0.99\columnwidth]{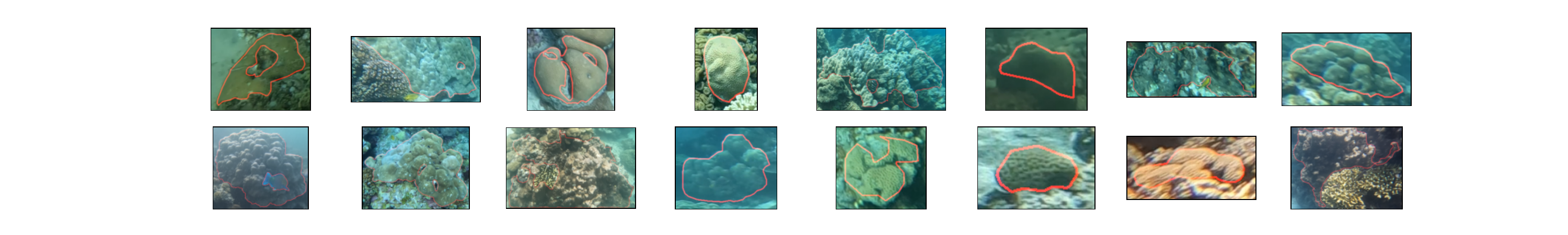}\\
\paragraph{Millepora} \fcolorbox{black}{millepora_color}{\phantom{X}} The `fire coral' Millepora is technically not a coral, but a hydrozoan. Appears most commonly in a branching form (Millepora Dichotoma).\\
    \includegraphics[trim={150px 20px 150px 20px},clip,width=0.99\columnwidth]{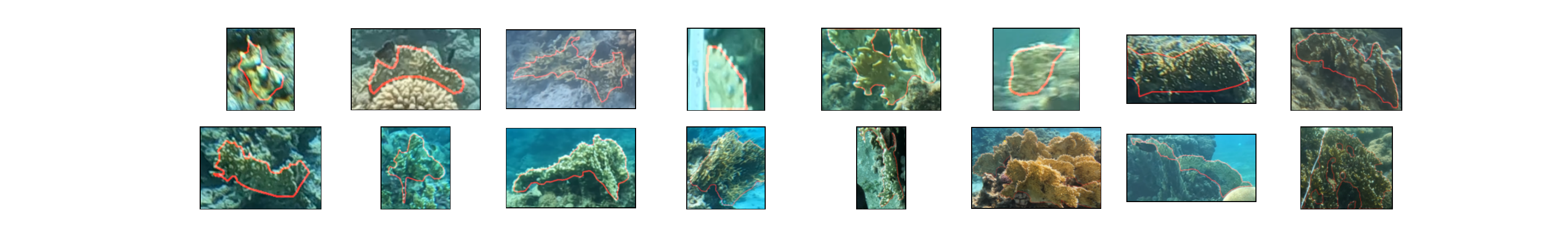}\\
    \paragraph{Other Coral Alive} \fcolorbox{black}{other_coral_alive_color}{\phantom{X}} All other live corals. Includes corals in thin plate or encrusting growth form, soft corals, and corals that can not be clearly classified into the other classes. \\
    \includegraphics[trim={150px 20px 150px 20px},clip,width=0.99\columnwidth]{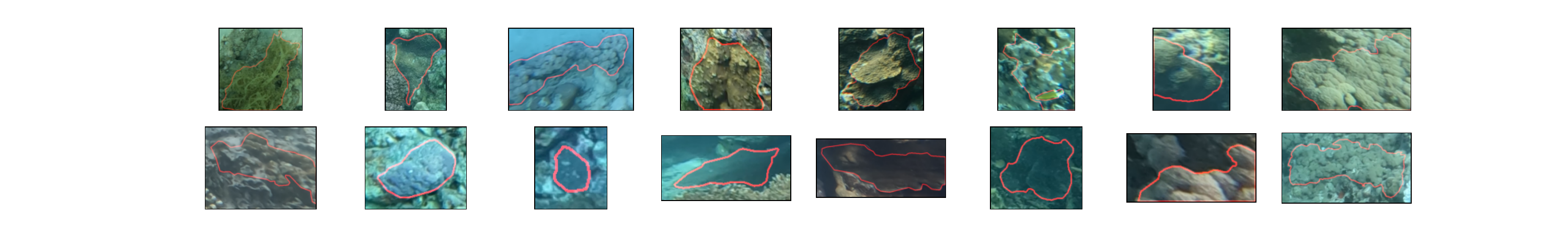}\\
\paragraph{Table Acropora Dead} \fcolorbox{black}{table_acropora_dead_color}{\phantom{X}} Dead Acropora tables which are dead or collapsed but are not yet overgrown by algae. Parts visibly overgrown by algae are labeled as `algae covered substrate'.\\
    \includegraphics[trim={150px 20px 150px 20px},clip,width=0.99\columnwidth]{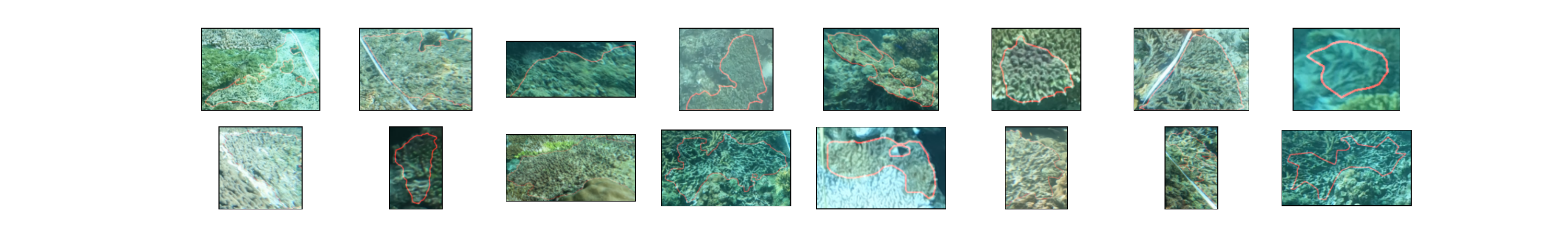}\\
\paragraph{Branching Dead} \fcolorbox{black}{branching_dead_color}{\phantom{X}} Other dead branching coral, including acropora, pocillopora, and stylophora. When overgrown by algae, labeled as `algae covered substrate'.\\
    \includegraphics[trim={150px 20px 150px 20px},clip,width=0.99\columnwidth]{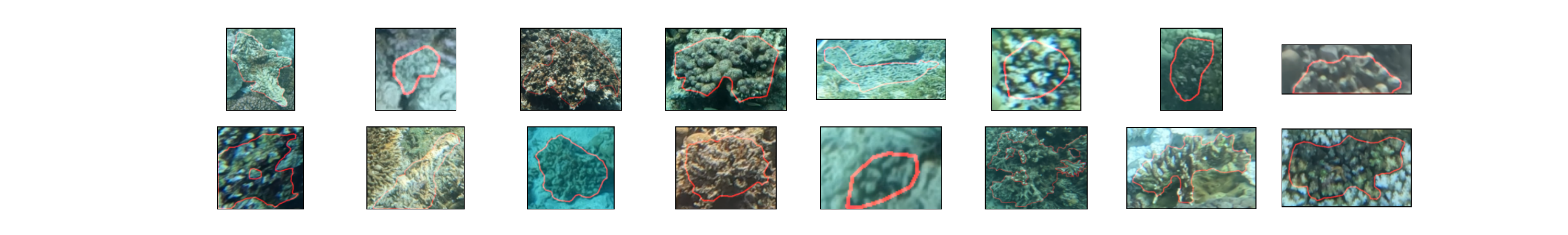}\\
\paragraph{Meandering Dead} \fcolorbox{black}{meandering_dead_color}{\phantom{X}} Dead meandering corals (Lobophyllia, Symphyllia, Platygyra, etc.). \\
    \includegraphics[trim={150px 20px 150px 20px},clip,width=0.99\columnwidth]{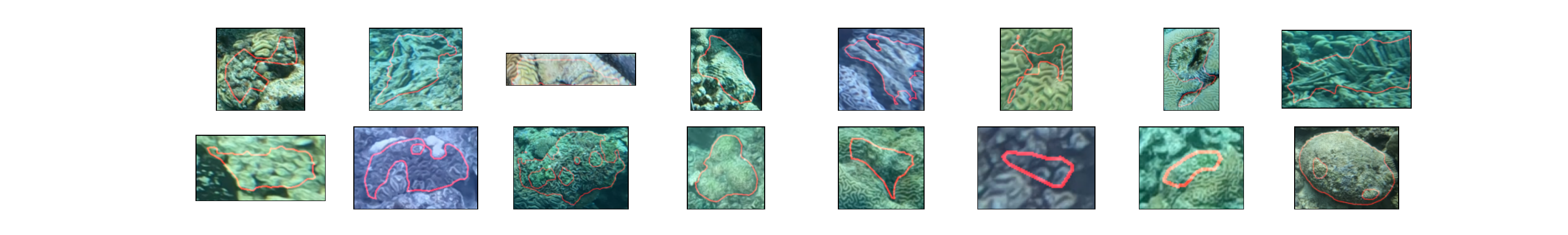}\\
\paragraph{Massive/Meandering Dead} \fcolorbox{black}{massive_meandering_dead_color}{\phantom{X}}  Dead massive corals or meandering corals that have decomposed enough so that the meandering structure is no longer visible.\\
    \includegraphics[trim={150px 20px 150px 20px},clip,width=0.99\columnwidth]{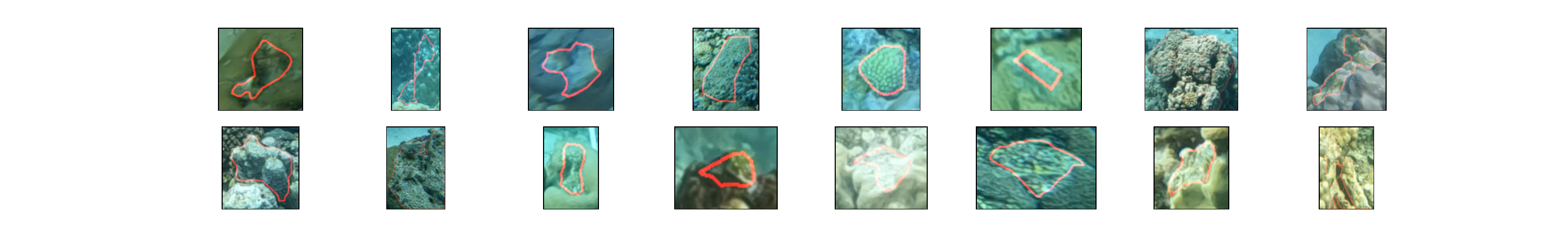}\\
    \paragraph{Other Coral Dead} \fcolorbox{black}{other_coral_dead_color}{\phantom{X}} All other dead coral skeletons. \\
    \includegraphics[trim={150px 20px 150px 20px},clip,width=0.99\columnwidth]{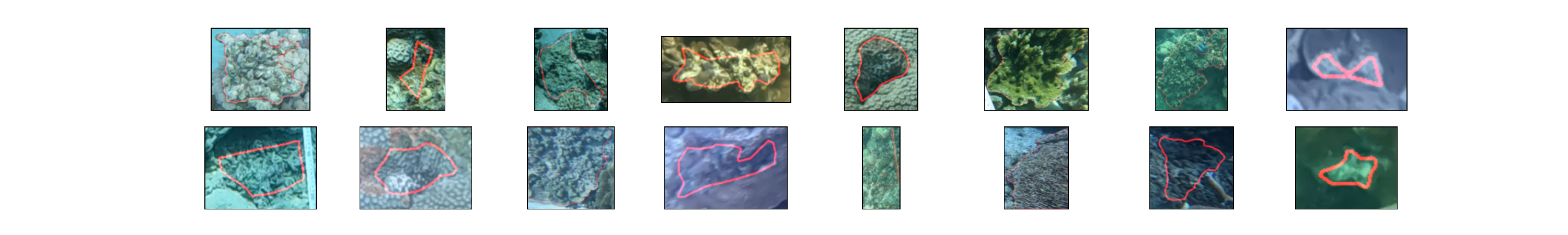}\\

\paragraph{Branching Bleached} \fcolorbox{black}{branching_bleached_color}{\phantom{X}} Bleached branching coral of all kinds, including bleached table acropora.\\
    \includegraphics[trim={150px 20px 150px 20px},clip,width=0.99\columnwidth]{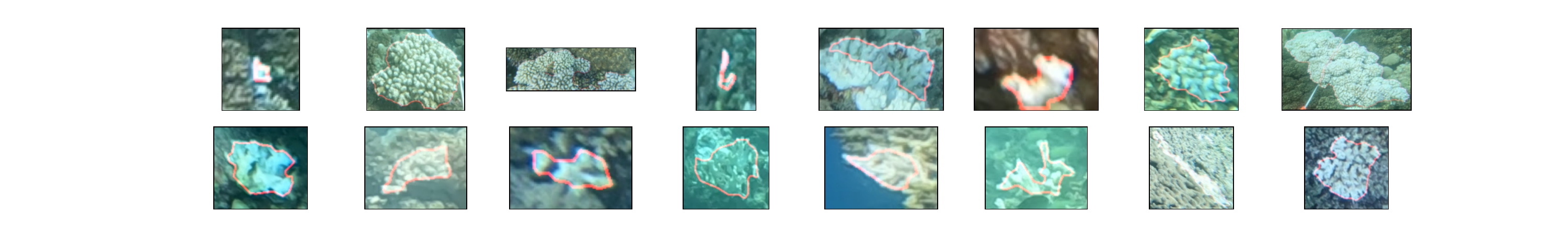}\\

\paragraph{Meandering Bleached} \fcolorbox{black}{meandering_bleached_color}{\phantom{X}} Bleached meandering corals (Lobophyllia, Symphyllia, Platygyra, etc.).\\
    \includegraphics[trim={150px 20px 150px 20px},clip,width=0.99\columnwidth]{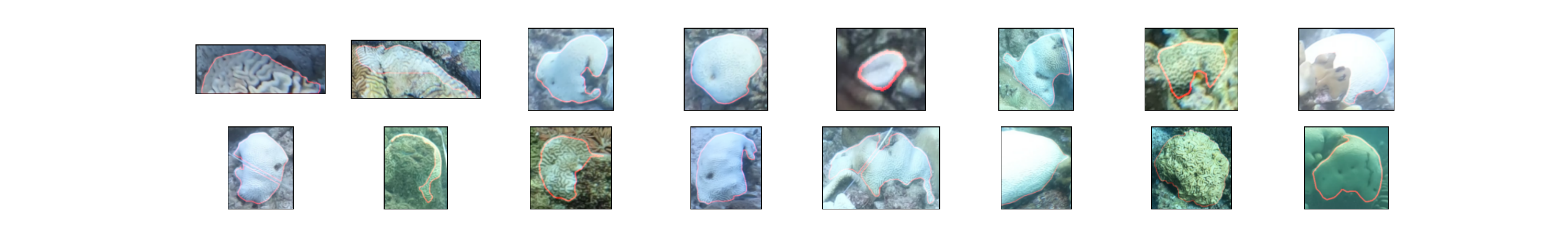}\\
    
\paragraph{Massive/Meandering Bleached} \fcolorbox{black}{massive_meandering_bleached_color}{\phantom{X}} Bleached massive corals or meandering corals that have decomposed enough so that the meandering structure is no longer visible.\\
    \includegraphics[trim={150px 20px 150px 20px},clip,width=0.99\columnwidth]{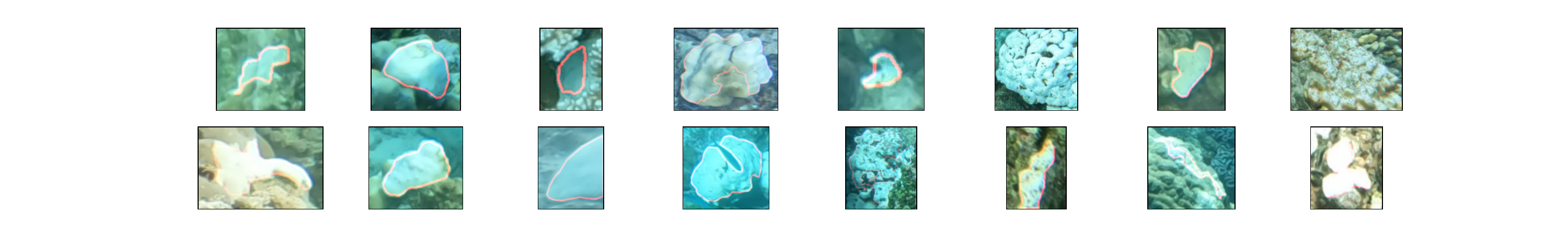}\\
    
\paragraph{Other Coral Bleached} \fcolorbox{black}{other_coral_bleached_color}{\phantom{X}} All other bleached coral, including bleached soft coral.\\
    \includegraphics[trim={150px 20px 150px 20px},clip,width=0.99\columnwidth]{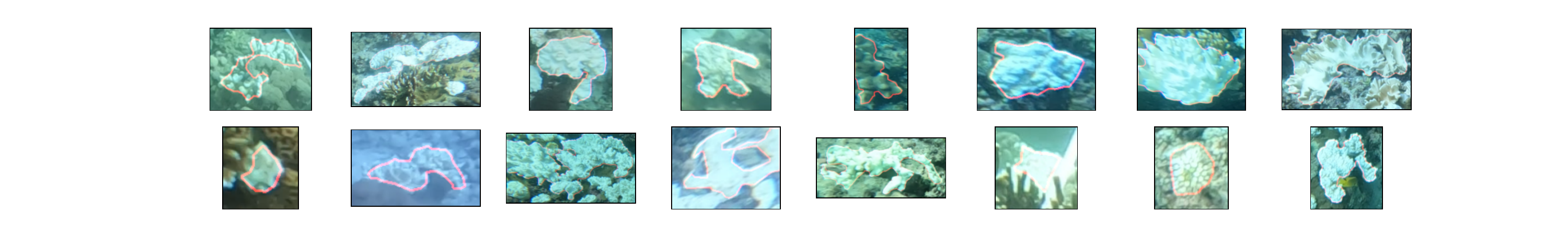}\\

\paragraph{Sponge} \fcolorbox{black}{sponge_color}{\phantom{X}} Sponges of all kind.\\
    \includegraphics[trim={150px 20px 150px 20px},clip,width=0.99\columnwidth]{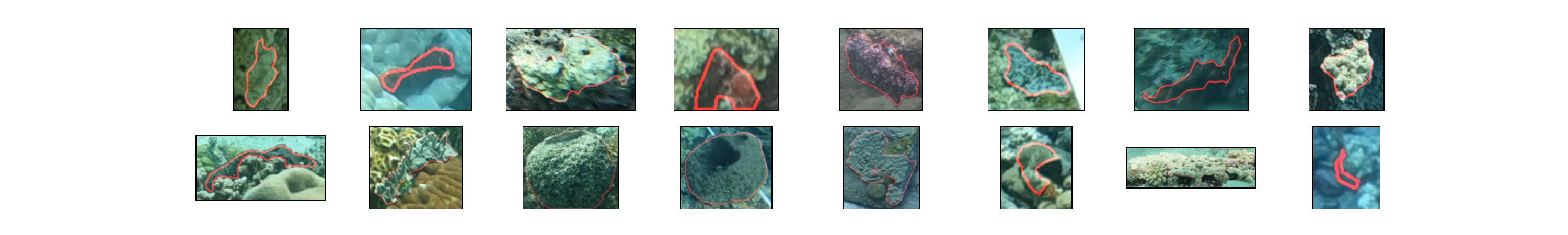}\\

\paragraph{Crown of Thorns} \fcolorbox{black}{crown_of_thorn_color}{\phantom{X}} Acanthaster planci, known to cause outbreaks in which it can severely reduce coral cover. \\
    \includegraphics[trim={150px 20px 150px 20px},clip,width=0.99\columnwidth]{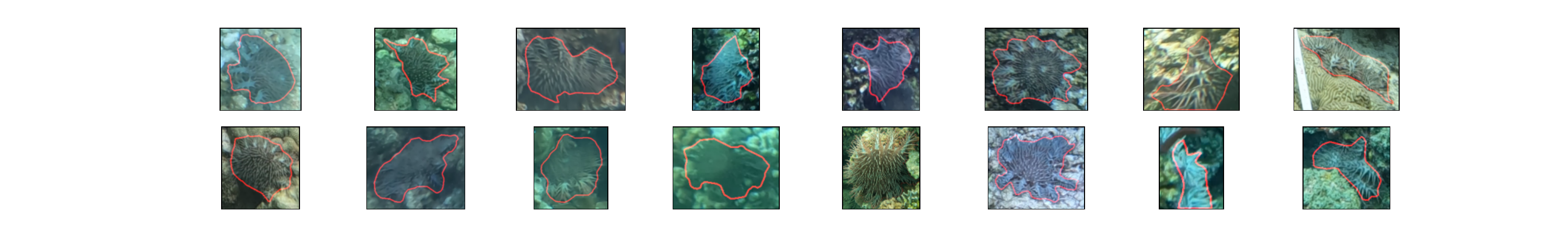}\\
\paragraph{Sea Urchin} \fcolorbox{black}{sea_urchin_color}{\phantom{X}} \\
    \includegraphics[trim={150px 20px 150px 20px},clip,width=0.99\columnwidth]{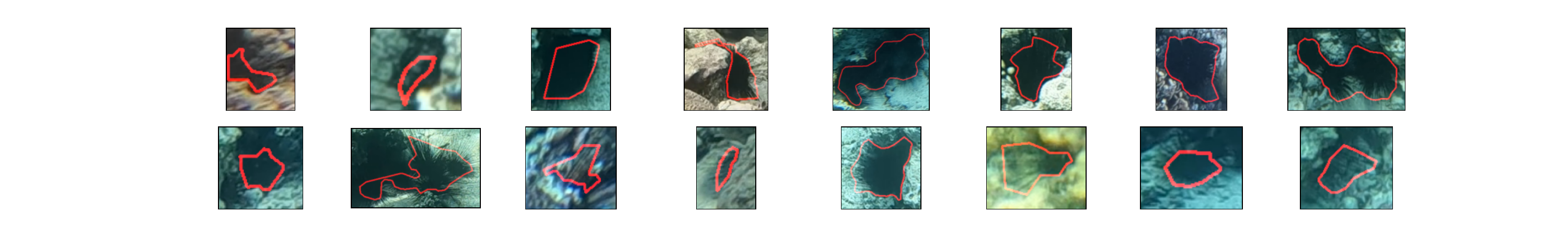}\\
\paragraph{Sea Cucumber} \fcolorbox{black}{sea_cucumber_color}{\phantom{X}} \\
    \includegraphics[trim={150px 20px 150px 20px},clip,width=0.99\columnwidth]{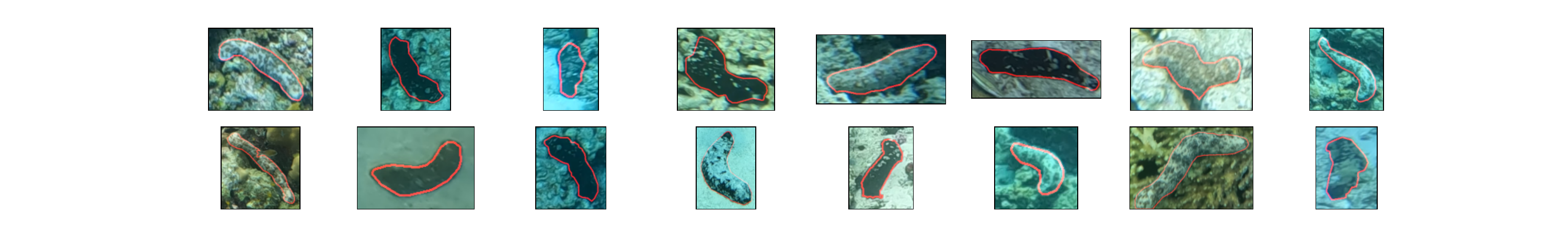}\\
\paragraph{Anemone} \fcolorbox{black}{anemone_color}{\phantom{X}} \\
    \includegraphics[trim={150px 20px 150px 20px},clip,width=0.99\columnwidth]{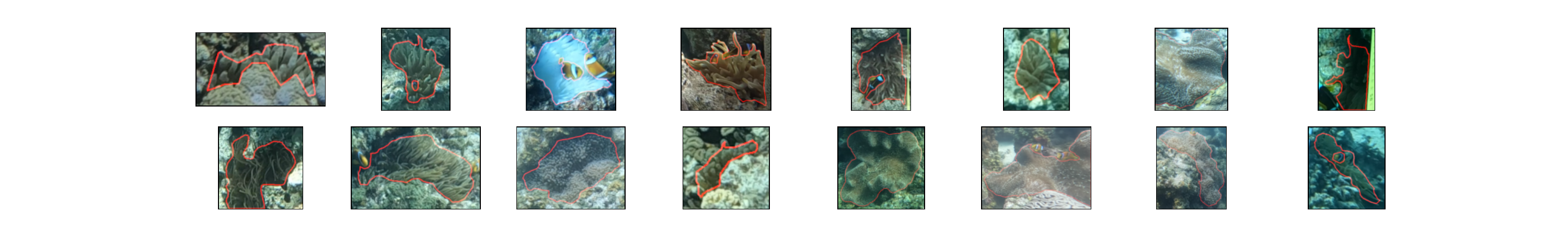}\\
\paragraph{Clam} \fcolorbox{black}{clam_color}{\phantom{X}} Live giant clams.\\
    \includegraphics[trim={150px 20px 150px 20px},clip,width=0.99\columnwidth]{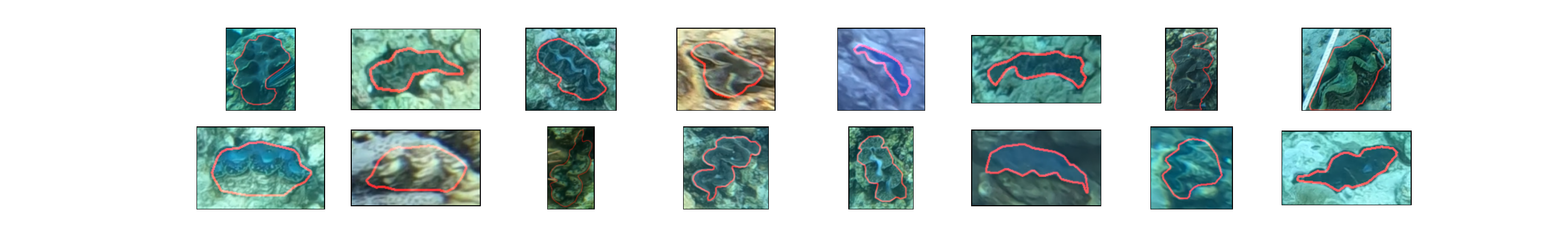}\\
\paragraph{Dead Clam} \fcolorbox{black}{dead_clam_color}{\phantom{X}} Dead giant clams.\\
    \includegraphics[trim={150px 20px 150px 20px},clip,width=0.99\columnwidth]{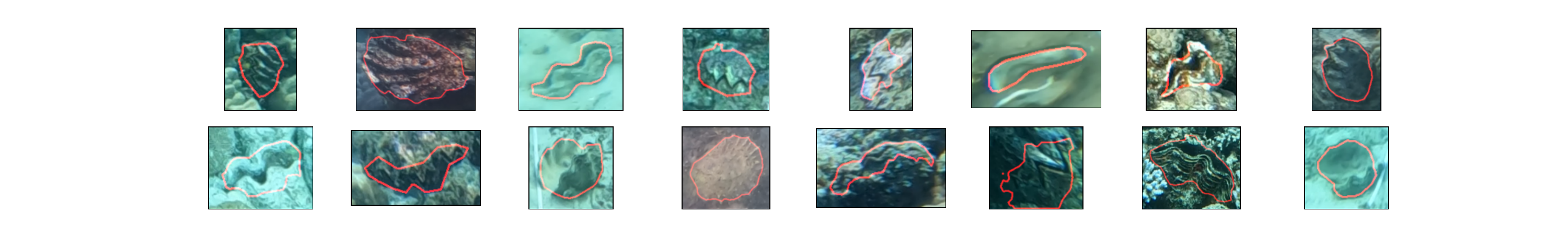}\\
    
\paragraph{Other Animal} \fcolorbox{black}{other_animal_color}{\phantom{X}} Includes starfish (except the crown-of-thorns starfish), feather worms, sea turtles, and other non-identifiable invertebrates or animals of which there are not enough annotations to warrant a separate class.\\
    \includegraphics[trim={150px 20px 150px 20px},clip,width=0.99\columnwidth]{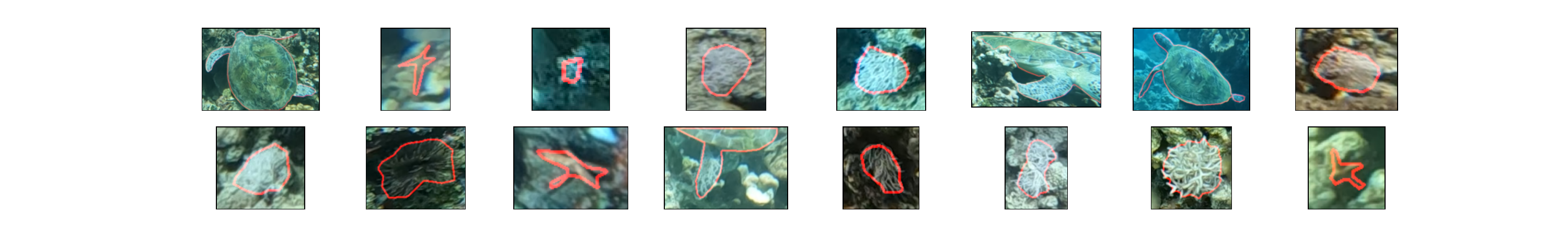}\\
\paragraph{Trash} \fcolorbox{black}{trash_color}{\phantom{X}} Includes all kinds of marine litter. Most common are plastic items including bags, cups, and bottles, aluminum cans and glass bottles, as well as abandoned fishing material, and parts of boats and machines.\\
    \includegraphics[trim={150px 20px 150px 20px},clip,width=0.99\columnwidth]{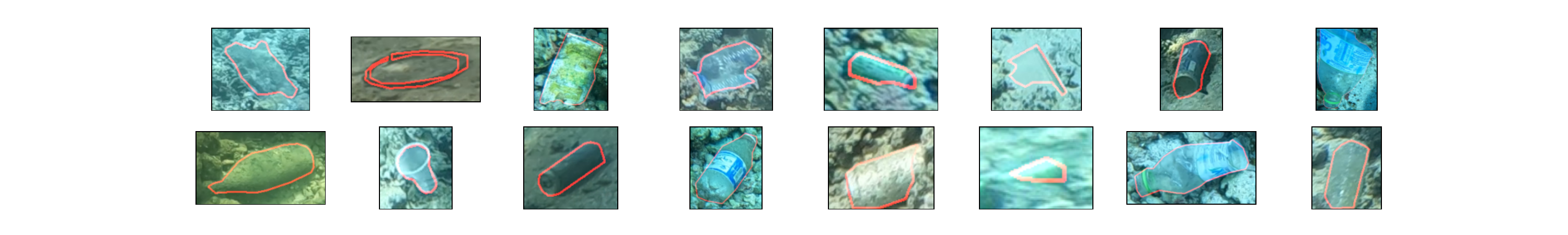}\\
\paragraph{Transect Line} \fcolorbox{black}{transect_line_color}{\phantom{X}} Rolled out transect tape. Excludes reels or other strands of rope that are laid out.\\
    \includegraphics[trim={150px 20px 150px 20px},clip,width=0.99\columnwidth]{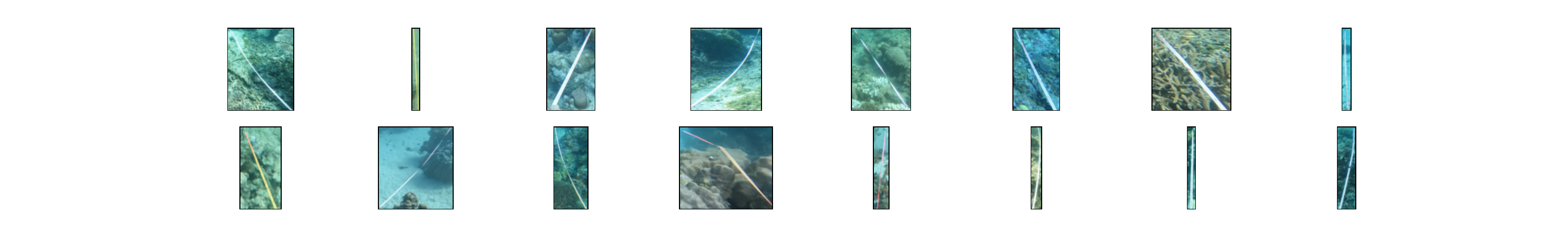}\\
\paragraph{Transect Tools} \fcolorbox{black}{transect_tools_color}{\phantom{X}} This includes transect reels \& spools, tags and markers placed on the reef, diving weights, surface marker buoys and the string attaching them to the ground or weights on the ground.\\
\includegraphics[trim={150px 20px 150px 20px},clip,width=0.99\columnwidth]{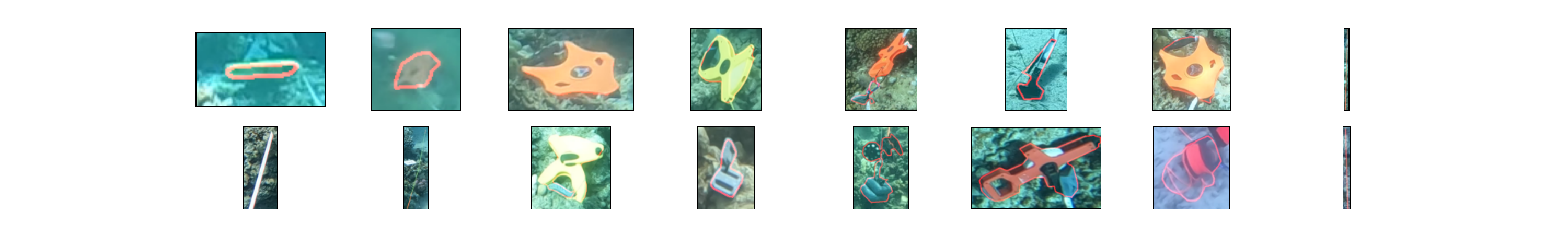}\\\paragraph{Human} \fcolorbox{black}{human_color}{\phantom{X}} The human class includes divers and snorkelers, of which sometimes only a hand or a fin is in the frame. Some ambiguity can arise when a human carries a transect tool that is not laid out on the ground, like a transect reel. In these cases, we generally decide this tool becomes part of the human polygon.\\
    \includegraphics[trim={150px 20px 150px 20px},clip,width=0.99\columnwidth]{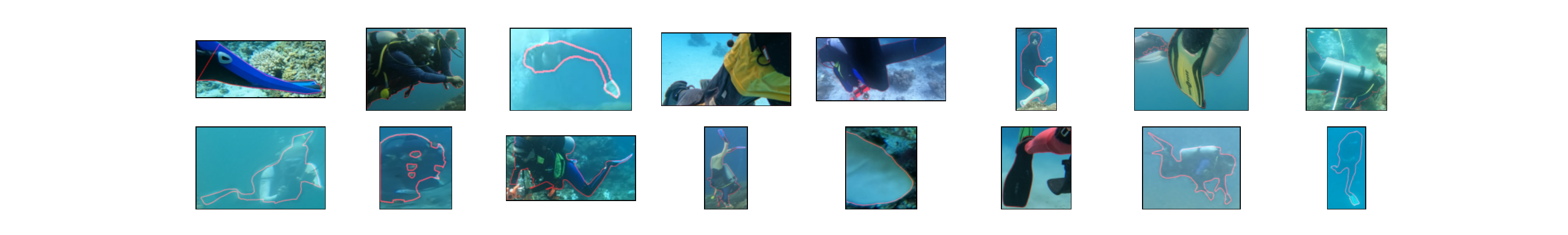}\\
\paragraph{Fish} \fcolorbox{black}{fish_color}{\phantom{X}} Fish of all kinds. \\
    \includegraphics[trim={150px 20px 150px 20px},clip,width=0.99\columnwidth]{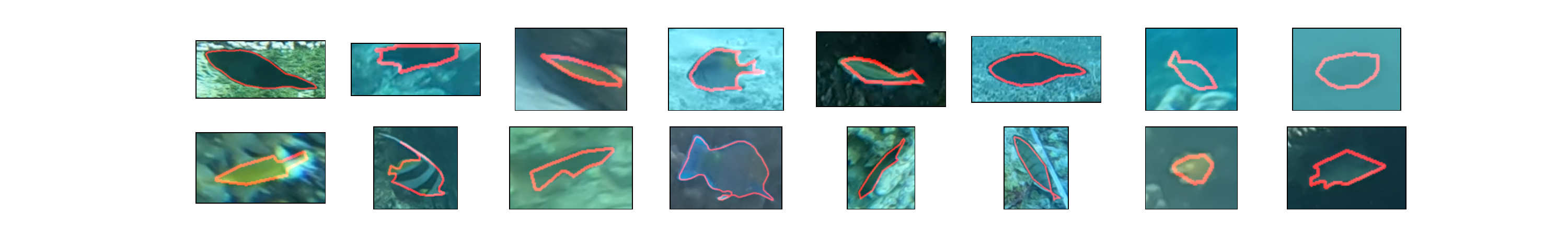}\\
\paragraph{Background} \fcolorbox{black}{background_color}{\phantom{X}} This class is assigned to pixels that are too far away or too blurry to be classifiable into any other class. This includes the water surface, which is sometimes visible. If Coralscapes is used in the context of benthic mapping, we recommend masking or excluding background. \\
    \includegraphics[trim={150px 20px 150px 20px},clip,width=0.99\columnwidth]{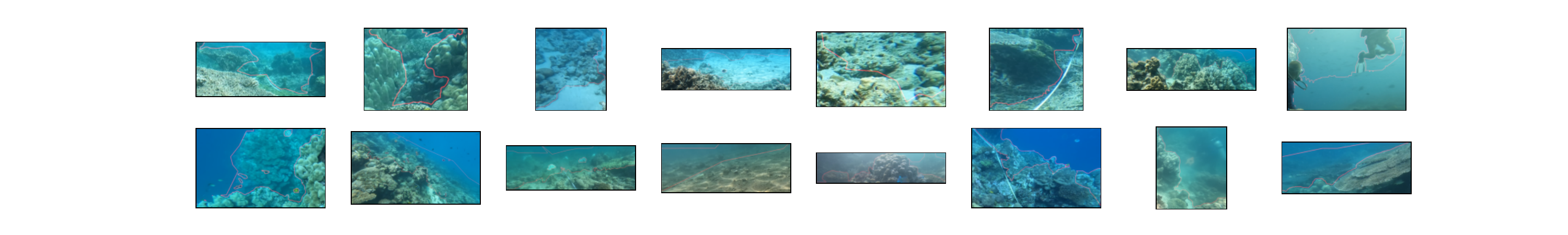}\\
\paragraph{Dark} \fcolorbox{black}{dark_color}{\phantom{X}} Parts of the image that are too dark to discern the benthic class, but that are within the substrate. Includes holes, crevices, cracks, or dark shadows cast onto the substrate. If Coralscapes is used in the context of benthic mapping, dark areas should be included in the outputs.\\
    \includegraphics[trim={150px 20px 150px 20px},clip,width=0.99\columnwidth]{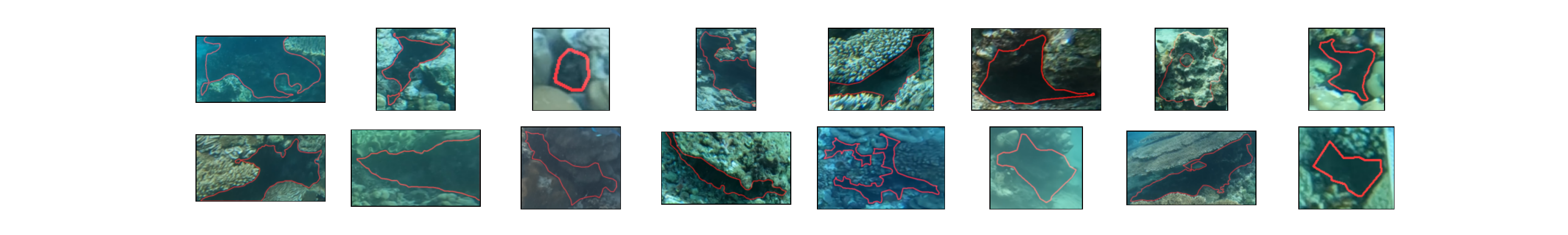}\\    
    \FloatBarrier
\paragraph{Sand} \fcolorbox{black}{sand_color}{\phantom{X}} Loose, fine sand. \\
    \includegraphics[trim={150px 20px 150px 20px},clip,width=0.99\columnwidth]{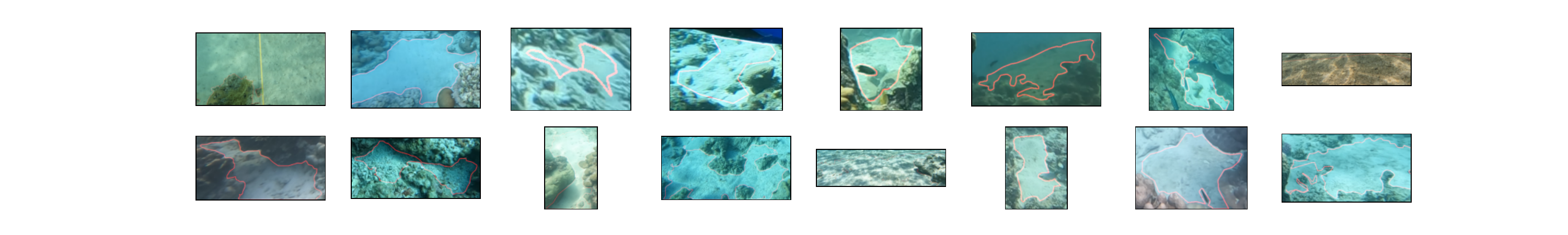}\\
\paragraph{Rubble} \fcolorbox{black}{rubble_color}{\phantom{X}} Small loose fragments of rocky substrate or dead coral. \\
    \includegraphics[trim={150px 20px 150px 20px},clip,width=0.99\columnwidth]{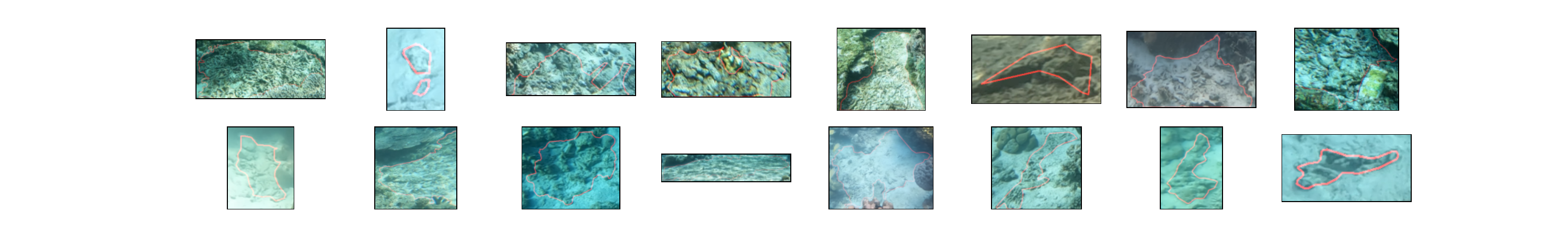}\\
\paragraph{Algae Covered Substrate} \fcolorbox{black}{algae_covered_substrate_color}{\phantom{X}} Substrate covered in turf algae or other macroalgae, including fleshy algae and Turbinaria.\\
    \includegraphics[trim={150px 20px 150px 20px},clip,width=0.99\columnwidth]{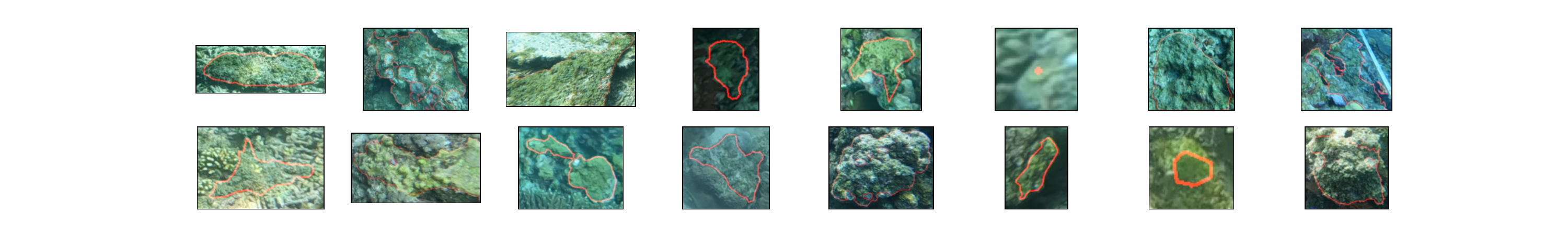}\\
\paragraph{Hard Substrate} \fcolorbox{black}{unknown_hard_substrate_color}{\phantom{X}} Hard substrate that is part of the reef, which can not be identified into any of the other classes. Includes rocks and heavily decomposed coral skeletons. Also includes human-made structures (underwater infrastructure) such as pipes, pier columns, coral nursery tables, buoys and their lines, as well as boat anchors. \\
    \includegraphics[trim={150px 20px 150px 20px},clip,width=0.99\columnwidth]{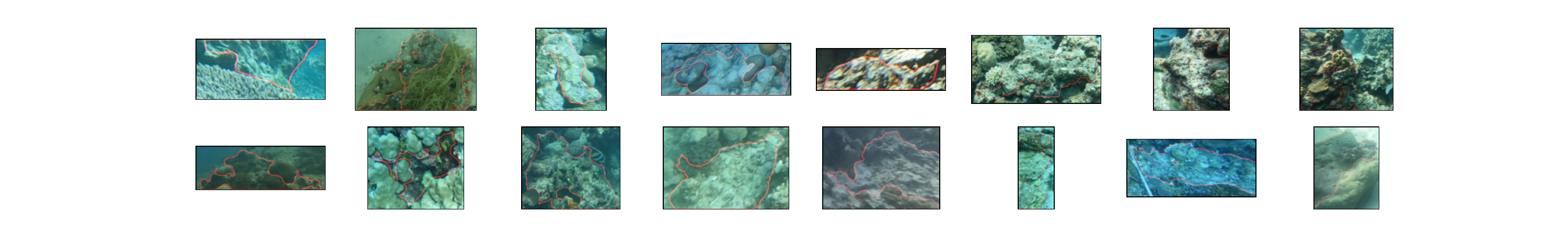}\\

\paragraph{Seagrass} \fcolorbox{black}{seagrass_color}{\phantom{X}} \\
    \includegraphics[trim={150px 20px 150px 20px},clip,width=0.99\columnwidth]{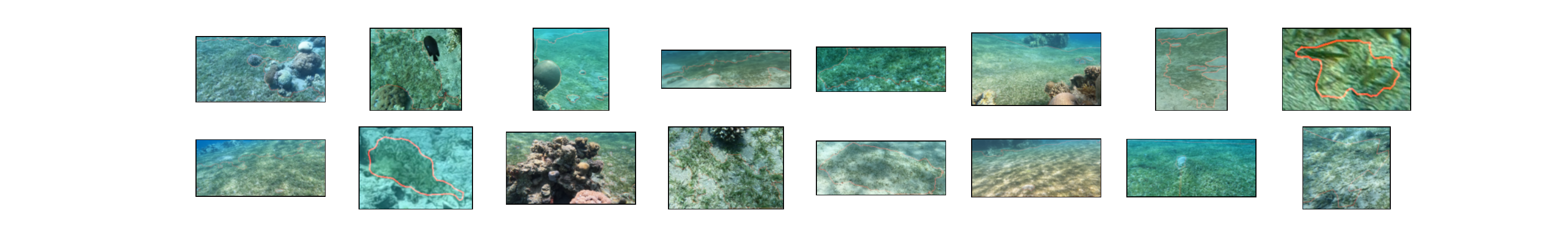}\\

\newpage

\section{Appendix: Implementation Details}
\label{appendix:implementation_details}
\subsection{Benchmarking}

All models are implemented in PyTorch, with convolutional architectures (UNet++, DeepLabV3+) built using the segmentation-models-pytorch library \cite{Iakubovskii:2019} and transformer-based architectures (SegFormer, DPT) implemented using the Hugging Face transformers library \cite{wolf2019huggingface}. Parameter efficient fine-tuning with LoRA was done using the Hugging Face PEFT library \cite{peft}. Training and evaluation are conducted on a system with a h100 GPU. The code used to train and evaluate the models is available on GitHub\footnote{\tiny{\url{https://github.com/eceo-epfl/coralscapesScripts}}}.

For the DeepLabV3+ model with a ResNet-50 backbone we follow the training strategy of \cite{deeplabv3plus}, training for 1000 epochs  with a batch size of 16 using stochastic gradient descent (SGD) with an initial learning rate of 1e-3, momentum of 0.9, weight decay of 1e-4, and a polynomial learning rate scheduler with a power of 0.9. The model is optimized using a cross-entropy loss. During training, images are randomly scaled within a range of 0.75 and 2, flipped horizontally with a 0.5 probability and randomly cropped to 768×768 pixels. All input images are then normalized using the ImageNet mean and standard deviation. When training on the train+validation set and evaluating on the test set, we trained the model for 700 epochs based on the optimal validation set mIoU.

For the UNet++ model with a ResNet-50 backbone we follow the training strategy of \cite{zhou2018unet++}, training for 1000 epochs with a batch size of 16 using stochastic gradient descent (SGD) with an initial learning rate of 1e-2, momentum of 0.99, weight decay of 3e-5 and polynomial learning rate scheduler with a power of 0.9. The model uses a combination of cross-entropy and Dice loss, with equal weighting. During training, images are randomly cropped to 512×512 pixels and flipped horizontally with a 0.5 probability. Input images are normalized using the ImageNet mean and standard deviation. When training on the train+validation set and evaluating on the test set, we trained the model for 925 epochs based on the optimal validation set mIoU.

For the SegFormer models, training is conducted following \cite{xie2021segformer}, using a batch size of 8 when using the MiT-b2 backbone and 4 when using the MiT-b5 backbone for 1000 epochs, using the AdamW optimizer with an initial learning rate of 6e-5 (multiplied by 10 when using LoRA), weight decay of 1e-2 and polynomial learning rate scheduler with a power of 1. During training, images are randomly scaled within a range of 1 and 2, flipped horizontally with a 0.5 probability and randomly cropped to 1024×1024 pixels. Input images are normalized using the ImageNet mean and standard deviation. For evaluation, a non-overlapping sliding window strategy is employed, using a window size of 1024x1024 and a stride of 1024. When performing parameter efficient fine-tuning with LoRA, a rank size and alpha of 128 was used. We tried using ranks of 64 and 256, however that had no significant effect on the results. When training on the train+validation set and evaluating on the test set, we trained the model for 265 epochs for the MiT-b2 backbone, 75 epochs for the MiT-b5 backbone and 765 epochs for both backbones trained with LoRA.

For the Linear DINOv2 model we followed the implementation described in the DINOv2 paper \cite{oquab2023dinov2}. The model is trained for 1000 epochs with a batch size of 16, using the AdamW optimizer with an initial learning rate of 5e-5 (5e-6 for the backbone layers), weight decay of 1e-2, and a polynomial learning rate scheduler with a power of 1. During training, images are resized to 1036×518 pixels, randomly horizontally flipped with a 0.5 probability, and normalized using the ImageNet mean and standard deviation. For validation and testing, images are resized to the same dimensions and normalized similarly, while keeping the ground truth masks at their original size. The performance of the model is then evaluated over the interpolated prediction mask at the original dimension of the mask (2048x1024). When training on the train+validation set and evaluating on the test set, we trained the model for 155 epochs based on the optimal validation set mIoU.

For the DPT models \cite{ranftl2021vision} with a DINOv2 backbone, we followed the training procedure inspired by the Linear DINOv2 approach described above and the implementation of DepthAnythingV2 \cite{depthanythingv2}. As such the model is trained for 1000 epochs with a batch size of 16 when using the base backbone and 8 when using the giant backbone, using the AdamW optimizer with an initial learning rate of 5e-5 (5e-6 for the backbone layers; multiplied by 10 when using LoRA), weight decay of 1e-2, and a polynomial learning rate scheduler with a power of 1. During training, images are resized to 1036×518 pixels, randomly horizontally flipped with a 0.5 probability, randomly cropped to size 518x518 and normalized using the ImageNet mean and standard deviation. For validation and testing, images are resized to the same dimensions and normalized similarly, while keeping the ground truth masks at their original size, using the same performance evaluation procedure as for the Linear DINOv2 model. When training on the train+validation set and evaluating on the test set, we trained the model for 165 epochs for the DINOv2-Base backbone, 55 epochs for the DINOv2-Giant backbone and 365 epochs for both backbones trained with LoRA.

\subsection{UCSD Mosaics Transfer Learning}

\begin{figure}[h]
    \centering
    \begin{subfigure}{0.154\textwidth}
    \includegraphics[trim={80px 38px 80px 20px},clip,width=\linewidth,height=68px]{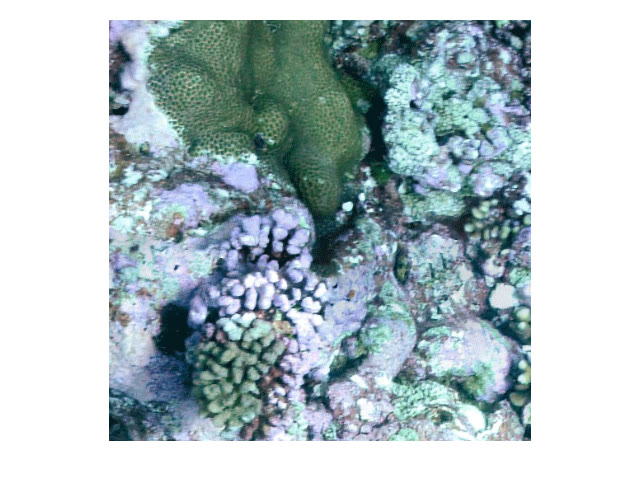}
        \caption{Image}
    \end{subfigure}
    \begin{subfigure}{0.15\textwidth}
    \includegraphics[trim={80px 20px 80px 20px},clip,width=\linewidth,height=68px]{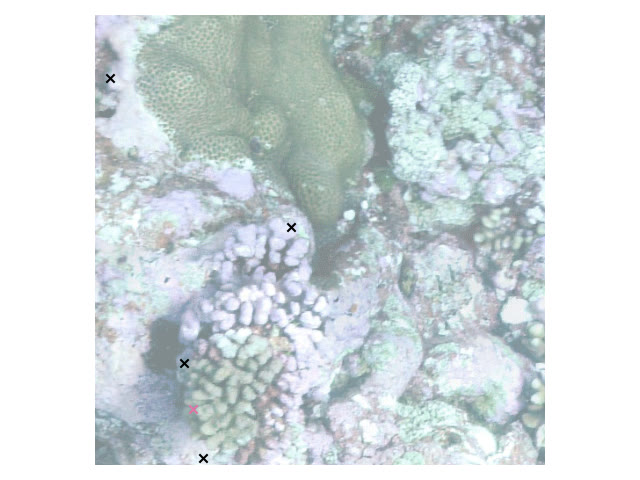}
        \caption{5}
    \end{subfigure}
    \begin{subfigure}{0.15\textwidth}
    \includegraphics[trim={80px 20px 80px 20px},clip,width=\linewidth,height=68px]{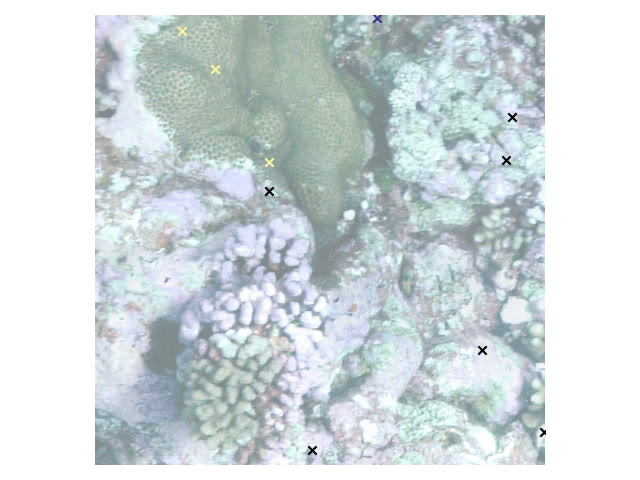}
        \caption{10}
    \end{subfigure}
    \begin{subfigure}{0.15\textwidth}
    \includegraphics[trim={80px 20px 80px 20px},clip,width=\linewidth,height=68px]{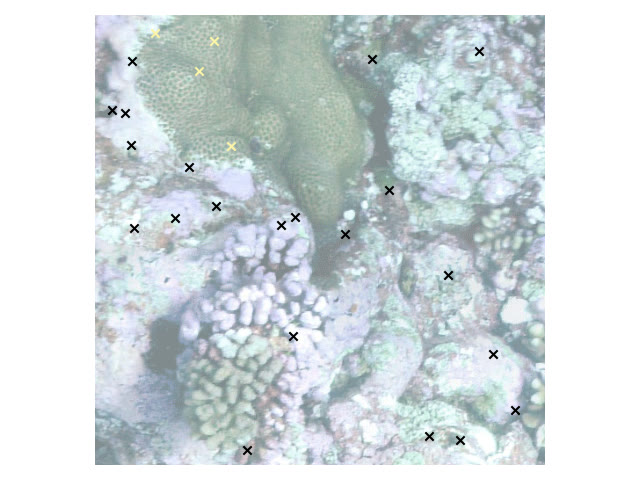}
        \caption{25}
    \end{subfigure}
    \begin{subfigure}{0.15\textwidth}
    \includegraphics[trim={80px 20px 80px 20px},clip,width=\linewidth,height=68px]{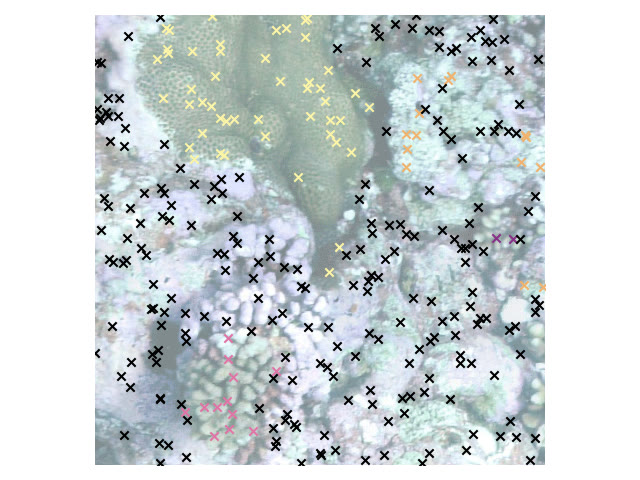}
        \caption{300}
    \end{subfigure}
    \begin{subfigure}{0.15\textwidth}
    \includegraphics[trim={80px 20px 80px 20px},clip,width=\linewidth,height=68px]{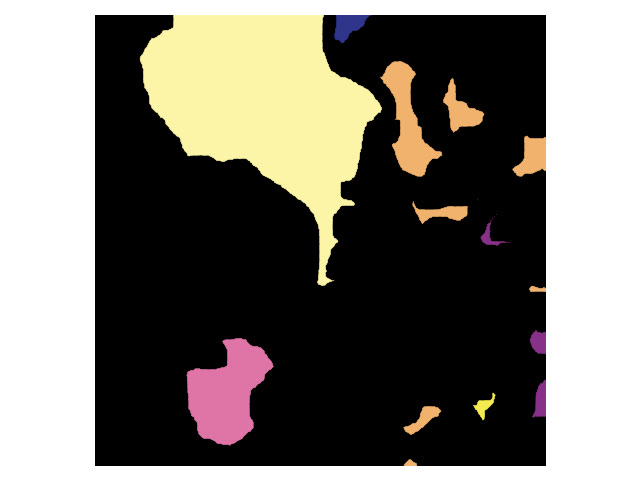}
        \caption{All}
    \end{subfigure}
    \caption{Example training sample for the UCSD mosaics transfer setting.}
    \label{fig:mosaics_sample}
\end{figure}

\begin{figure}[h]
    \centering
    \begin{subfigure}{0.15\textwidth}
    \includegraphics[width=\linewidth]{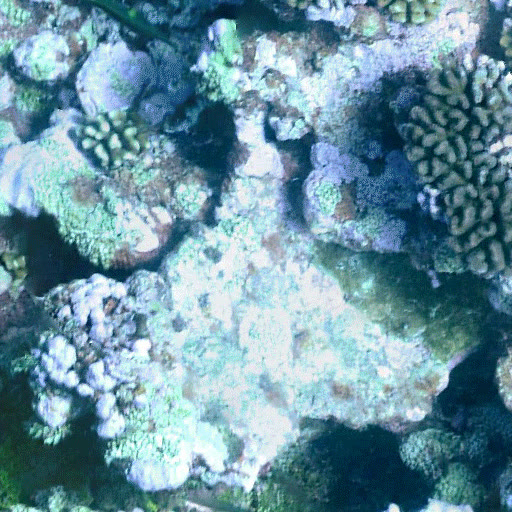}
        \caption{\\Image\\\phantom{x}}
    \end{subfigure}
    \begin{subfigure}{0.15\textwidth}
    \includegraphics[width=\linewidth]{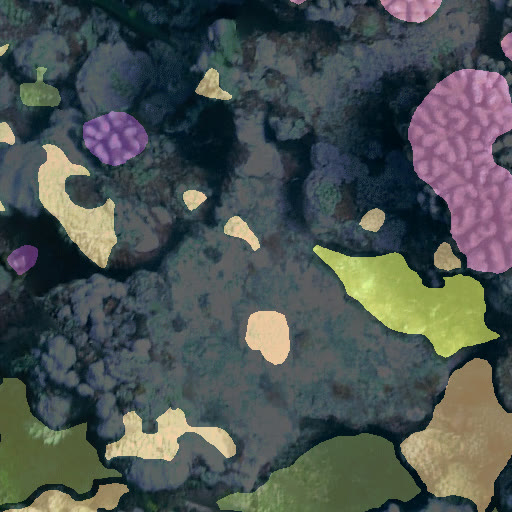}
        \caption{\\GT\\\phantom{x}}
    \end{subfigure}
    \begin{subfigure}{0.15\textwidth}
    \includegraphics[width=\linewidth]{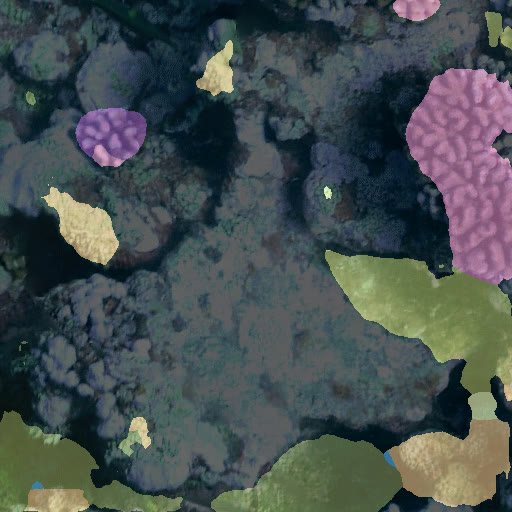}
        \caption{\\Pred\\Ours}
    \end{subfigure}
    \begin{subfigure}{0.15\textwidth}
    \includegraphics[width=\linewidth]{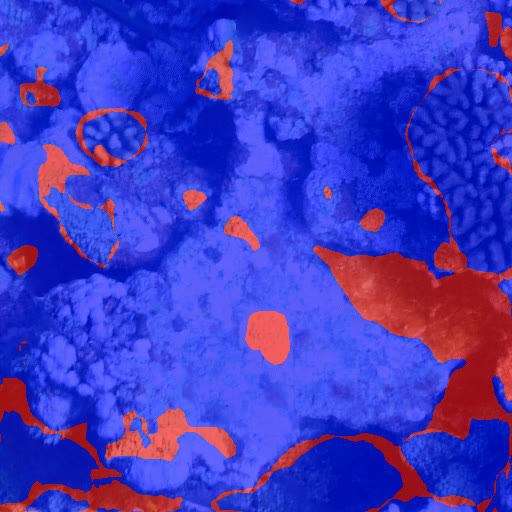}
        \caption{\\Correct\\Ours}
    \end{subfigure}
    \begin{subfigure}{0.15\textwidth}
    \includegraphics[width=\linewidth]{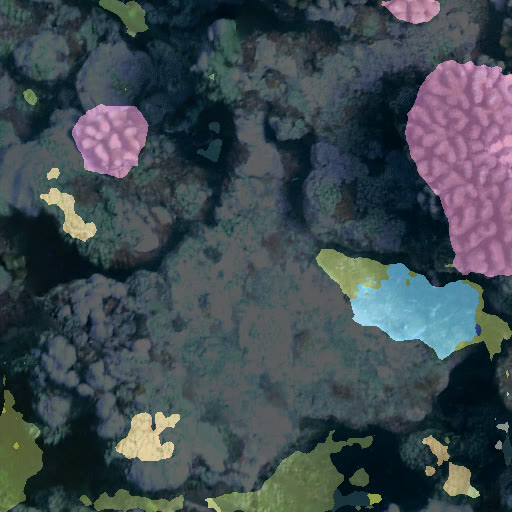}
        \caption{\\Pred\\ImageNet}
    \end{subfigure}
    \begin{subfigure}{0.15\textwidth}
    \includegraphics[width=\linewidth]{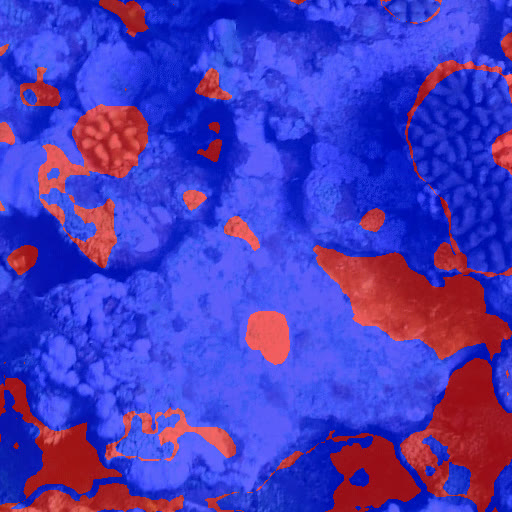}
        \caption{\\Correct\\ImageNet}
    \end{subfigure}
    \caption{Example predictions of DeepLabV3+ pre-trained on Coralscapes and an off-the-shelf (ImageNet) model, trained with 5 labels per image.}
    \label{fig:mosaics_result}
\end{figure}

All models are trained using the same optimizer and learning rate setting as in the Coralscapes benchmarking experiments. During training, only random horizontal and vertical flips of the 512$\times$512px patches during training as data augmentations. The models are trained for 100 epochs in the  5 \& 10 labels per image setting, and for 200 epochs otherwise. Example training labels in each of the settings are shown in Figure~\ref{fig:mosaics_sample}, and example model outputs for DeepLabV3+ trained in the sparsest setting are shown in Figure~\ref{fig:mosaics_result}.

\subsection{Crown-of-Thorns Starfish Survey}

The CSIRO Crown-of-Thorns dataset \cite{cots} was originally a challenge on Kaggle. As the competition has finished, the original test dataset is no longer available. The segmentation models that were pre-trained on Coralscapes were saved and the last (output) layer manually reshaped from 39 classes to 2 classes, with the `COTS' neuron being the `Crown of Thorn' neuron from Coralscapes, and the other neuron the mean of the remaining 38 Coralscapes classes. We train the segmentation models for 10 epochs on square patches of size 720, with otherwise the same hyperparameters as training on Coralscapes. At inference, a sliding window with stride 280 is used for prediction. As a baseline, we train YoloV8-L \cite{yolov8} for 100 epochs using the `ultralytics' Python library \cite{ultralytics}, in the default settings, which initializes from a pre-trained model and applies a range of image augmentations during training. 

\begin{figure}[h]
    \centering
    \begin{subfigure}{0.19\textwidth}
    \includegraphics[width=\linewidth]{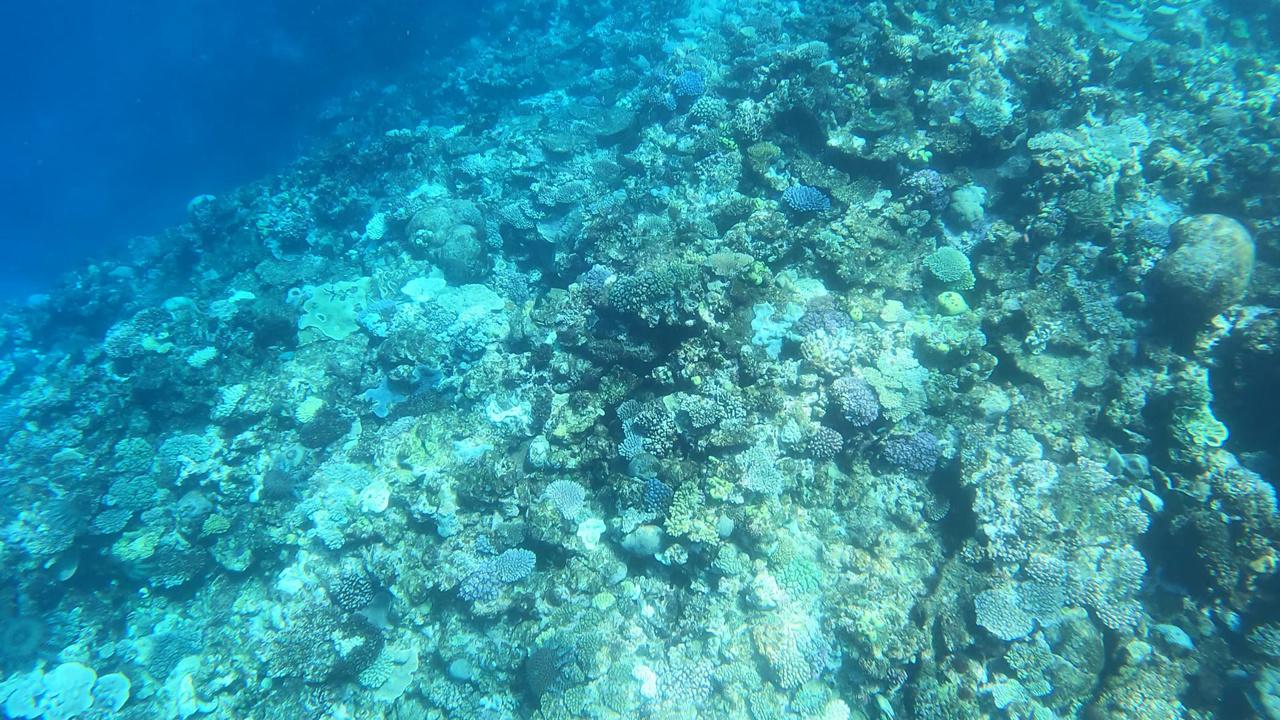}
        \caption{Image}
    \end{subfigure}
    \begin{subfigure}{0.19\textwidth}
    \includegraphics[width=\linewidth]{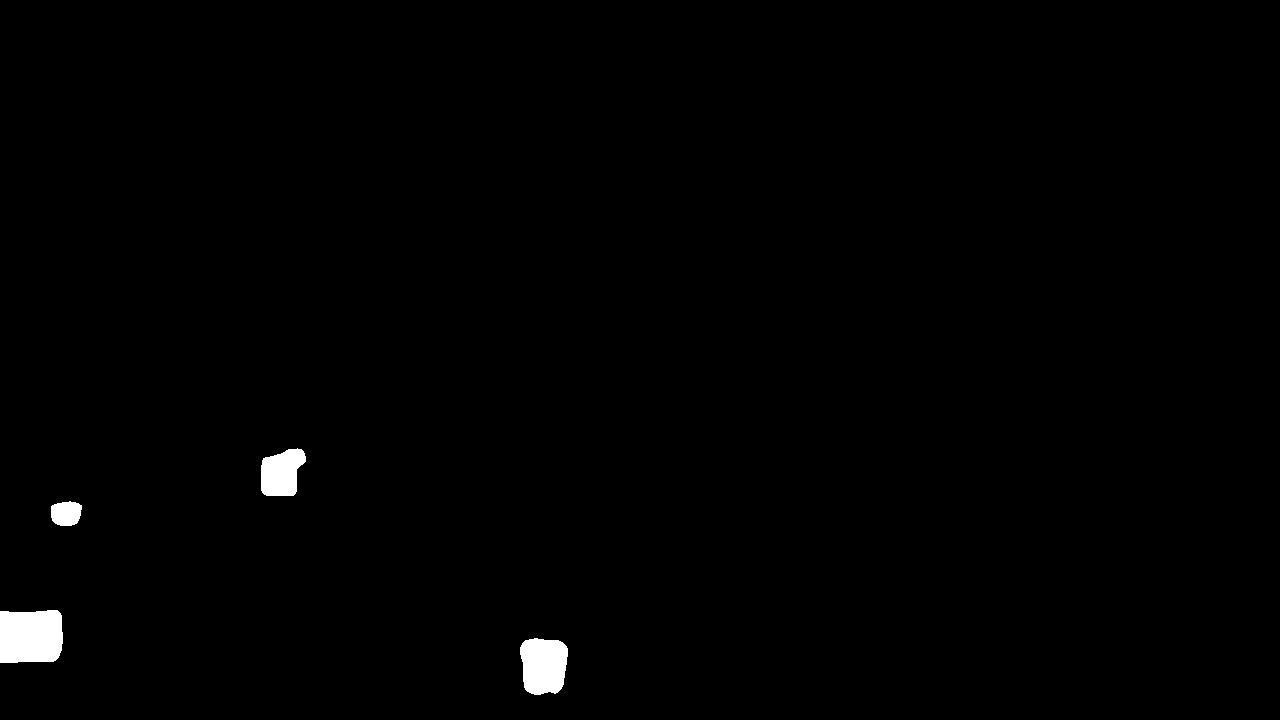}
        \caption{Segmentation}
    \end{subfigure}
    \begin{subfigure}{0.19\textwidth}
    \includegraphics[width=\linewidth]{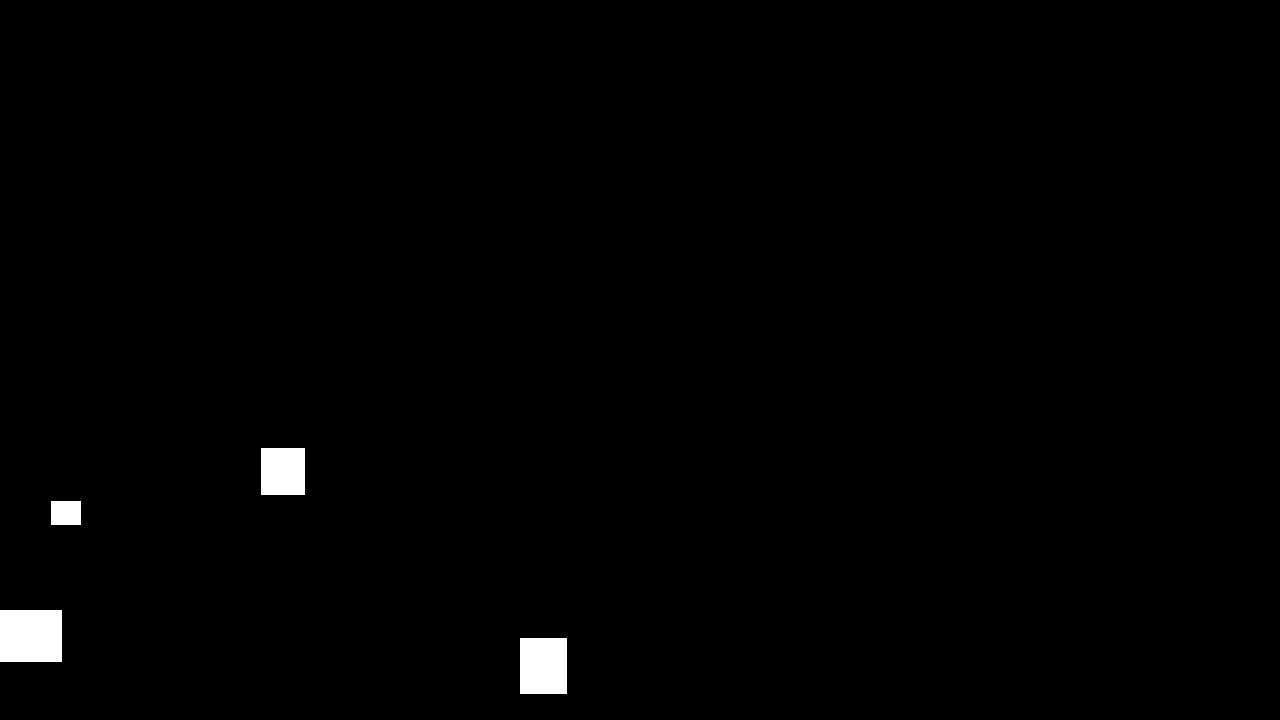}
        \caption{Bbox Prediction}
    \end{subfigure}
    \begin{subfigure}{0.19\textwidth}
    \includegraphics[width=\linewidth]{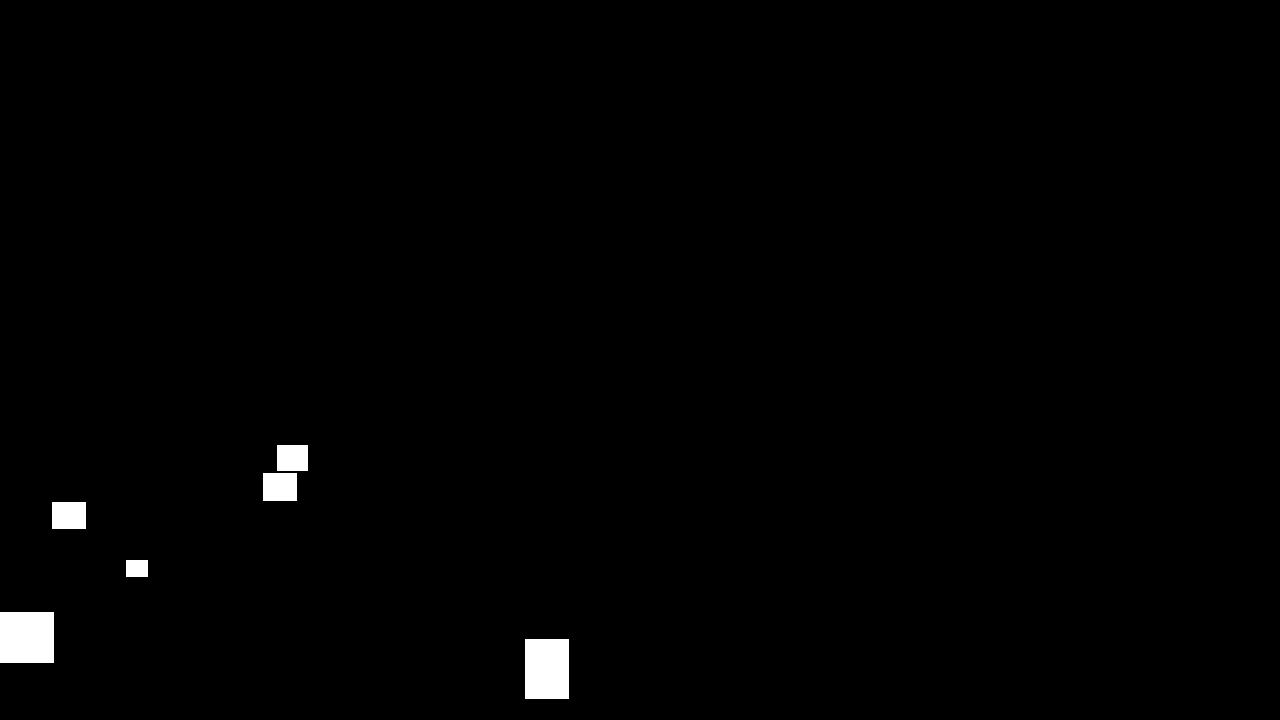}
        \caption{Bbox GT}
    \end{subfigure}
    \caption{Example of how the images are transformed to bounding boxes when using SegFormer in the COTS detection task.}
    \label{fig:enter-label}
\end{figure}

\clearpage
\section{Comparison to CoralSCOP}
\label{appendix:coralscop}
In this Section we emphasize the key differences from Coralscapes to the dataset from CoralSCOP \cite{zheng2024coralscop}: \vspace{-3pt}
\begin{enumerate}
    \item \textbf{General purpose:} CoralSCOP only has segmentation masks for visible corals, and no other classes. This makes CoralSCOP unsuitable for many applications (e.g. marine robotics), where it is critical to segment other classes (divers, fish, other substrates) to avoid collisions and ensure safe deployment. Coralscapes goes beyond coral segmentation, and aims for \textbf{scene understanding} in coral reefs. \vspace{-3pt}
    \item \textbf{Data quality:} Our data was collected as part of research \& monitoring campaigns by the authors. We can attest that all images are \textbf{original and authentic} (not altered or generated by AI), whereas the majority of CoralSCOP is crawled from the internet or recompiled from existing sources.\vspace{-3pt}
    \item \textbf{Annotation quality:} The Coralscapes segmentations follow a stricter quality standard than existing datasets w.r.t to coverage and a conservatively applied class hierarchy.
\end{enumerate}
These key points position Coralscapes as a high-quality dataset for general purpose reef segmentation, whereas CoralSCOP's main aim is breadth and coverage.

\begin{figure}[b!]
    \centering
    \includegraphics[width=0.95\linewidth]{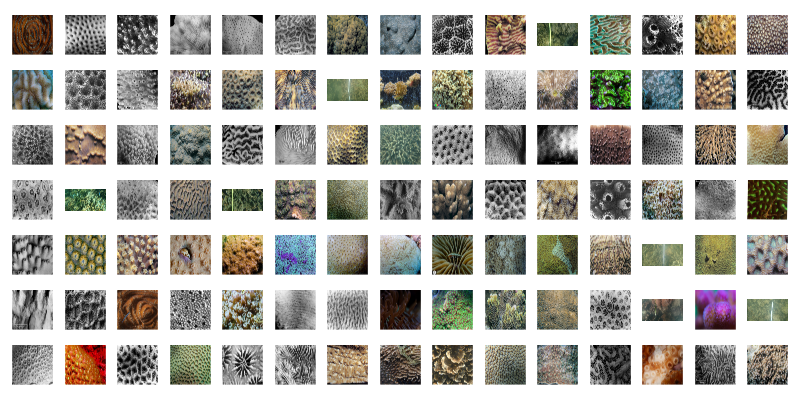}
    \includegraphics[width=0.93\linewidth]{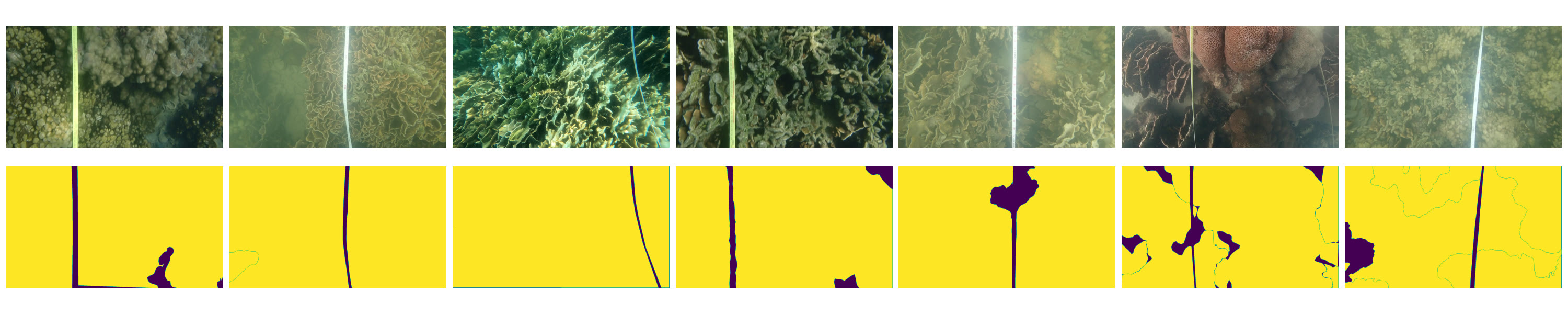}
    \caption{Samples from CoralSCOP. Top: 100 samples of images with $\geq$90\% annotation cover, showing that the vast majority are extreme close-ups of corals. Bottom: for images that are not close-ups, the coral segmentation masks are coarse.}
    \label{fig:coralmask}
\end{figure}

To supplement the findings of Fig.~3, which show that Coralscapes is more complex than CoralSCOP, we offer a direct comparison to CoralSCOP images that have a high proportion of the images annotated in Fig.~\ref{fig:coralmask}.
Of the first 105 CoralSCOP images with $geq$90\% cover (below), 98 are extreme close-ups of corals, and the remaining ones have severely imprecise segmentation masks: there is no clear distinction between dead and live coral, and often there are rocks \& large dark patches which are mistakenly included in the coral mask. The images shown are more than 6\% of CoralSCOP images with $\geq$90\% annotation cover. 
\clearpage

\section{Additional Samples}
\label{appendix:additional_samples}
\vspace{-10pt}
\begin{figure*}[b]
    \centering
    \includegraphics[width=\linewidth]{figures/legend.pdf}
    \begin{subfigure}{0.195\textwidth}
        \centering
        \includegraphics[width=\linewidth]{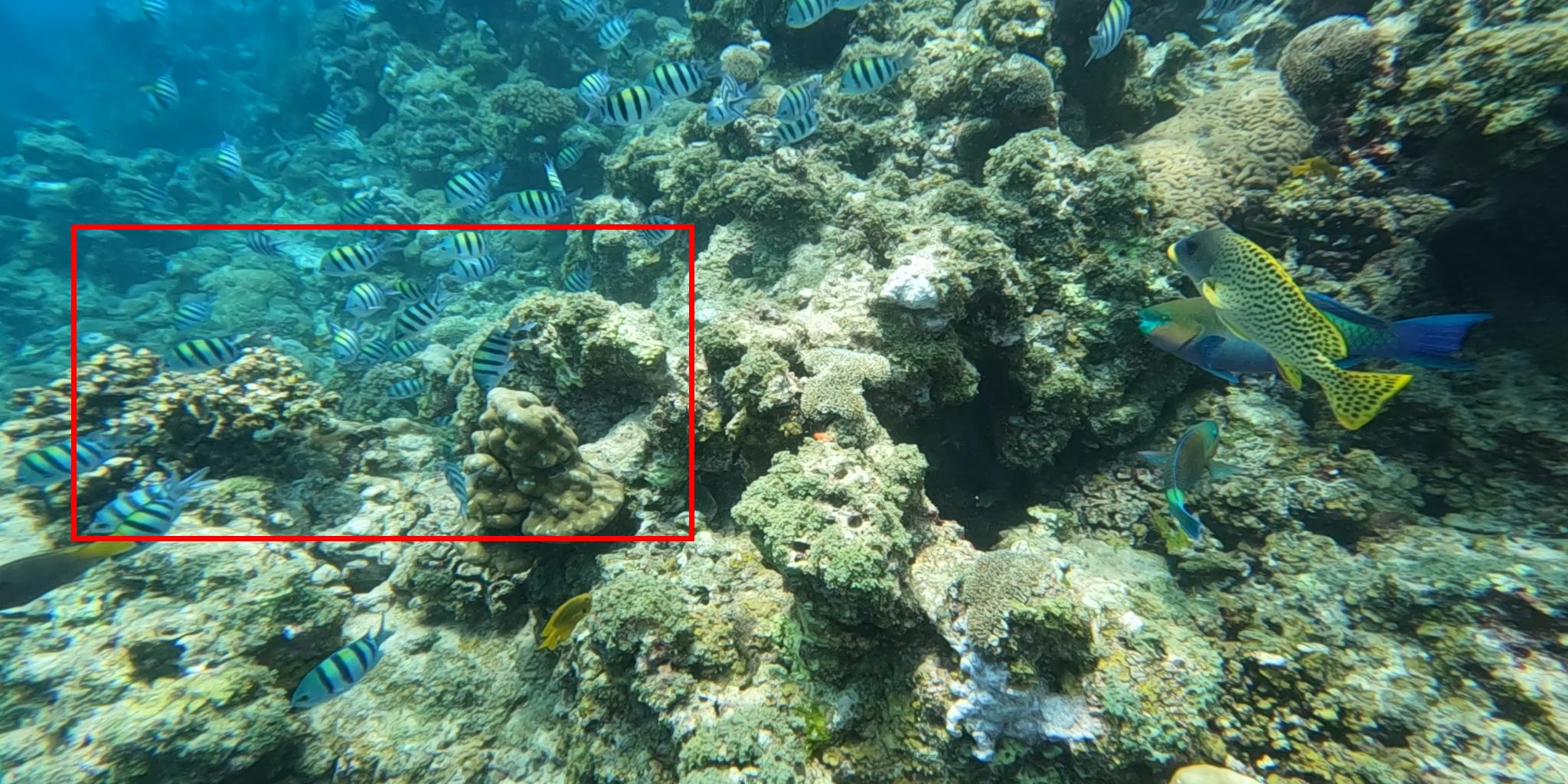} 
        \includegraphics[width=\linewidth]{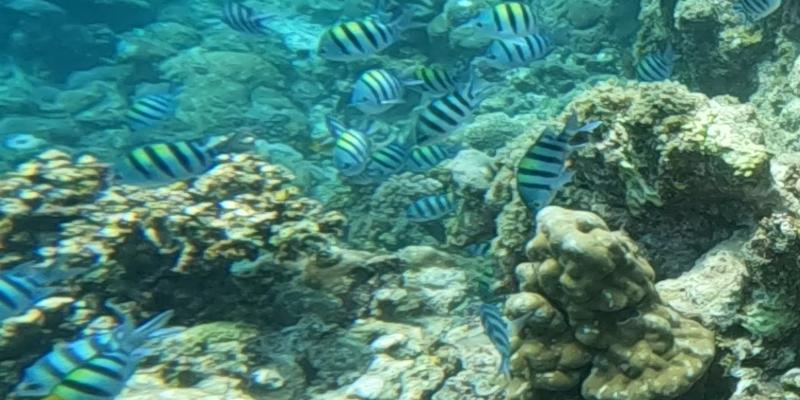}
        \includegraphics[width=\linewidth]{figures/benchmarking_examples/0039image.rgb.jpg} 
        \includegraphics[width=\linewidth]{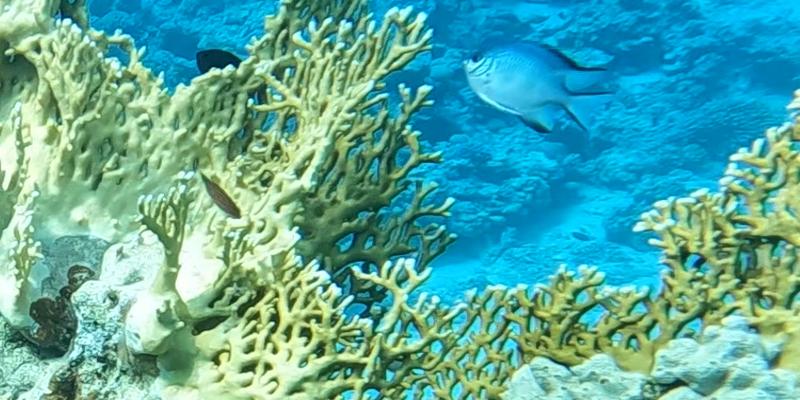} 
        \includegraphics[width=\linewidth]{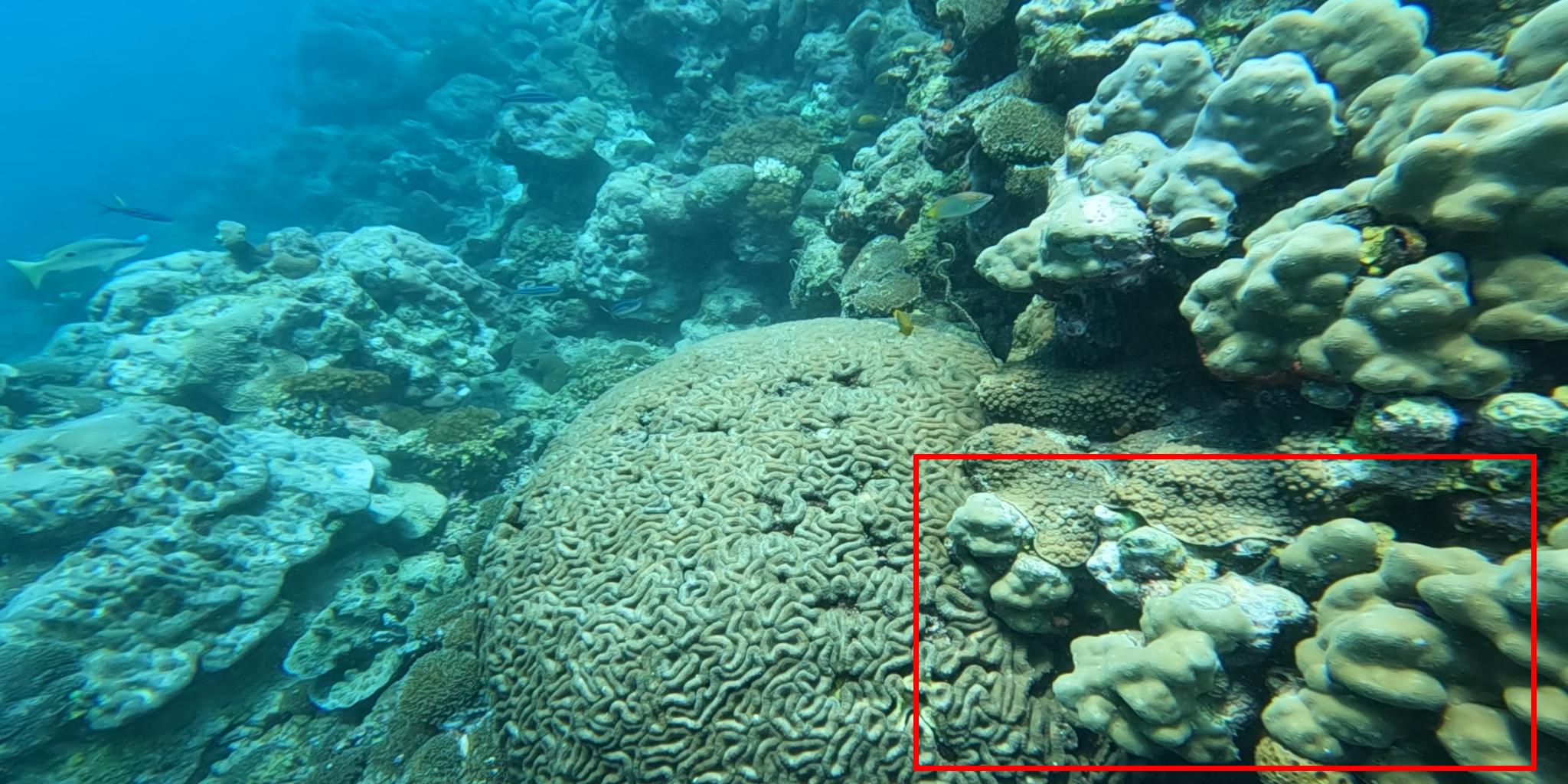} 
        \includegraphics[width=\linewidth]{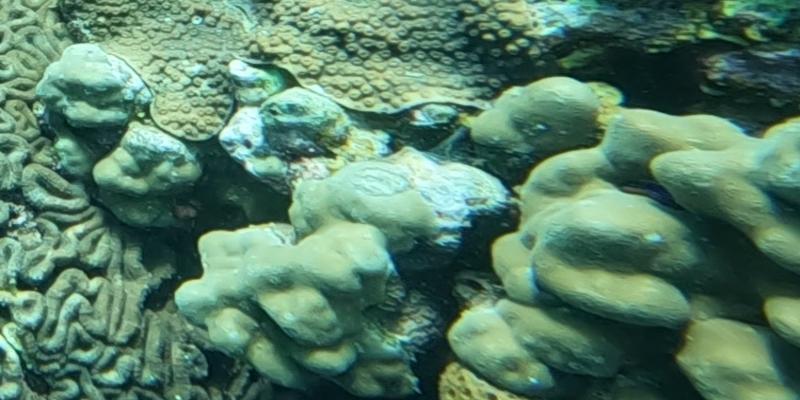} 
        \includegraphics[width=\linewidth]{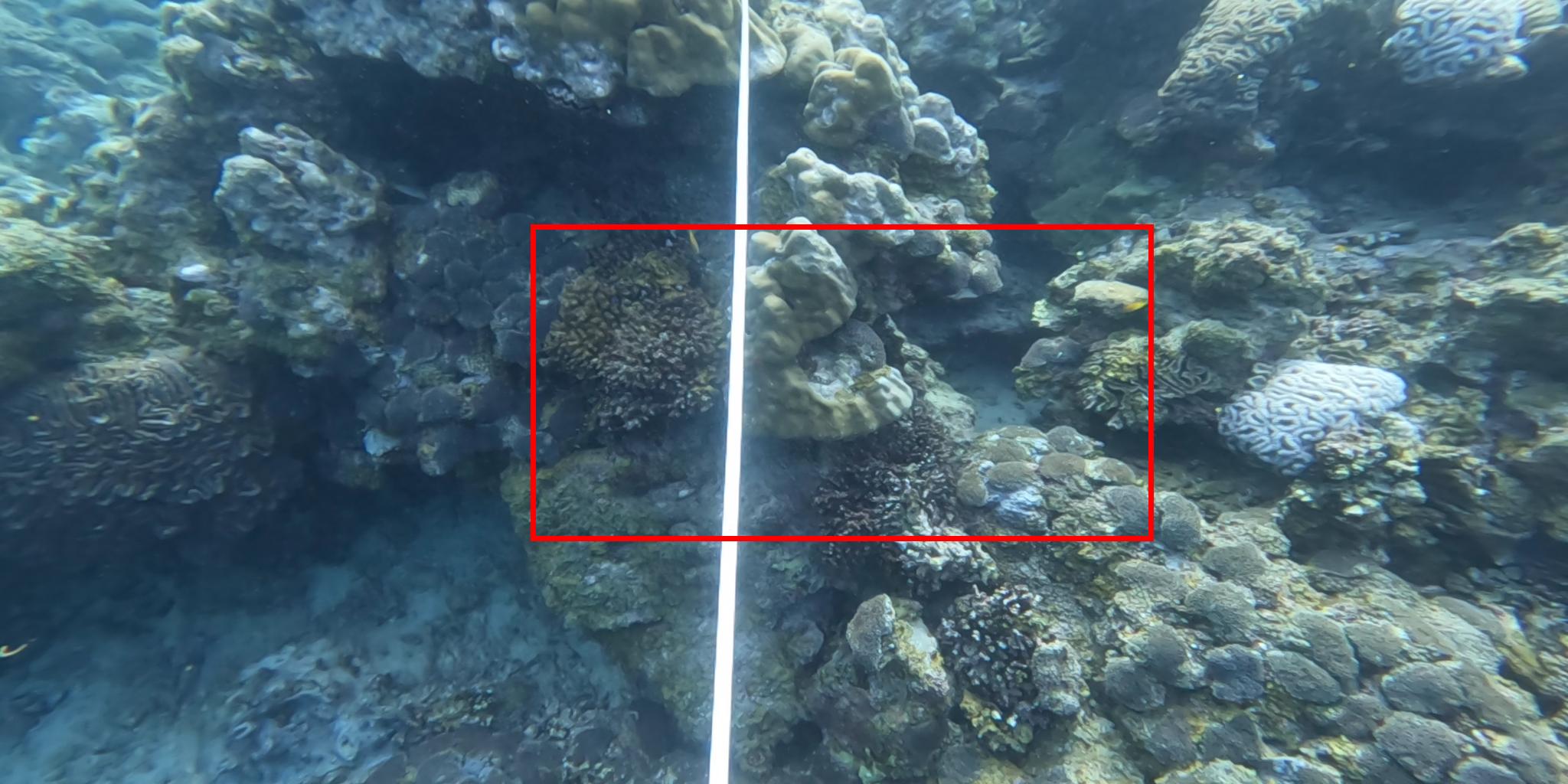} 
        \includegraphics[width=\linewidth]{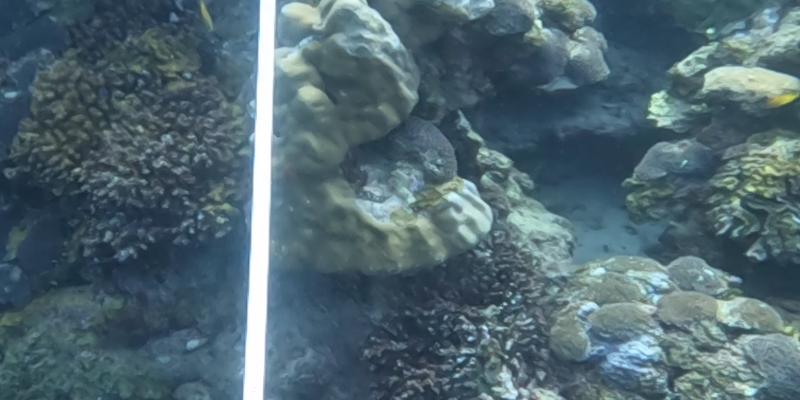} 
        \includegraphics[width=\linewidth]{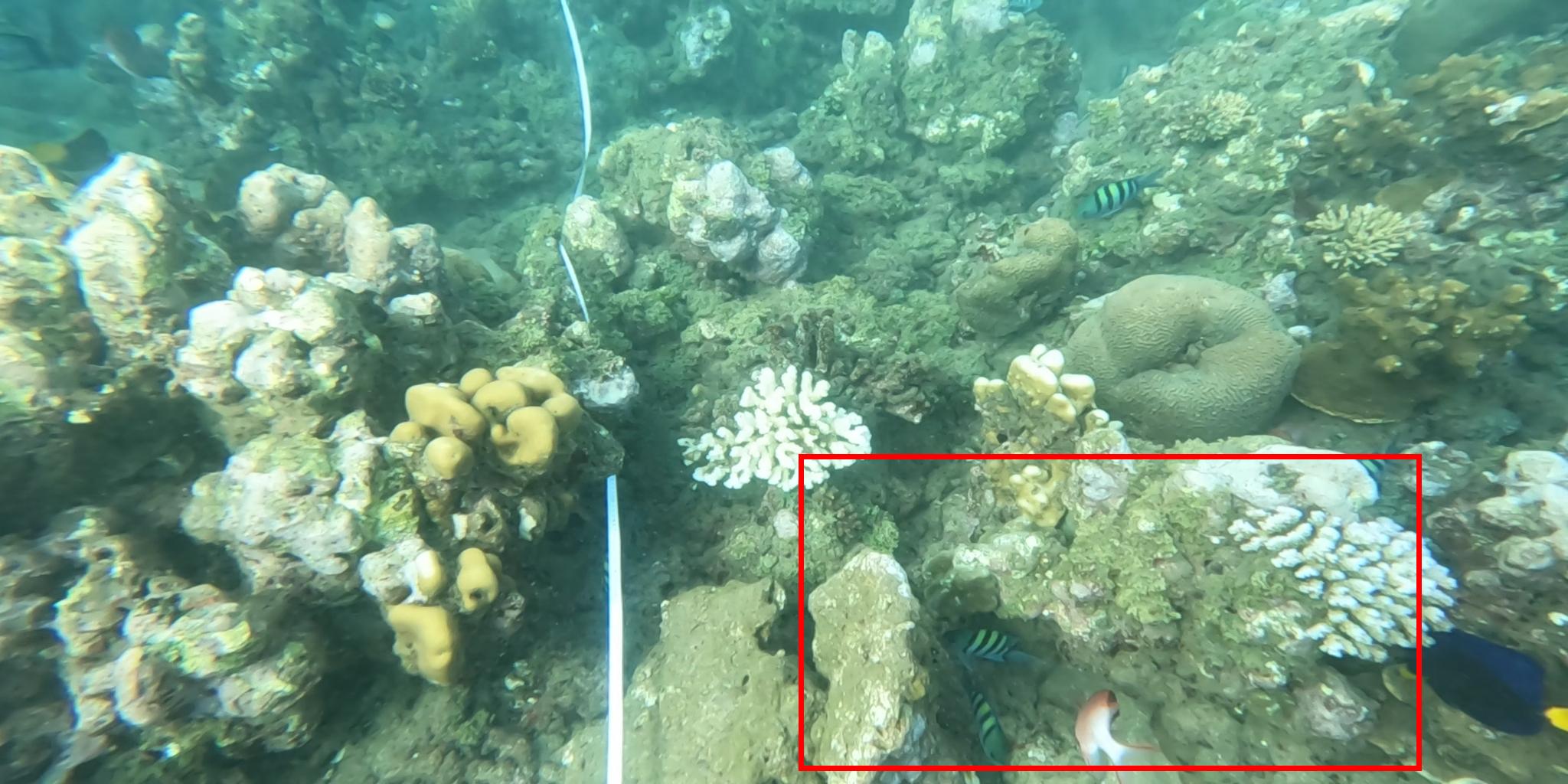} 
        \includegraphics[width=\linewidth]{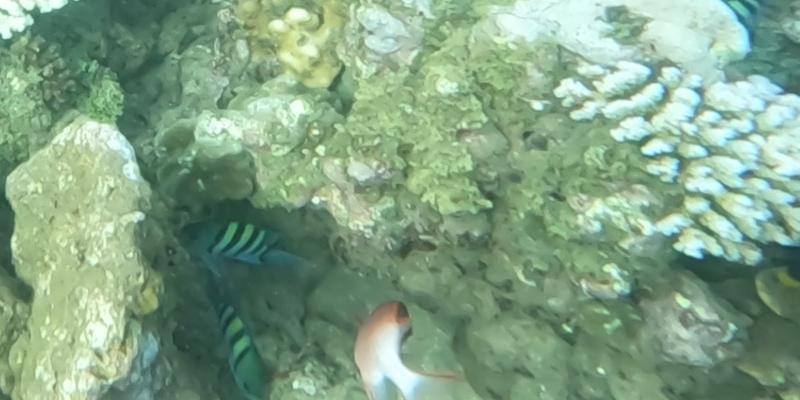} 
        \caption{\\Image}
        \label{fig:sub1_additional}
    \end{subfigure}
    \begin{subfigure}{0.195\textwidth}
        \centering
        \includegraphics[width=\linewidth]{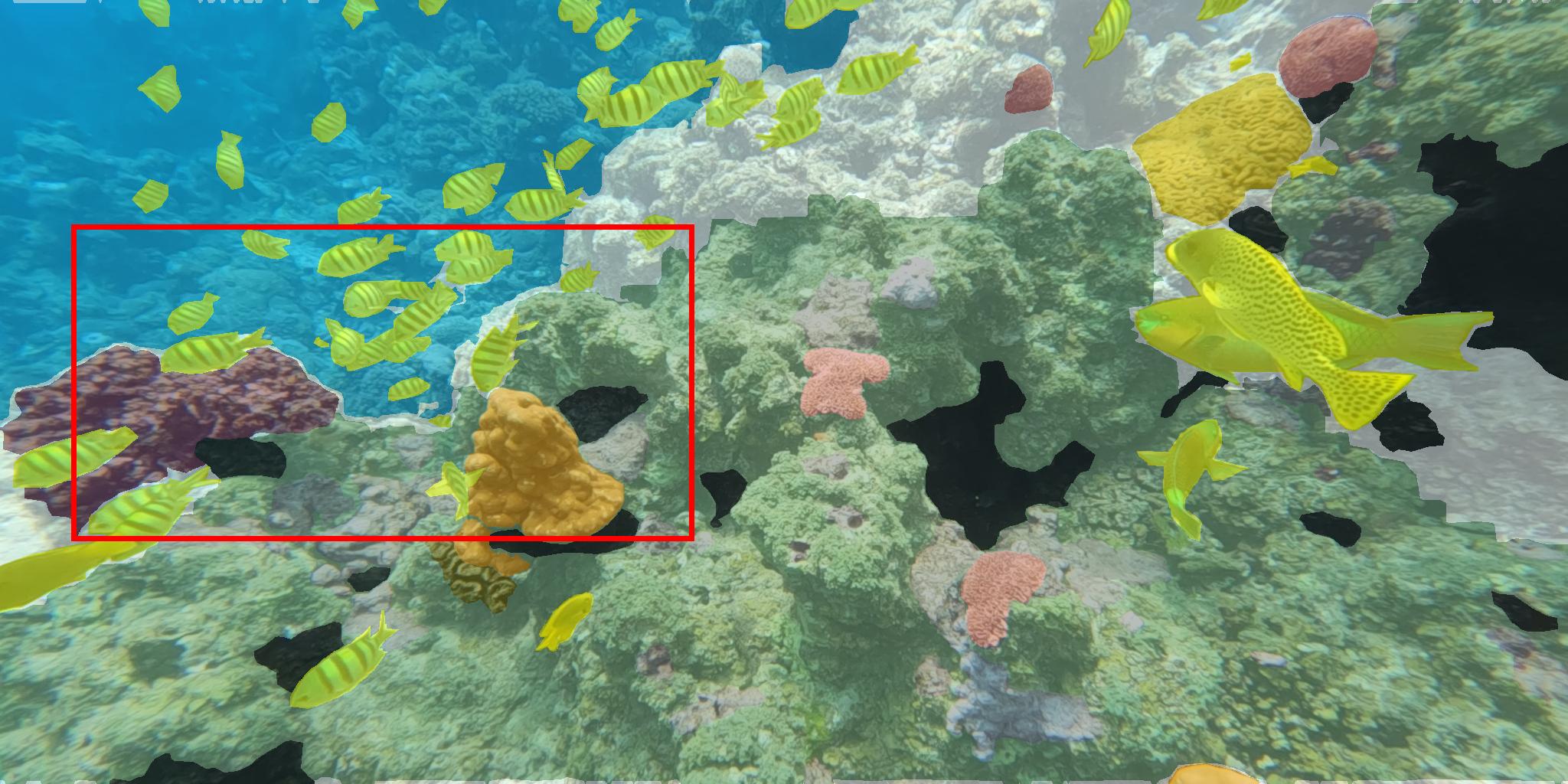} 
        \includegraphics[width=\linewidth]{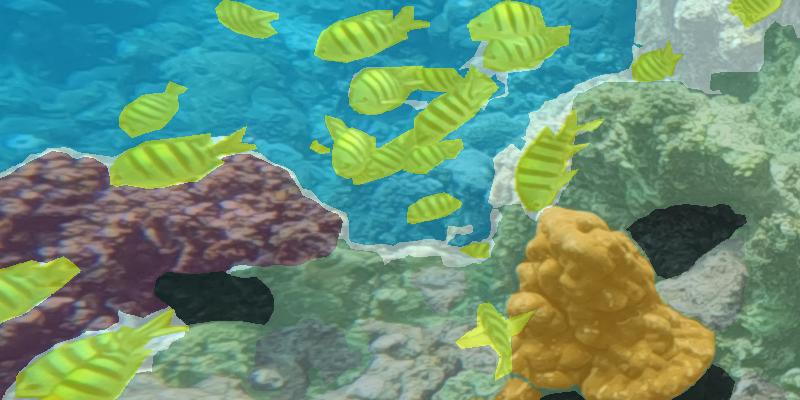} 
        \includegraphics[width=\linewidth]{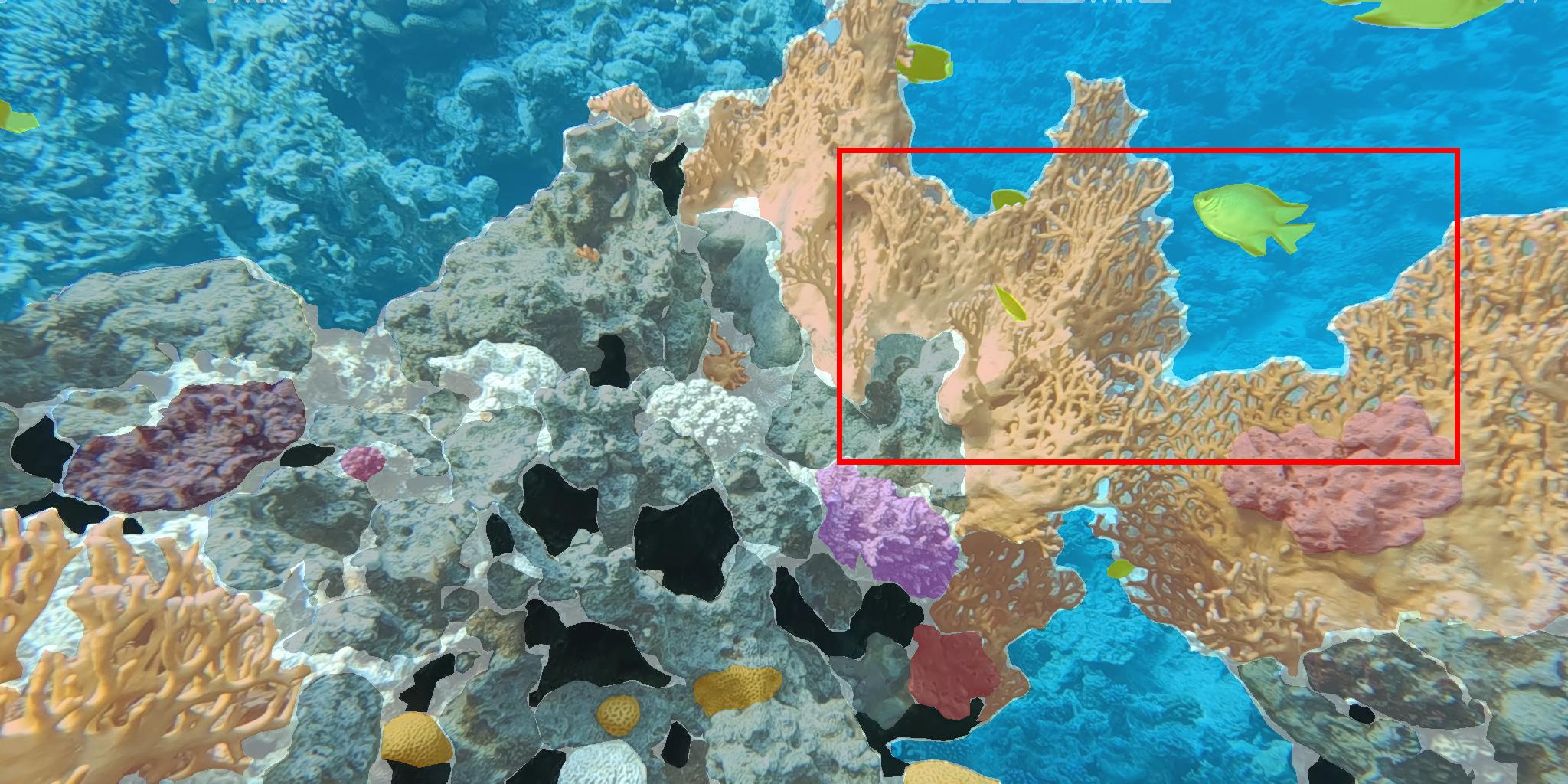} 
        \includegraphics[width=\linewidth]{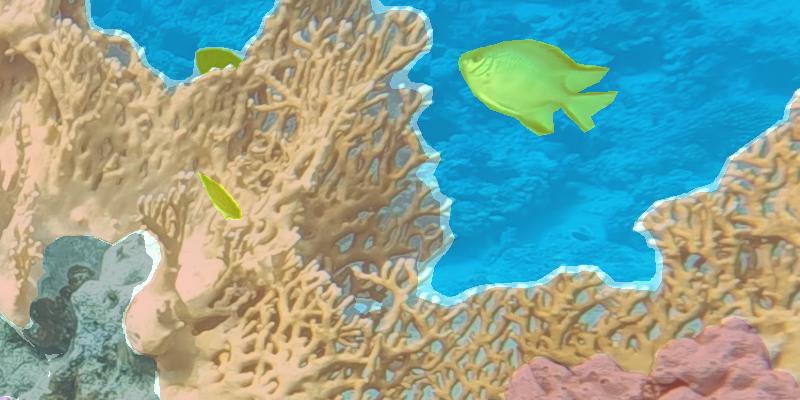}
        \includegraphics[width=\linewidth]{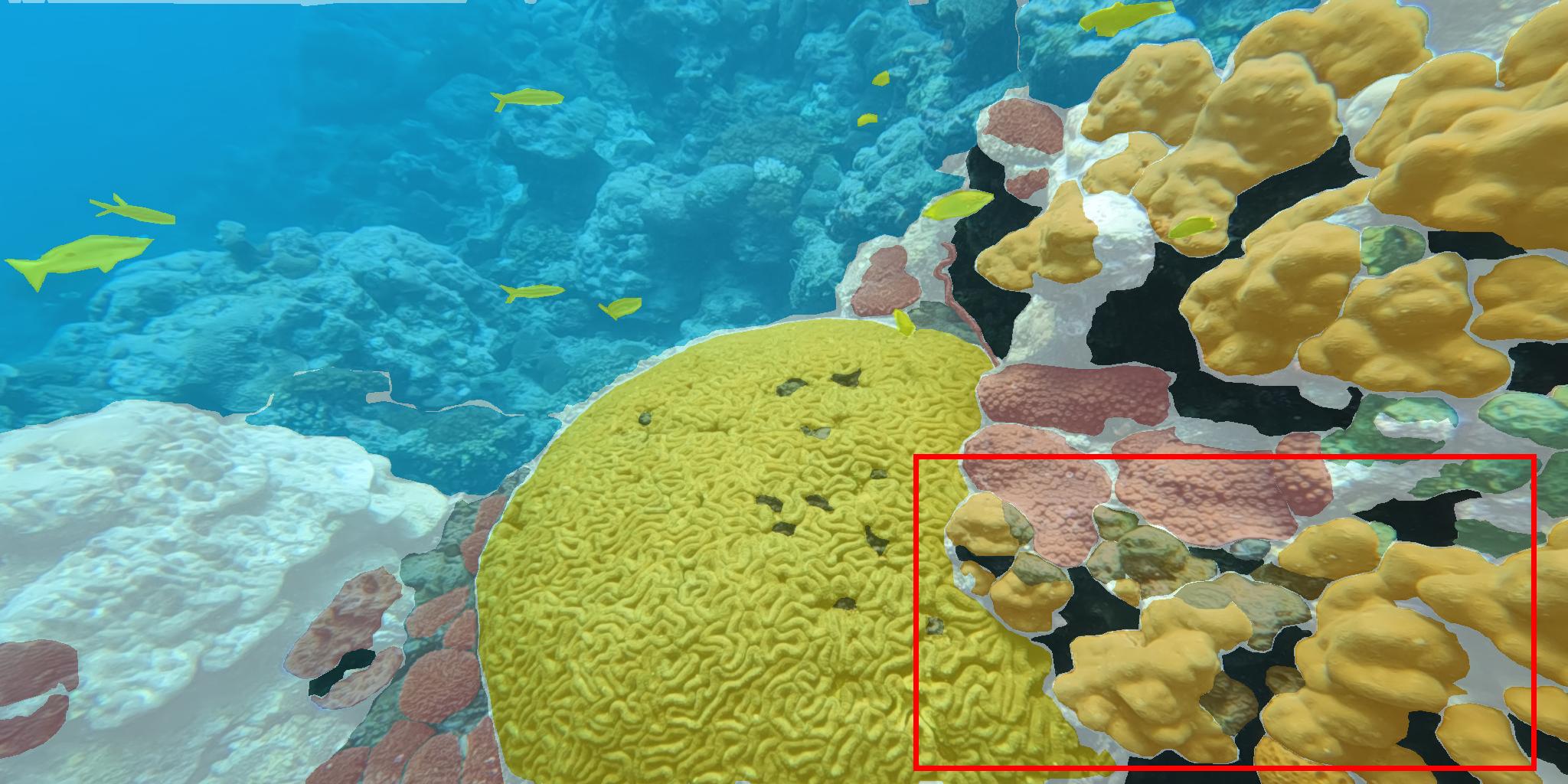} 
        \includegraphics[width=\linewidth]{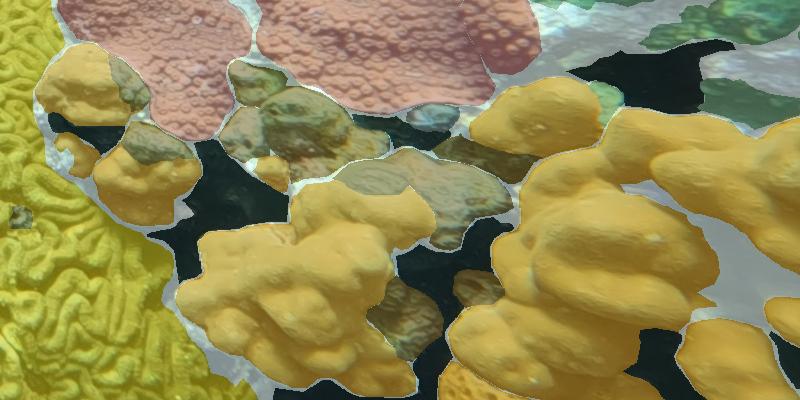} 
        \includegraphics[width=\linewidth]{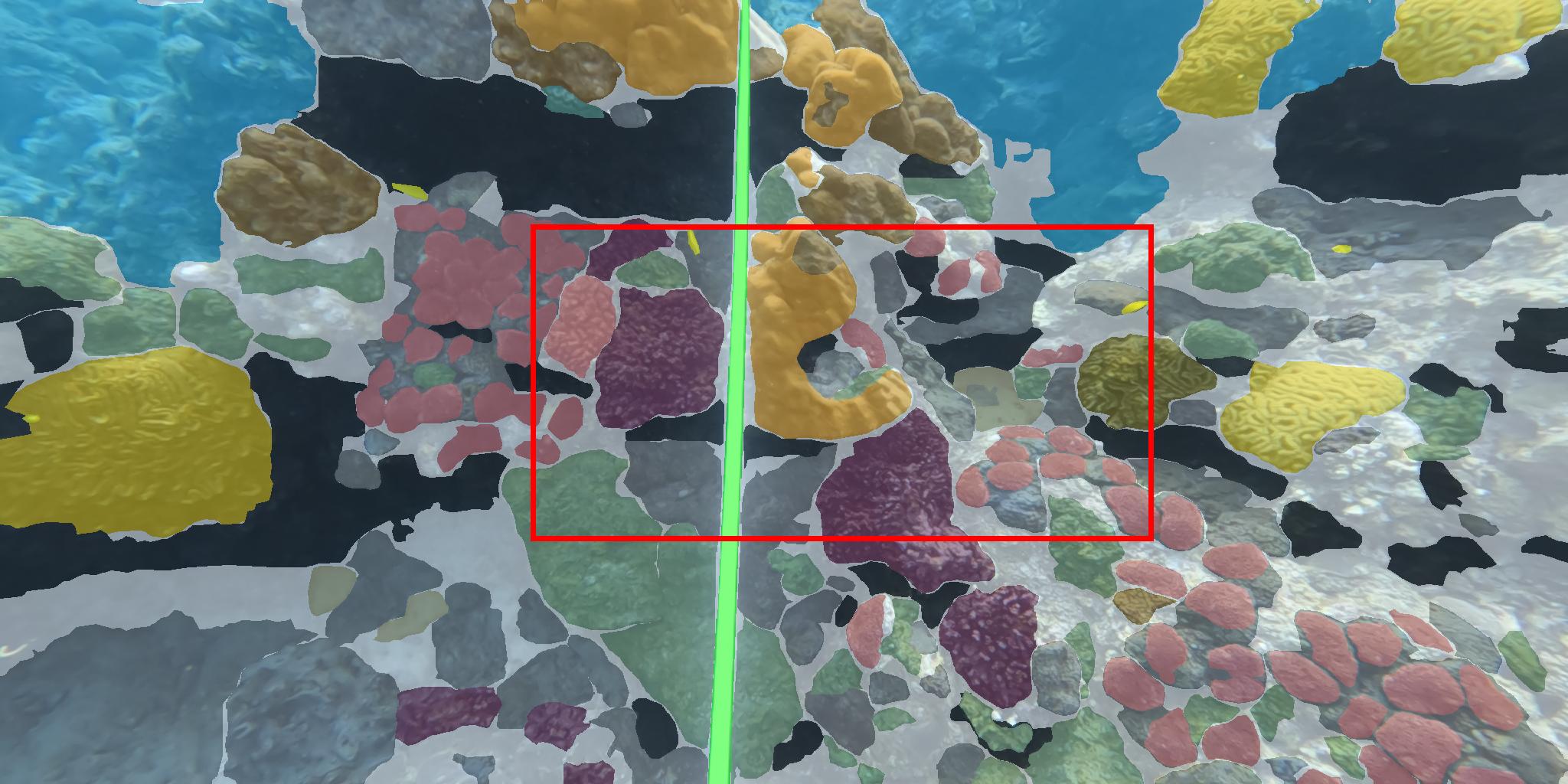} 
        \includegraphics[width=\linewidth]{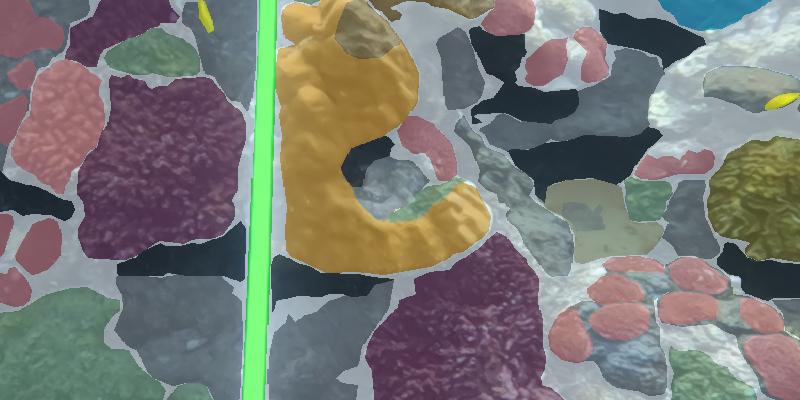}         
        \includegraphics[width=\linewidth]{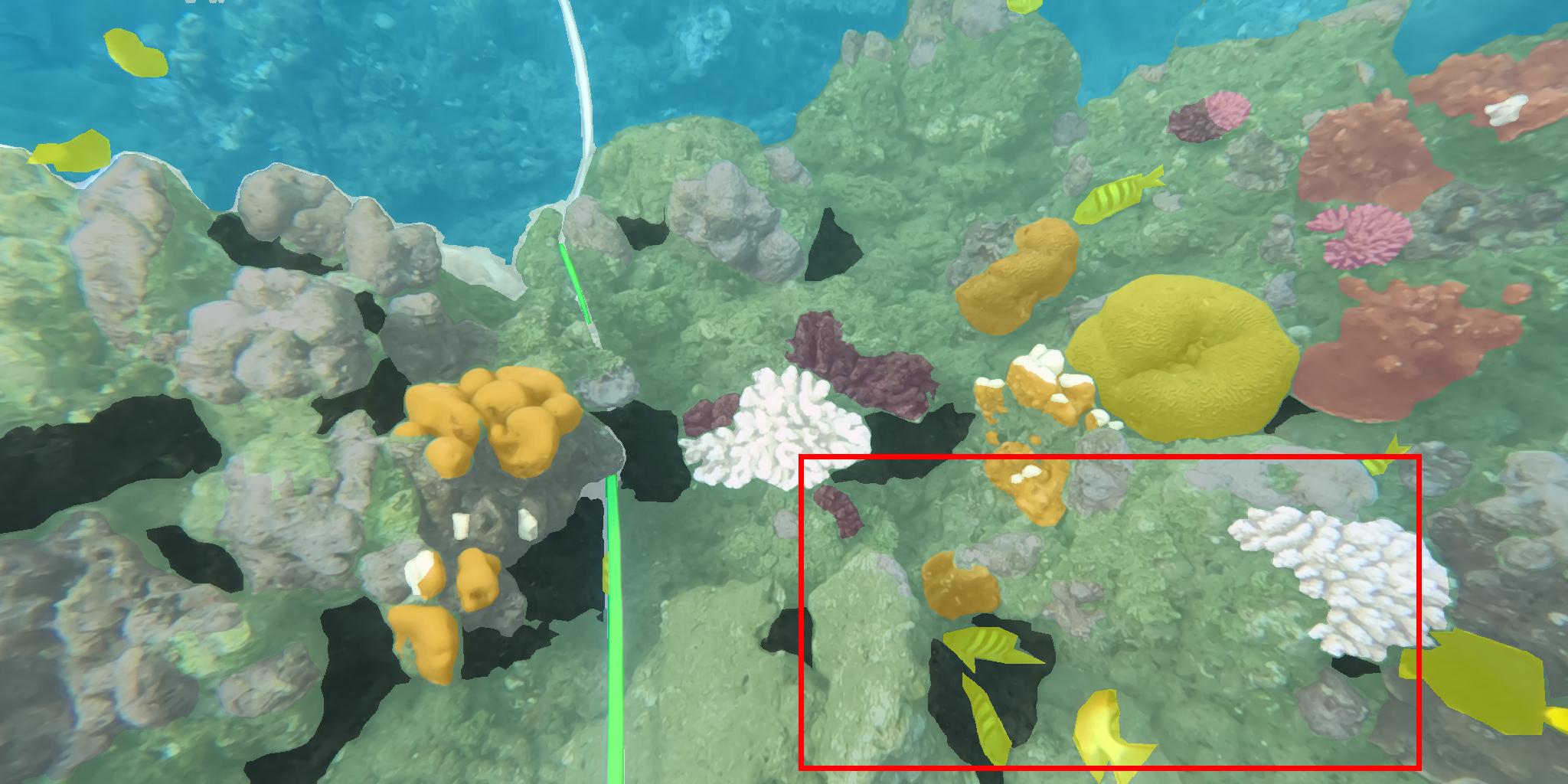} 
        \includegraphics[width=\linewidth]{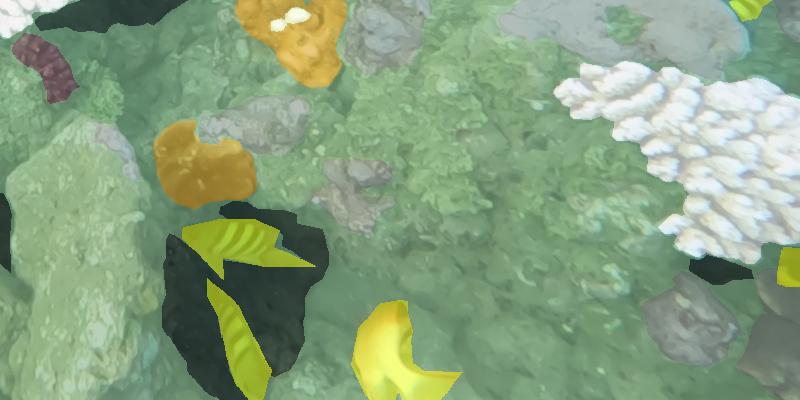}         
        \caption{\\Ground Truth}
        \label{fig:sub2_additional}
    \end{subfigure}
    \begin{subfigure}{0.195\textwidth}
        \centering
        \includegraphics[width=\linewidth]{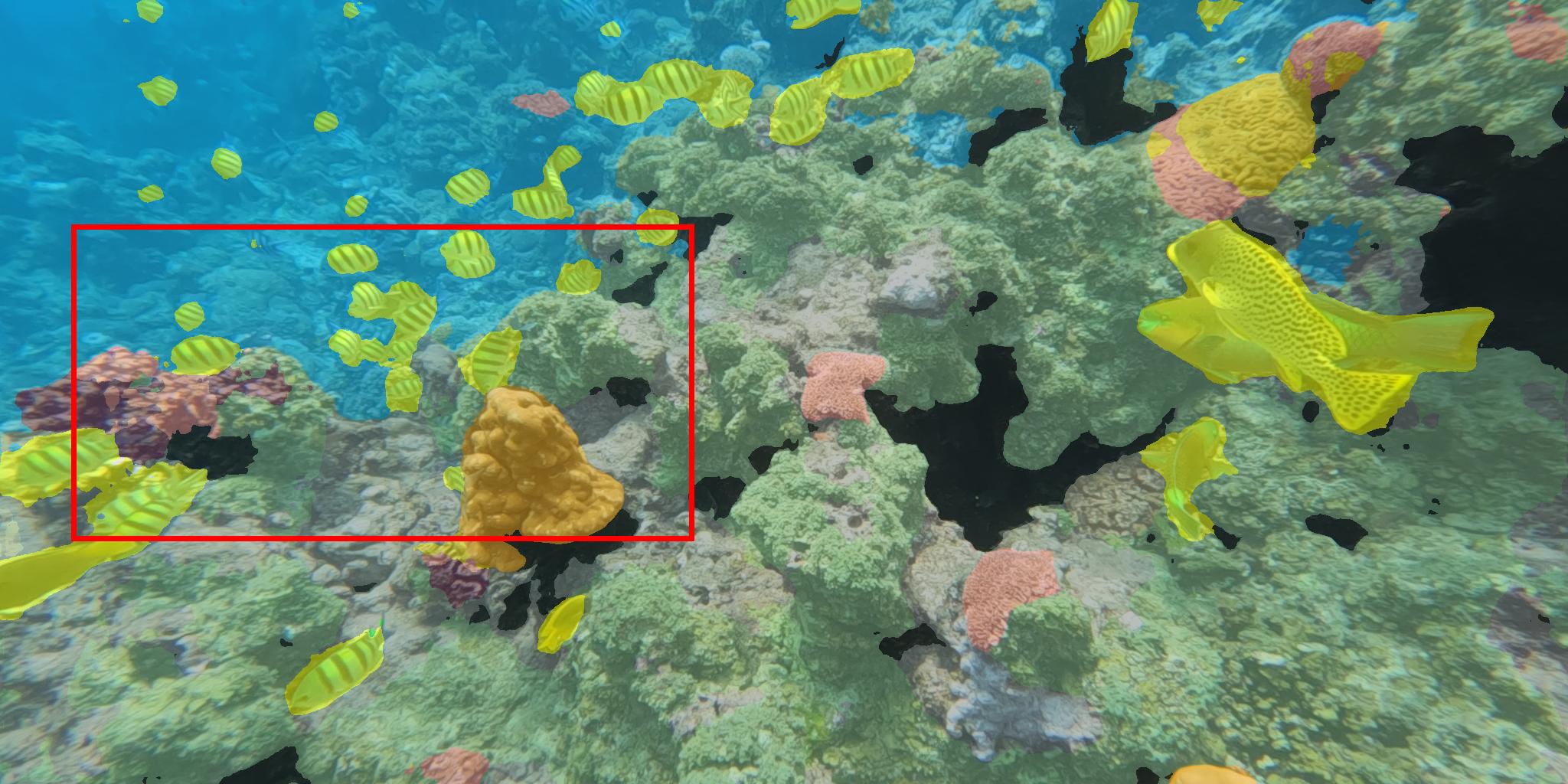} 
        \includegraphics[width=\linewidth]{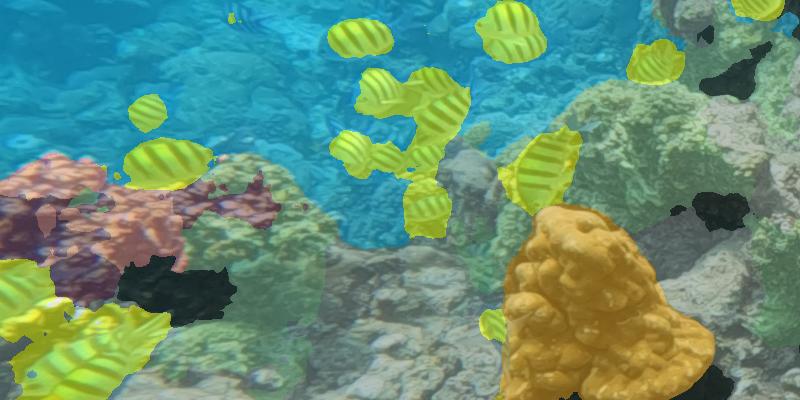} 
        \includegraphics[width=\linewidth]{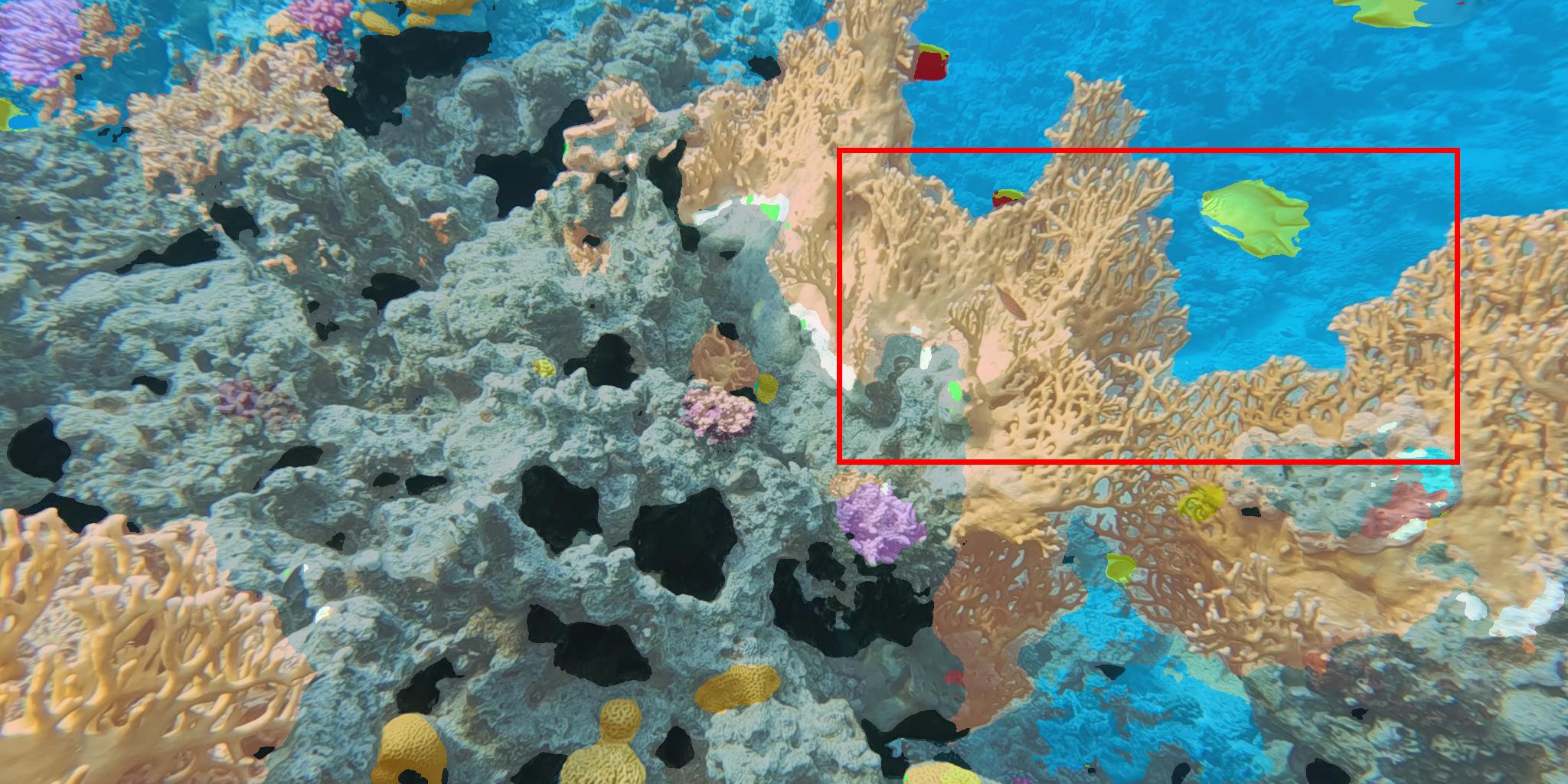} 
        \includegraphics[width=\linewidth]{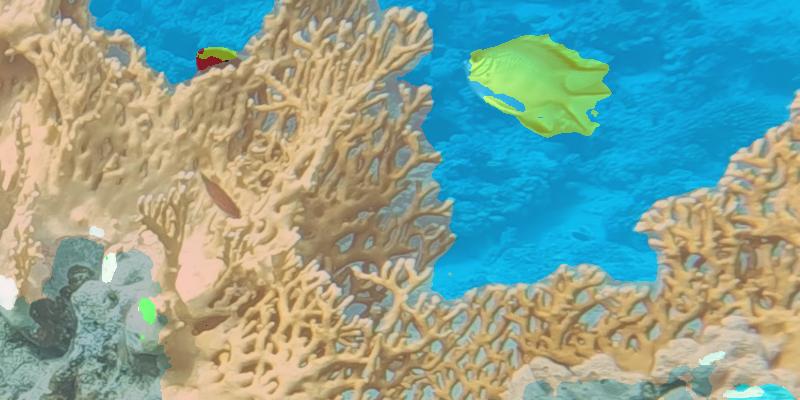} 
        \includegraphics[width=\linewidth]{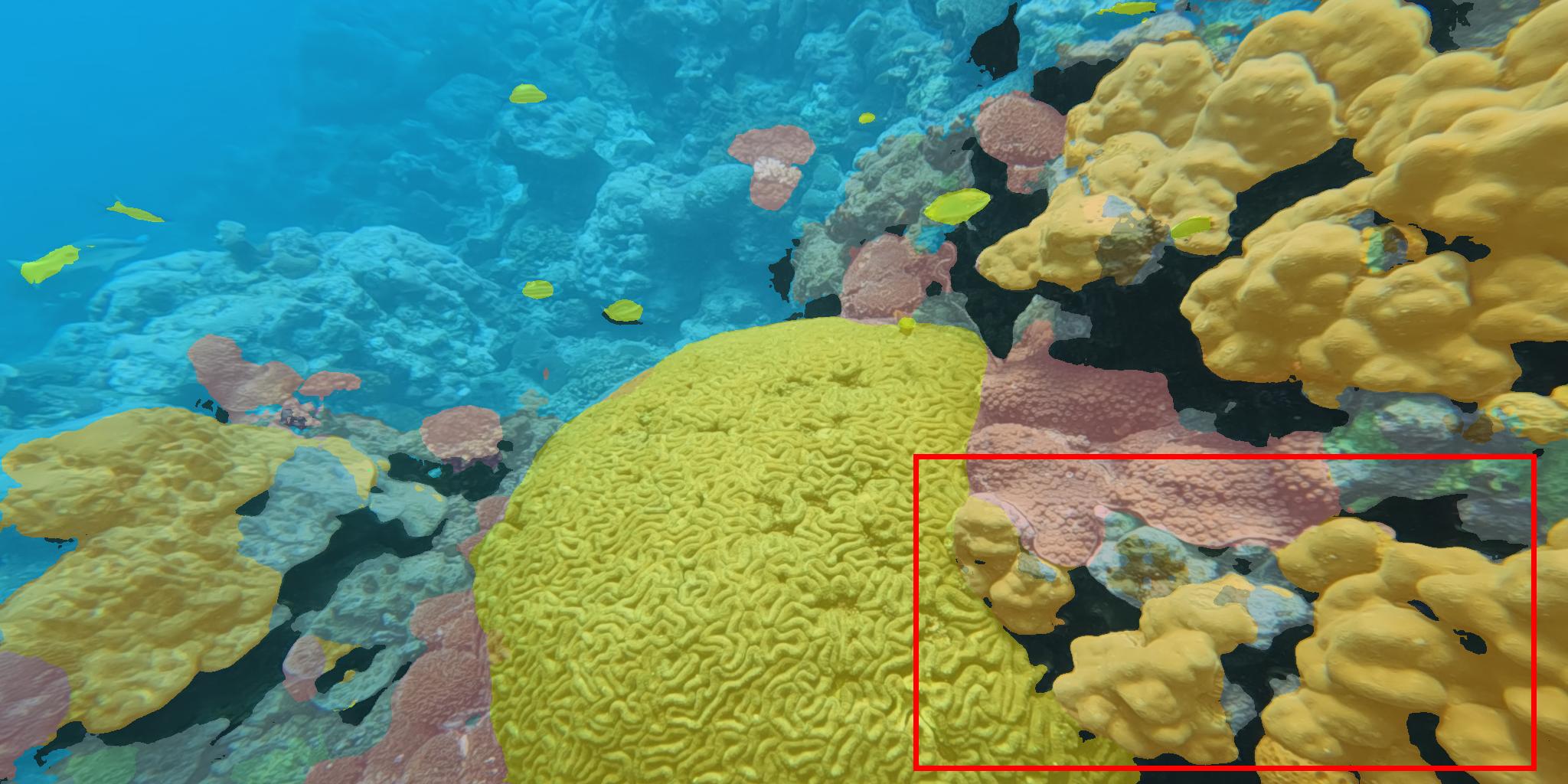} 
        \includegraphics[width=\linewidth]{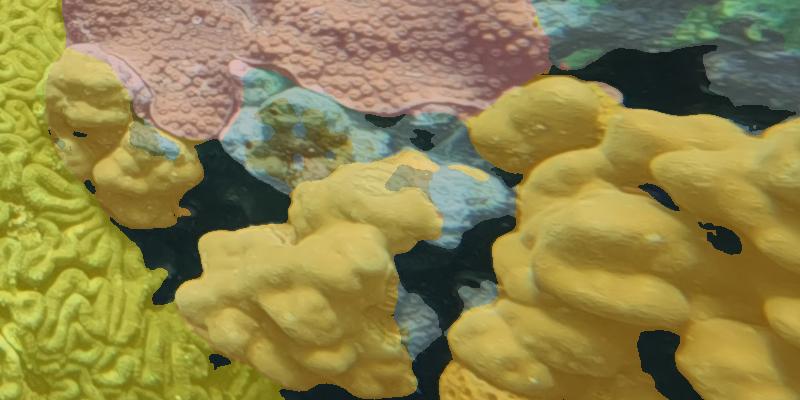} 
        \includegraphics[width=\linewidth]{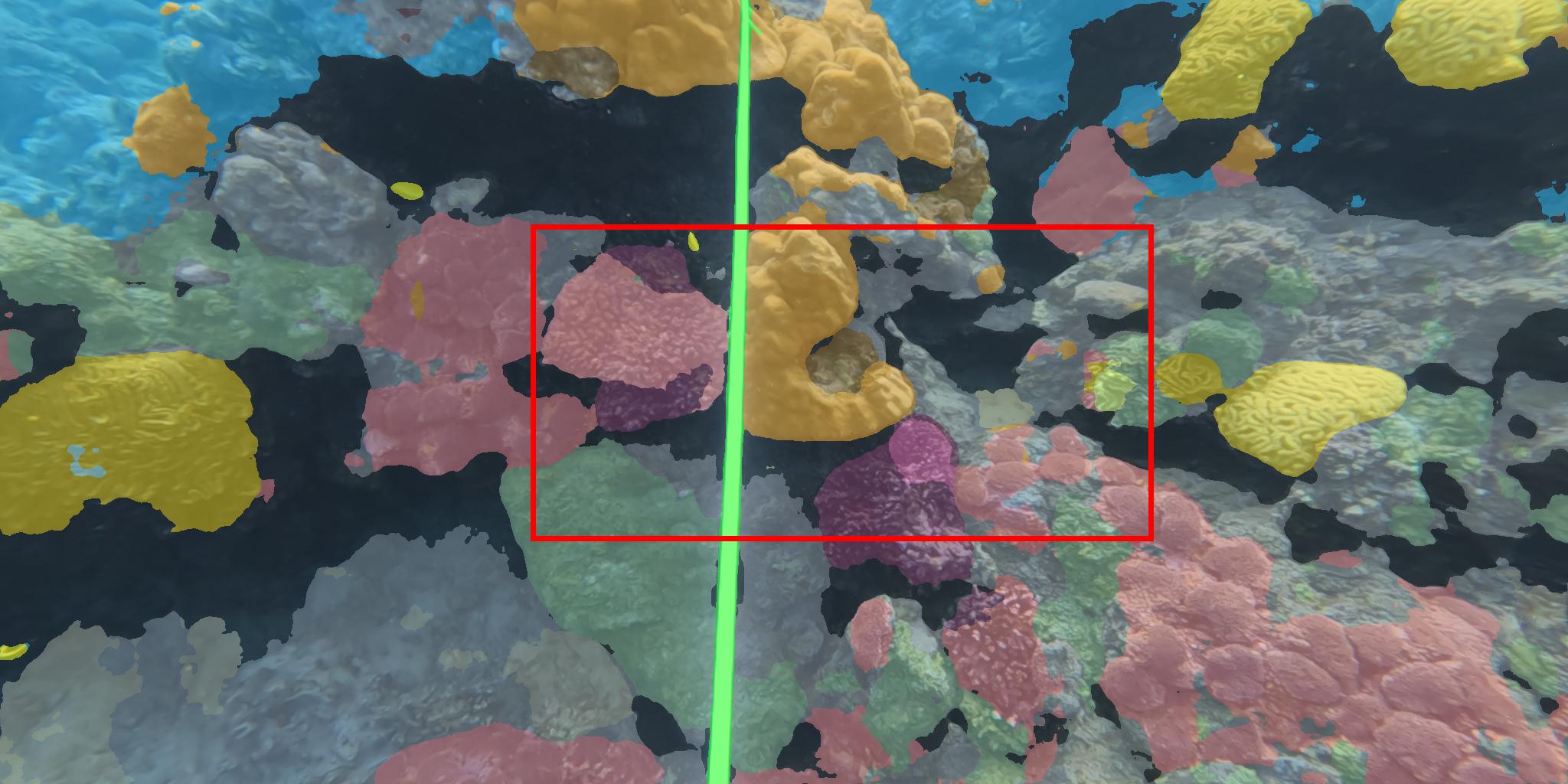} 
        \includegraphics[width=\linewidth]{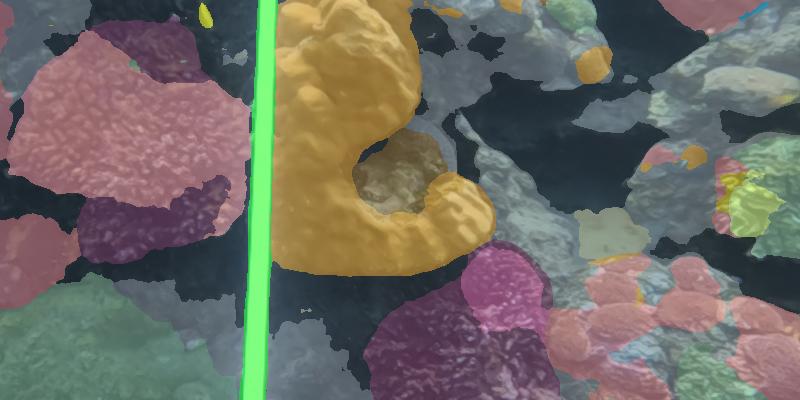}           
        \includegraphics[width=\linewidth]{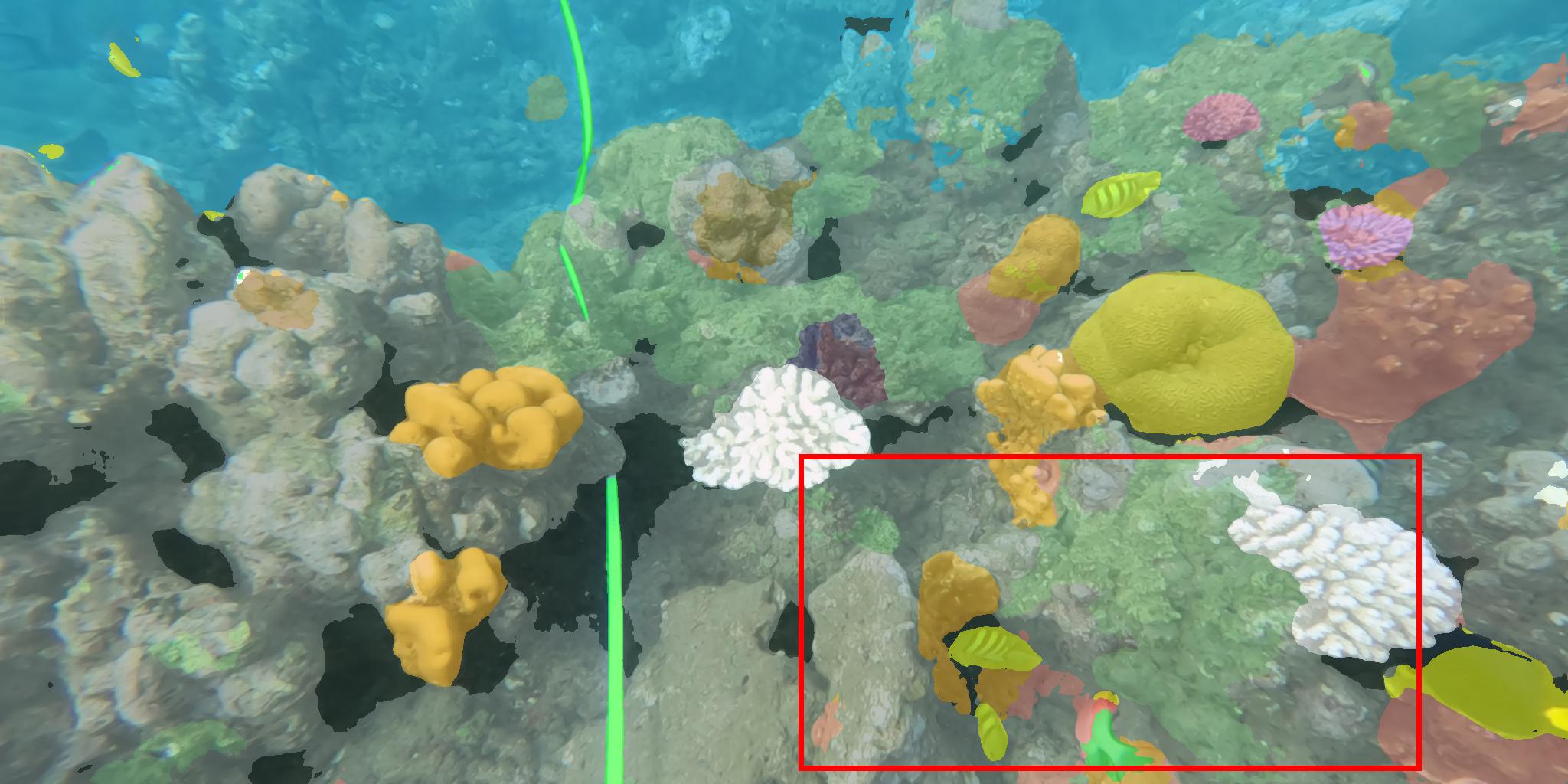} 
        \includegraphics[width=\linewidth]{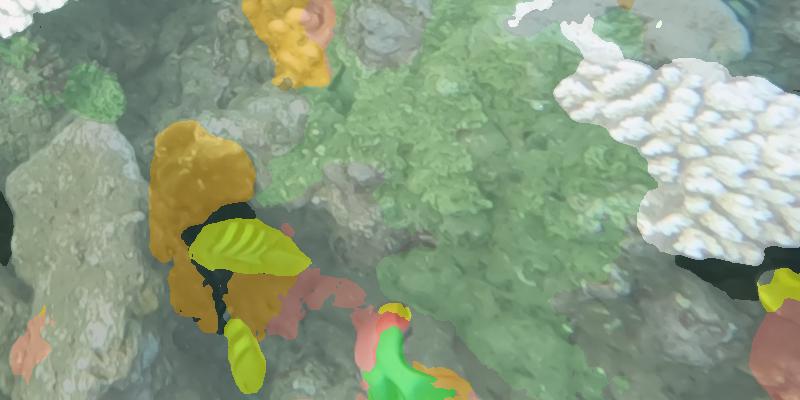}           
        \caption{\\DeepLabV3+}
        \label{fig:sub3_additional}
    \end{subfigure}
    \begin{subfigure}{0.195\textwidth}
        \centering
        \includegraphics[width=\linewidth]{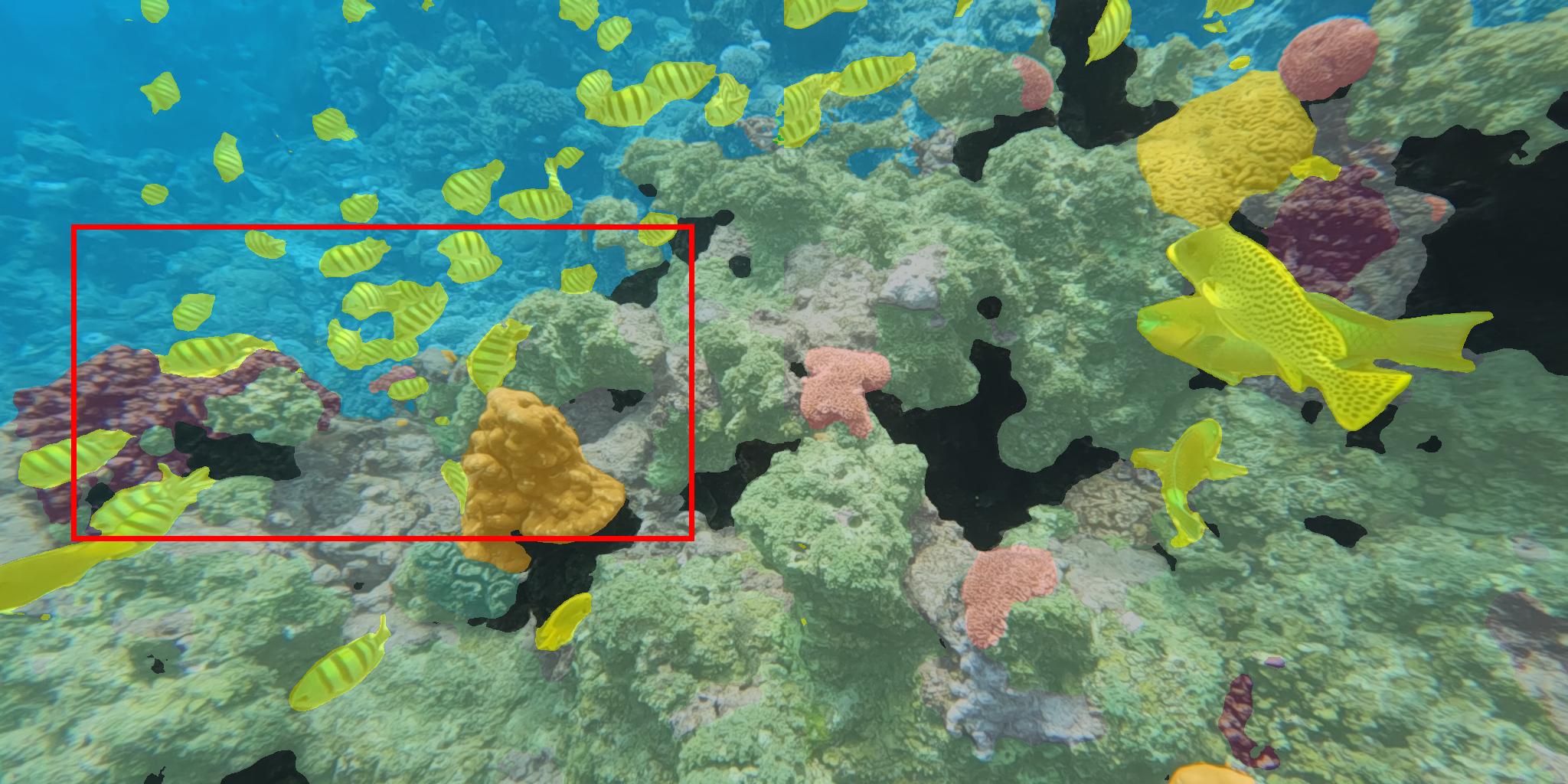} 
        \includegraphics[width=\linewidth]{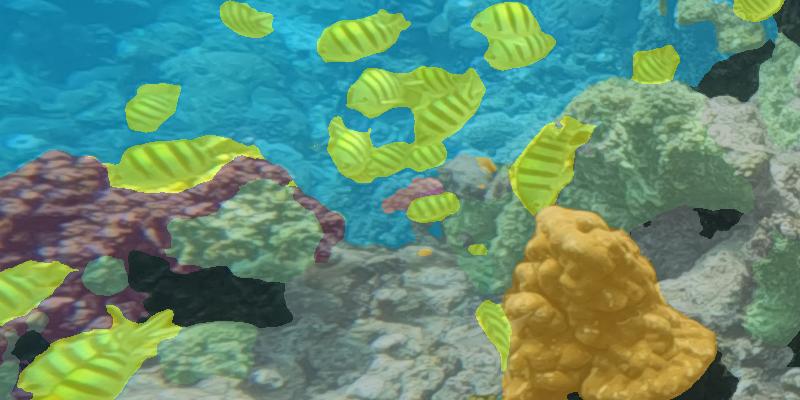} 
        \includegraphics[width=\linewidth]{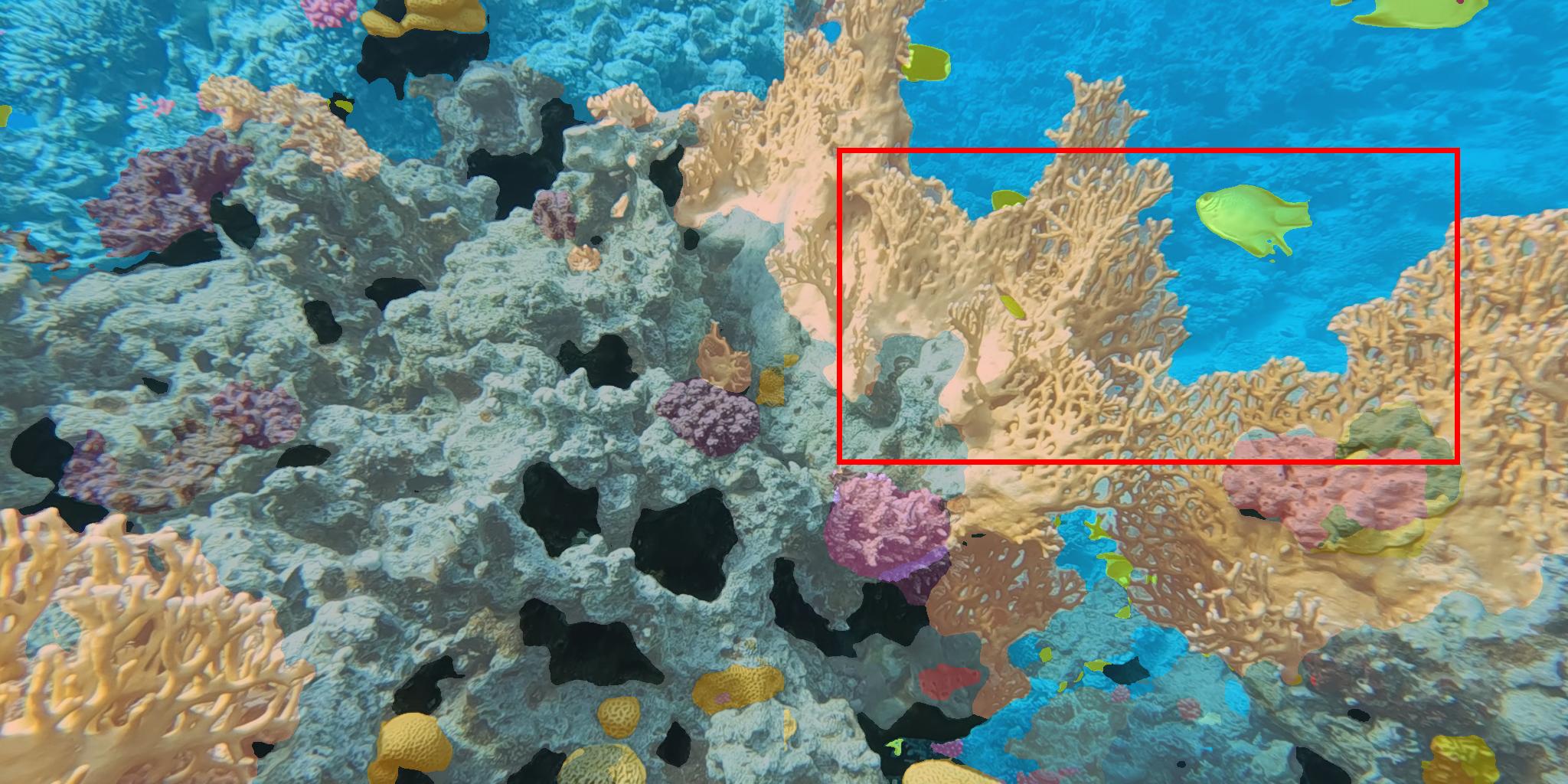}
        \includegraphics[width=\linewidth]{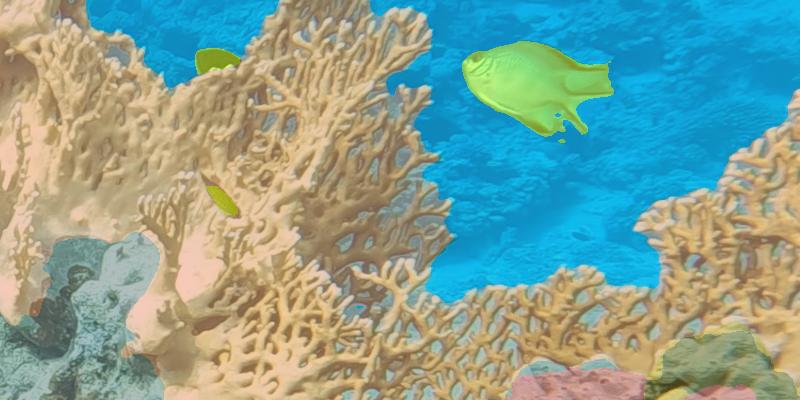} 
        \includegraphics[width=\linewidth]{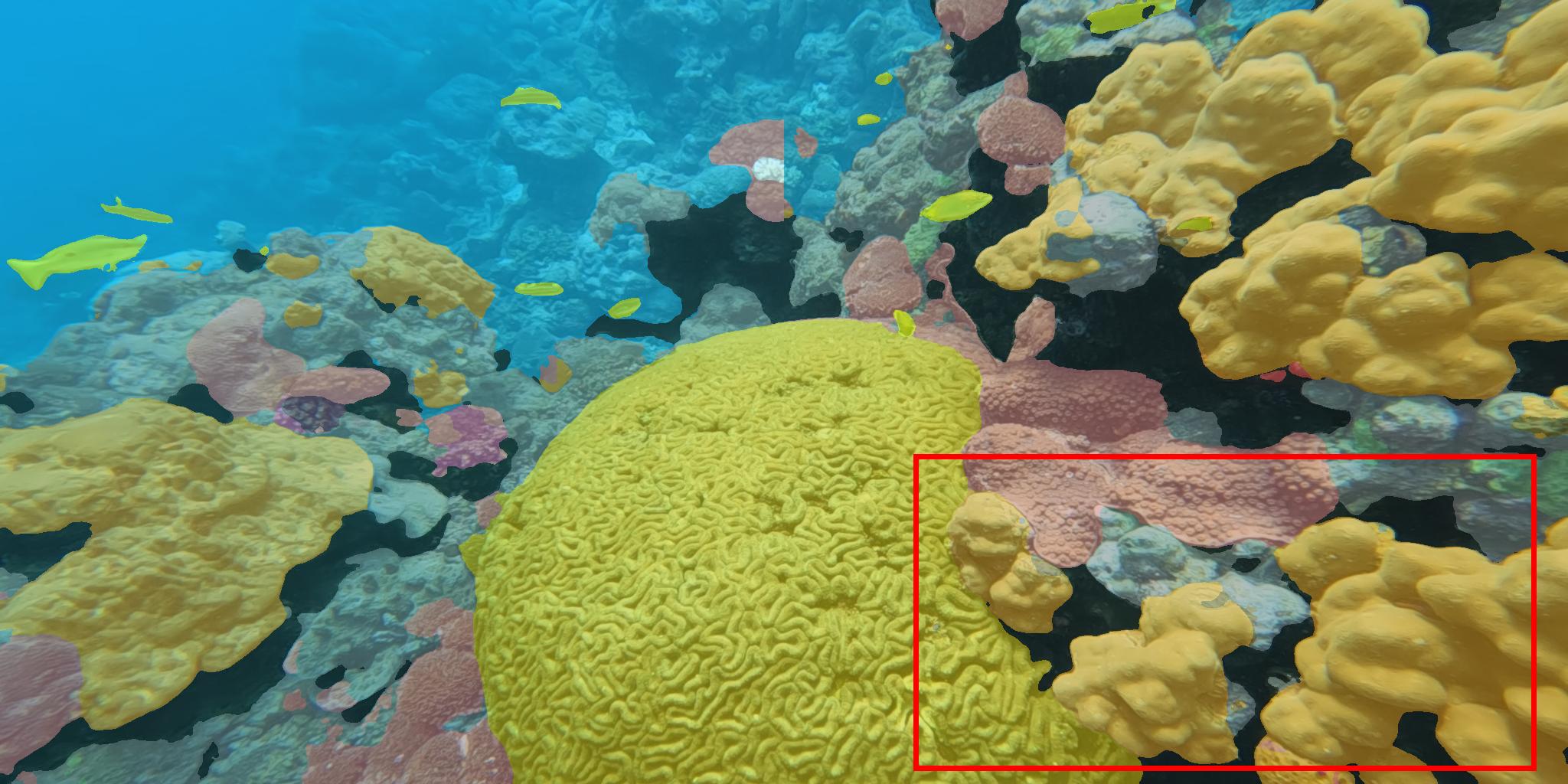} 
        \includegraphics[width=\linewidth]{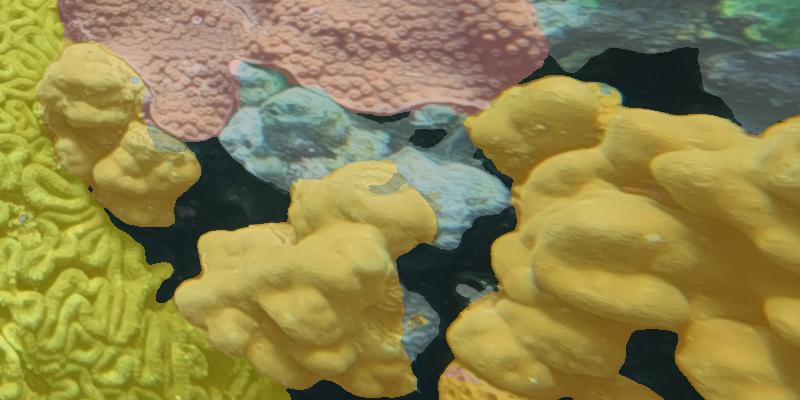} 
        \includegraphics[width=\linewidth]{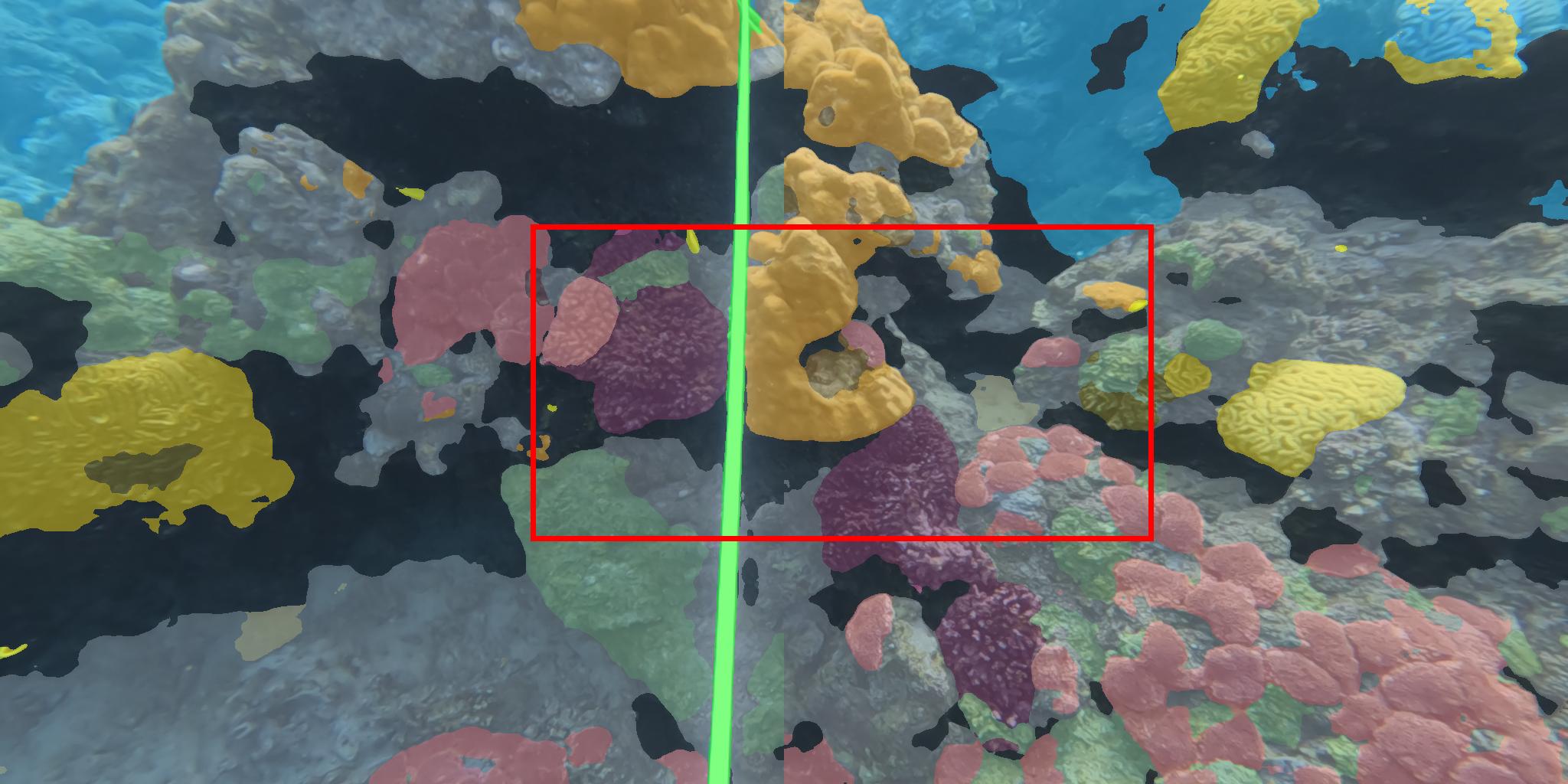} 
        \includegraphics[width=\linewidth]{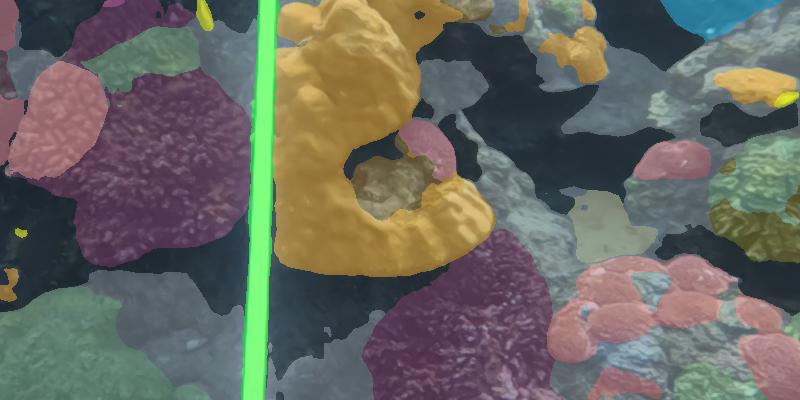}      \includegraphics[width=\linewidth]{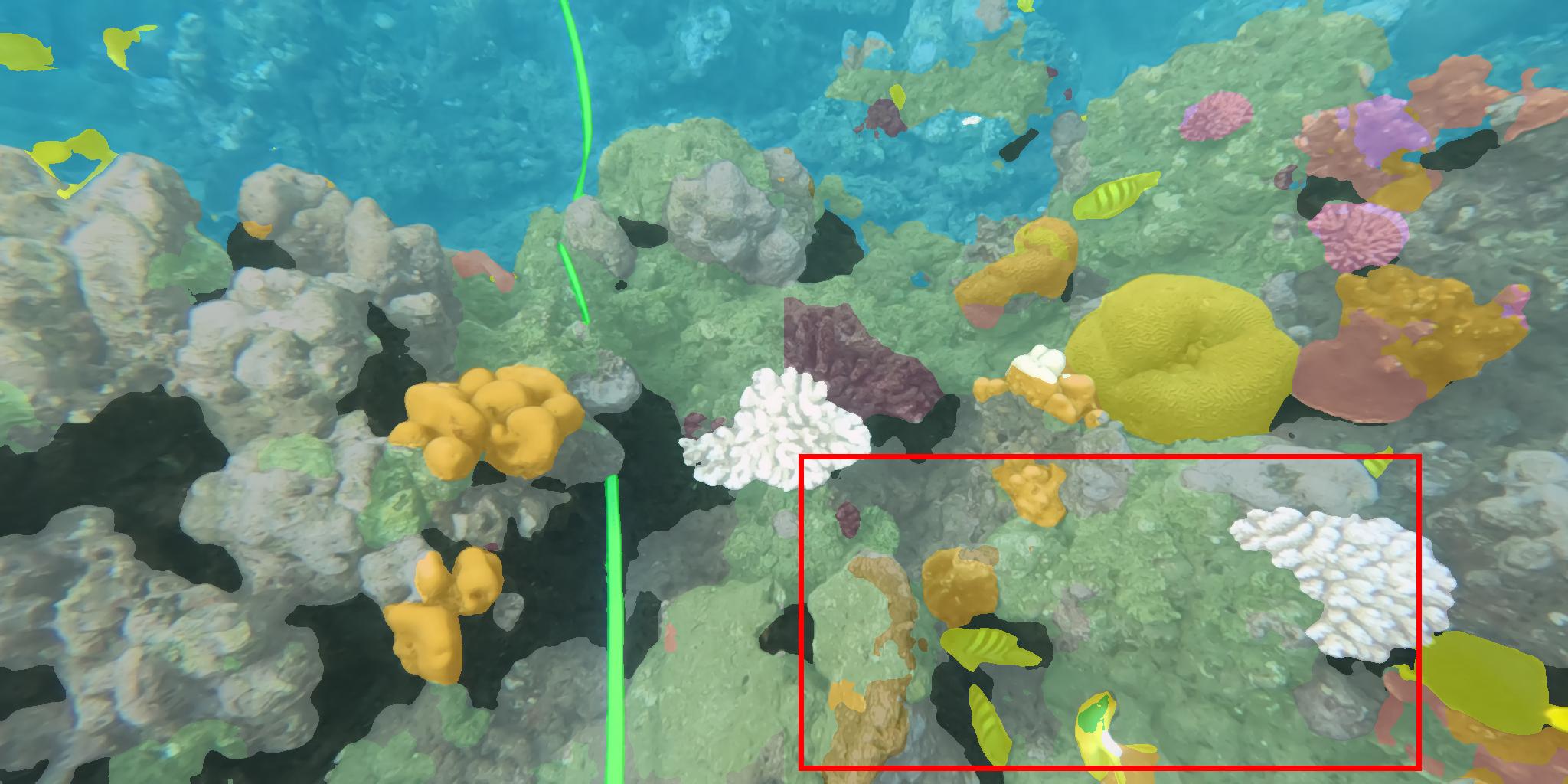} 
        \includegraphics[width=\linewidth]{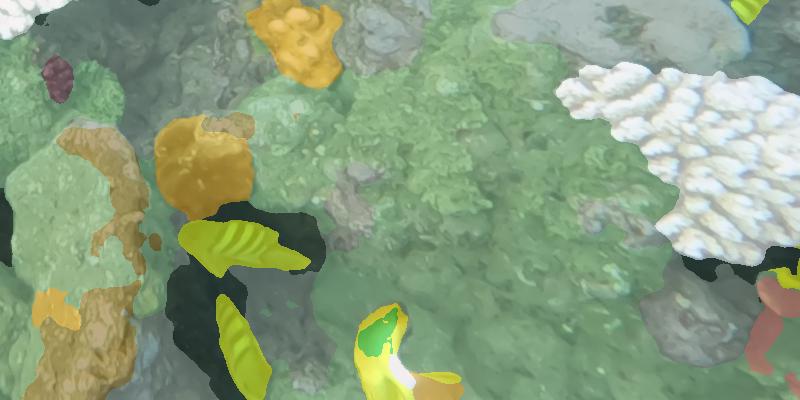}          
        \caption{\\SegFormer - MiT-b2}
        \label{fig:sub4_additional}
    \end{subfigure}
    \begin{subfigure}{0.195\textwidth}
        \centering
        \includegraphics[width=\linewidth]{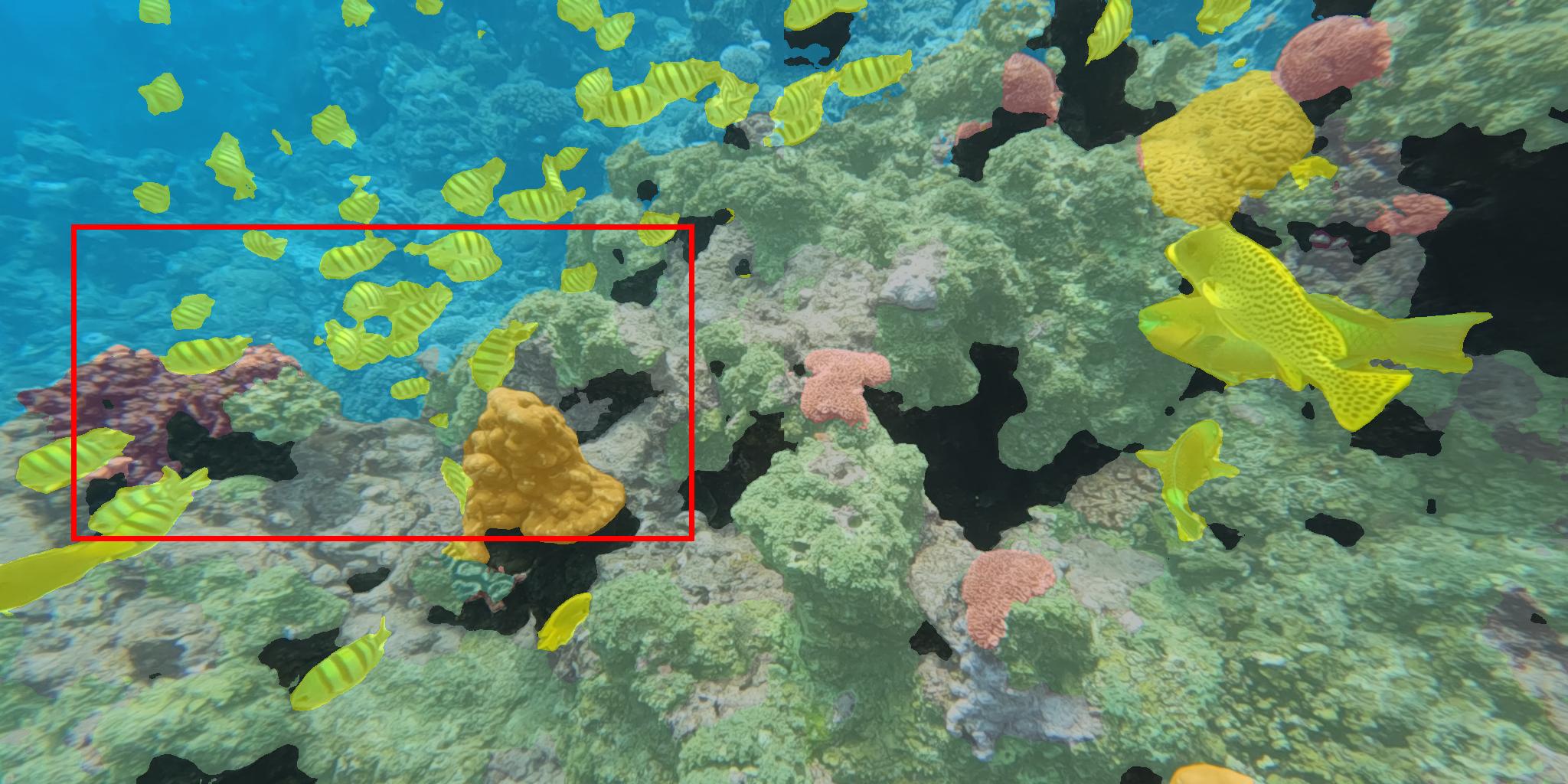} 
        \includegraphics[width=\linewidth]{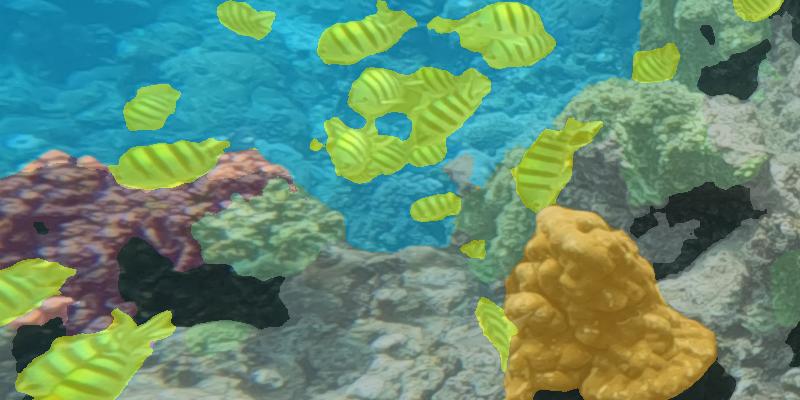} 
        \includegraphics[width=\linewidth]{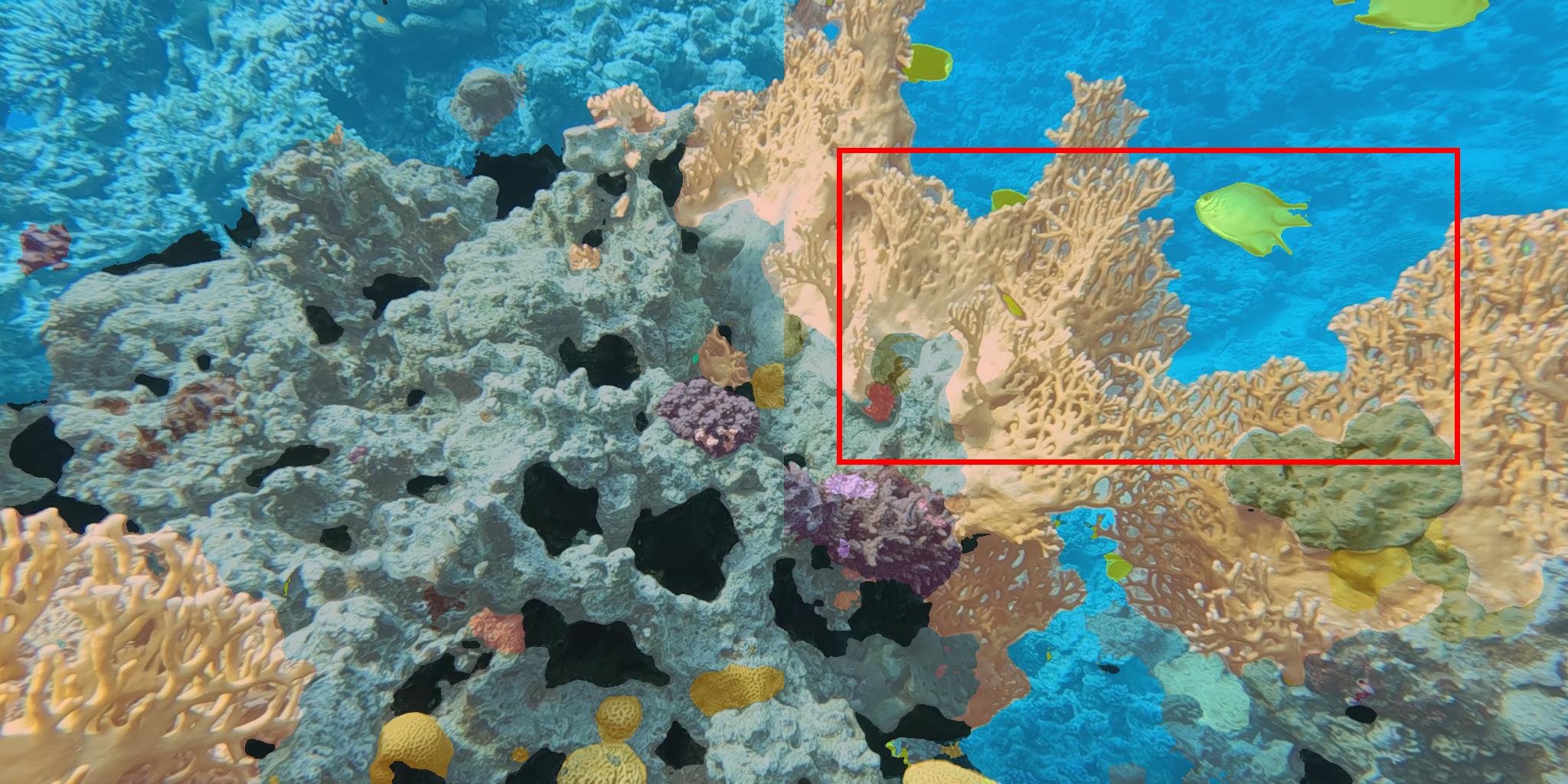}
        \includegraphics[width=\linewidth]{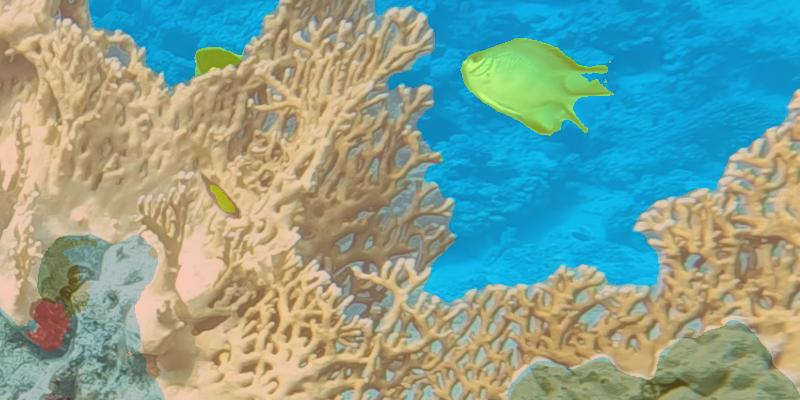} 
        \includegraphics[width=\linewidth]{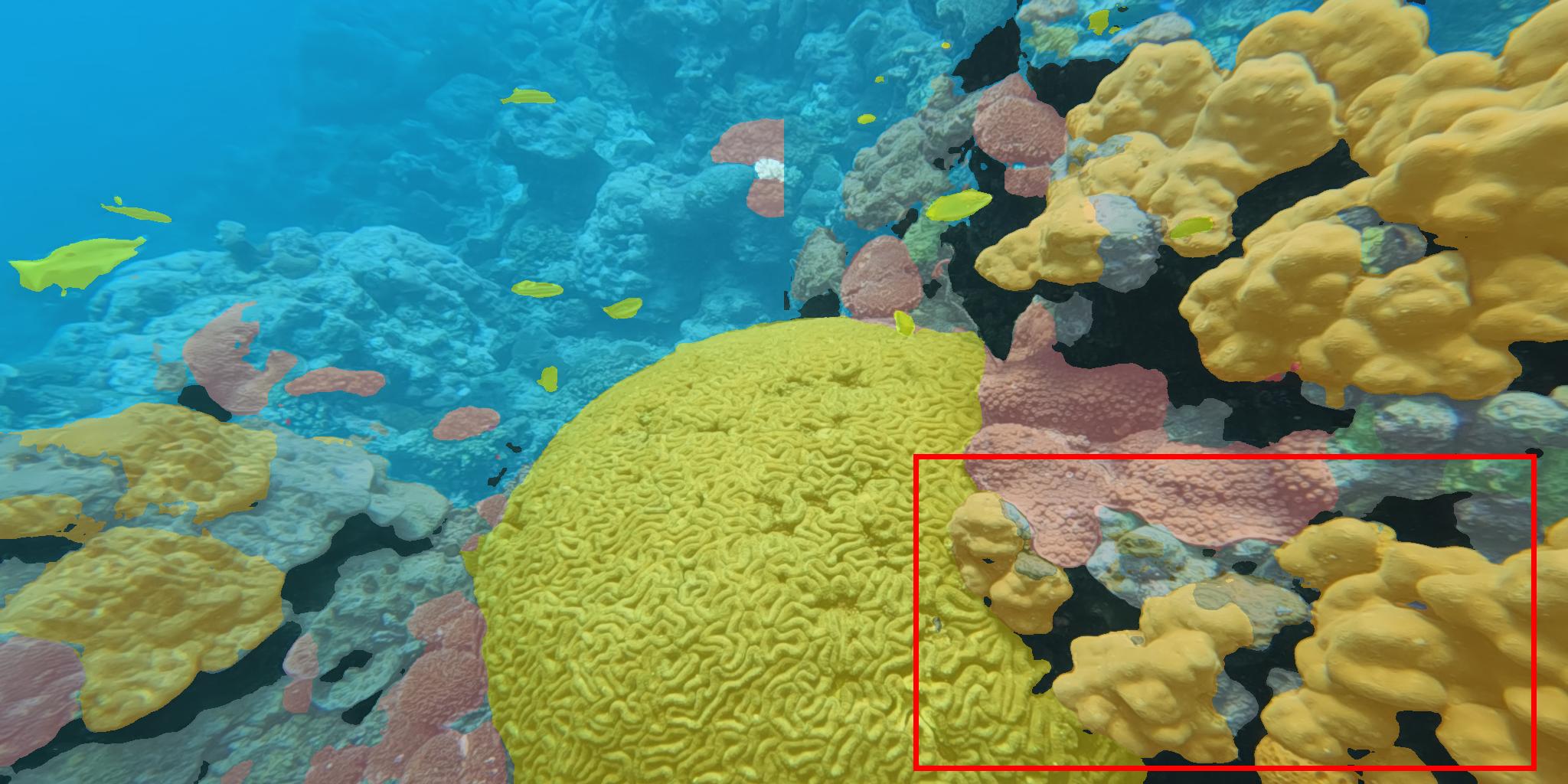} 
        \includegraphics[width=\linewidth]{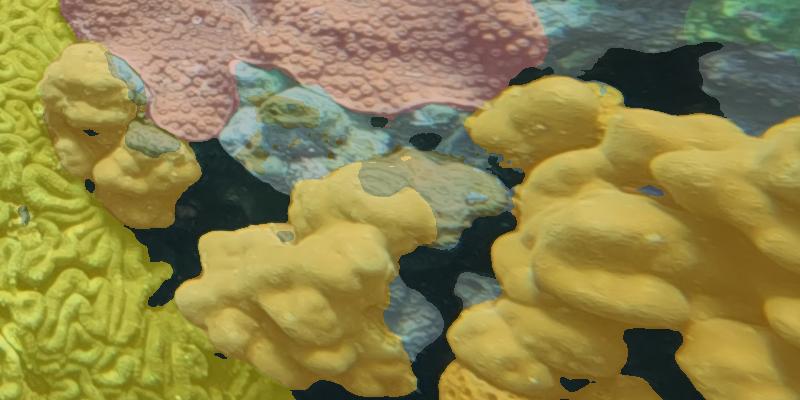} 
        \includegraphics[width=\linewidth]{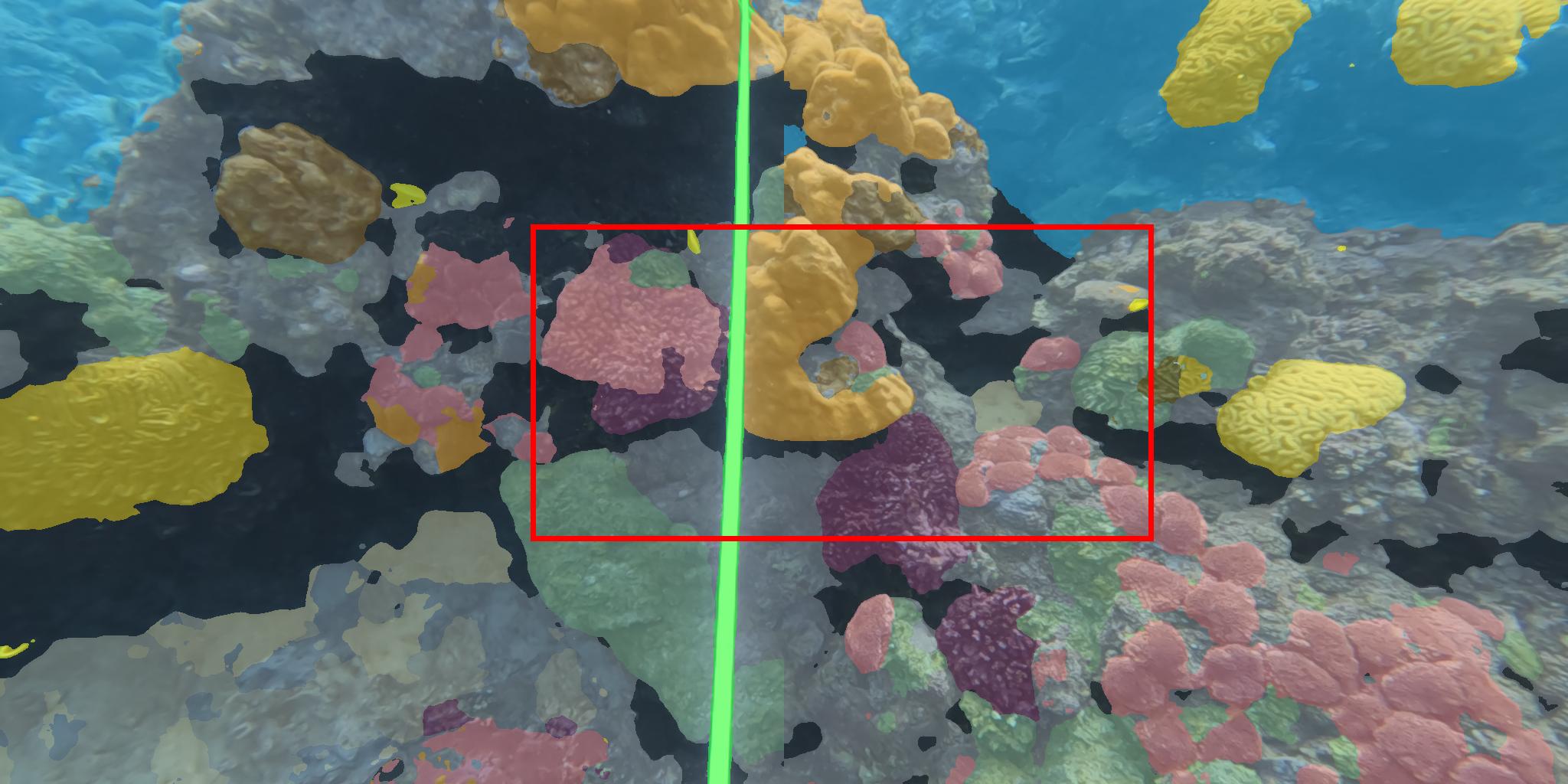} 
        \includegraphics[width=\linewidth]{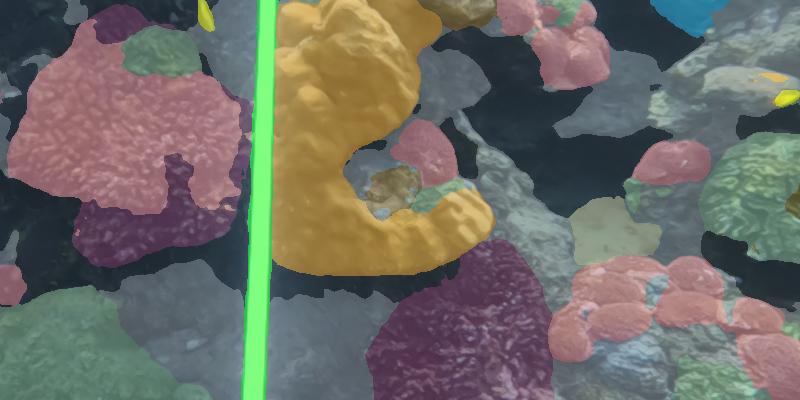}        
        \includegraphics[width=\linewidth]{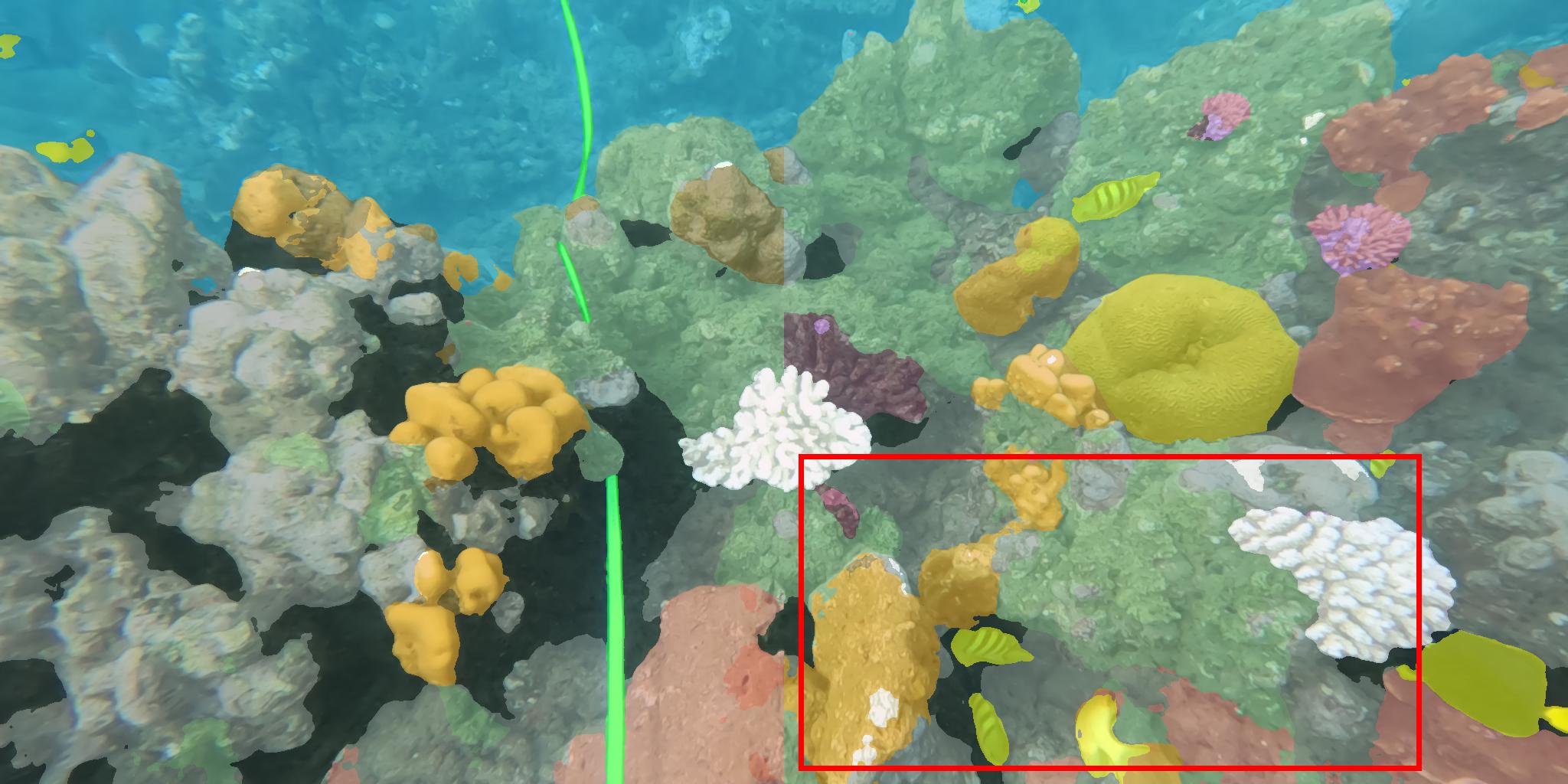} 
        \includegraphics[width=\linewidth]{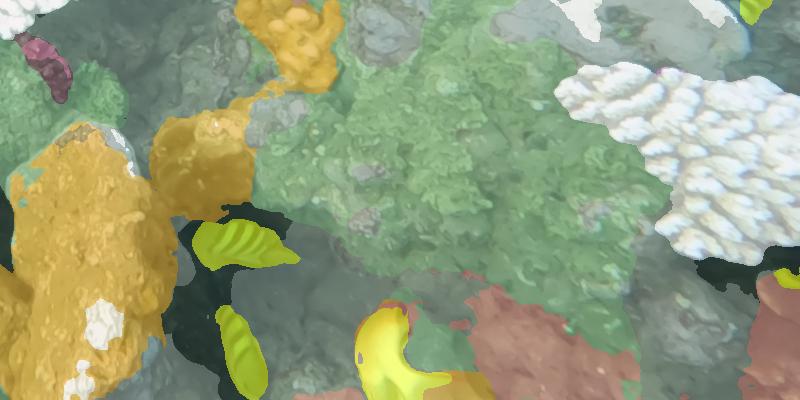}        
        \caption{\\SegFormer - MiT-b5}
        \label{fig:sub5_additional}
    \end{subfigure}
    \caption{Additional qualitative samples from the Coralscapes test set.}
    \label{fig:main_additional}
\end{figure*}

\end{document}